\documentclass[oneside,10pt]{article}

\usepackage{array}
\usepackage{amsthm,amsmath,amsfonts,amssymb}
\usepackage[caption=false,font=normalsize,labelfont=sf,textfont=sf]{subfig}
\usepackage{textcomp}
\usepackage{url}
\usepackage{verbatim}
\usepackage{graphicx}
\usepackage{tikz}
\usepackage{cite}
\usepackage{xfp}
\usepackage{adjustbox}
\usepackage{longtable}
\usepackage{enumitem}
\usepackage{graphicx} 
\usepackage{listings}   % For better code formatting
\usepackage{xcolor}     % For colored text and backgrounds
\usepackage[many]{tcolorbox}  % For nicer colored boxes (alternative to fbox)
\usepackage{makecell} 
\usepackage{pifont}% http://ctan.org/pkg/pifont
\newcommand{\cmark}{\ding{51}}%
\newcommand{\xmark}{\ding{55}}%
\usepackage{algorithm}  % For algorithm environments
\usepackage{algpseudocode} % For algorithm pseudocode
\usepackage[commentColor=black,beginLComment=/*~, endLComment=~*/]{algpseudocodex}
\usepackage{booktabs}       % professional-quality tables
\usepackage{multirow} 
\usepackage{nicefrac}       % compact symbols for 1/2, etc.
\usepackage{hyperref}       % hyperlinks
\usepackage{tcolorbox}
\tcbuselibrary{breakable, skins}

	\newtcolorbox{promptbox}[1][]{
		enhanced,
		breakable,
		colback=white,
		colframe=black,
		boxrule=0.4pt,
		left=6pt, right=6pt, top=6pt, bottom=6pt,
		fontupper=\ttfamily\footnotesize\raggedright,
		before upper={\setlength{\parindent}{0pt}\setlength{\parskip}{4pt}},
		#1
	}
\theoremstyle{plain}
\newtheorem{theorem}{Theorem}
\newtheorem{lemma}[theorem]{Lemma}

\newtheorem{corollary}[theorem]{Corollary}

\theoremstyle{definition}
\newtheorem{definition}{Definition}
\newtheorem{example}{Example}

\theoremstyle{remark}
\newtheorem{remark}{Remark}

\DeclareMathOperator{\tr}{tr}

\usepackage{etoolbox}
\makeatletter
\patchcmd{\@makecaption}
  {\scshape}
  {}
  {}
  {}
\makeatother

\tikzset{
	dotnode/.style = {
		circle, fill=black, inner sep=0pt,
		minimum size=3pt, outer sep=1.5pt
	},
	graph layout/.style args={nodes=#1}{
		execute at begin picture={
			\pgfkeys{/graph/nodes=#1}
			\setcounter{graphnode}{0}
		}
	},
	selfloop/.style={out=120, in=60, distance=15mm, looseness=8},
	selfloop left/.style={out=150, in=210, distance=15mm, looseness=8},
	selfloop right/.style={out=30, in=-30, distance=15mm, looseness=8},
	selfloop bottom/.style={out=210, in=330, distance=15mm, looseness=8},
	edge label/.style={midway, fill=white, inner sep=1.5pt, font=\footnotesize, text=red}
}

\newcounter{graphnode}
\pgfkeys{/graph/nodes/.initial}
\newcommand{\newnode}{%
	\stepcounter{graphnode}%
	\pgfkeysgetvalue{/graph/nodes}{\currentn}%
	\ifnum\currentn=2
	\ifcase\value{graphnode}
	\or \node[dotnode] (1) at (\fpeval{cosd(180)},\fpeval{sind(180)}) {};
	\or \node[dotnode] (2) at (\fpeval{cosd(0)},\fpeval{sind(0)}) {};
	\fi
	\else\ifnum\currentn=3
	\ifcase\value{graphnode}
	\or \node[dotnode] (1) at (\fpeval{cosd(180)},\fpeval{sind(180)}) {};
	\or \node[dotnode] (2) at (\fpeval{cosd(60)},\fpeval{sind(60)}) {};
	\or \node[dotnode] (3) at (\fpeval{cosd(-60)},\fpeval{sind(600)}) {};
	\fi
	\else\ifnum\currentn=4
	\ifcase\value{graphnode}
	\or \node[dotnode] (1) at (\fpeval{cosd(180)},\fpeval{sind(180)}) {};
	\or \node[dotnode] (2) at (\fpeval{cosd(90)},\fpeval{sind(90)}) {};
	\or \node[dotnode] (3) at (\fpeval{cosd(0)},\fpeval{sind(0)}) {};
	\or \node[dotnode] (4) at (\fpeval{cosd(-90)},\fpeval{sind(-90)}) {};
	\fi
	\else\ifnum\currentn=5
	\ifcase\value{graphnode}
	\or \node[dotnode] (1) at (\fpeval{cosd(180)},\fpeval{sind(180)}) {};
	\or \node[dotnode] (2) at (\fpeval{cosd(108)},\fpeval{sind(108)}) {};
	\or \node[dotnode] (3) at (\fpeval{cosd(36)},\fpeval{sind(36)}) {};
	\or \node[dotnode] (4) at (\fpeval{cosd(-36)},\fpeval{sind(-36)}) {};
	\or \node[dotnode] (5) at (\fpeval{cosd(-108)},\fpeval{sind(-108)}) {};
	\fi
	\else\ifnum\currentn=6
	\ifcase\value{graphnode}
	\or \node[dotnode] (1) at (\fpeval{cosd(180)},\fpeval{sind(180)}) {};
	\or \node[dotnode] (2) at (\fpeval{cosd(120)},\fpeval{sind(120)}) {};
	\or \node[dotnode] (3) at (\fpeval{cosd(60)},\fpeval{sind(60)}) {};
	\or \node[dotnode] (4) at (\fpeval{cosd(0)},\fpeval{sind(0)}) {};
	\or \node[dotnode] (5) at (\fpeval{cosd(-60)},\fpeval{sind(-60)}) {};
	\or \node[dotnode] (6) at (\fpeval{cosd(-120)},\fpeval{sind(-120)}) {};
	\fi
	\else\ifnum\currentn=7
	\ifcase\value{graphnode}
	\or \node[dotnode] (1) at (\fpeval{cosd(180)},\fpeval{sind(180)}) {};
	\or \node[dotnode] (2) at (\fpeval{cosd(180-360/7)},\fpeval{sind(180-360/7)}) {};
	\or \node[dotnode] (3) at (\fpeval{cosd(180-2360/7)},\fpeval{sind(180-2360/7)}) {};
	\or \node[dotnode] (4) at (\fpeval{cosd(180-3360/7)},\fpeval{sind(180-3360/7)}) {};
	\or \node[dotnode] (5) at (\fpeval{cosd(180-4360/7)},\fpeval{sind(180-4360/7)}) {};
	\or \node[dotnode] (6) at (\fpeval{cosd(180-5360/7)},\fpeval{sind(180-5360/7)}) {};
	\or \node[dotnode] (7) at (\fpeval{cosd(180-6360/7)},\fpeval{sind(180-6360/7)}) {};
	\fi
	\fi\fi\fi\fi\fi\fi
}

\begin{document}

%\title{Counting Cycles with AI}
%\title{Computationally Efficient Formulas for Cycle Count Statistics via Human--AI Collaboration}
\title{Counting Cycles with AI: Computationally Efficient Equivalent Forms with Applications}

\author{Jiashun Jin, Zheng Tracy Ke, Bingcheng Sui, and Zhenggang Wang
	\thanks{J. Jin is with the Department of Statistics and Data Science, Carnegie Mellon University, Pittsburgh, PA 15213, USA (e-mail: jiashun@stat.cmu.edu).}
	\thanks{Z.T. Ke is with the Department of Statistics, Harvard University, Cambridge, MA 02138, USA (e-mail: zke@fas.harvard.edu).}
	\thanks{B. Sui is with the School of the Gifted Young, University of Science and Technology of China, Hefei, Anhui 230026, China (e-mail: suibc@mail.ustc.edu.cn).}
	\thanks{Z. Wang is with the School of Statistics and Data Science, Southeast University, Nanjing, Jiangsu 211189, China (e-mail: zggwang@seu.edu.cn).}
}

\maketitle

% The \author macro works with any number of authors. There are two commands
% used to separate the names and addresses of multiple authors: \And and \AND.
%
% Using \And between authors leaves it to LaTeX to determine where to break the
% lines. Using \AND forces a line break at that point. So, if LaTeX puts 3 of 4
% authors names on the first line, and the last on the second line, try using
% \AND instead of \And before the third author name.

\begin{abstract}  
Cycle count statistics are fundamental tools in  statistics and engineering,   with applications in motif counting, channel coding, and statistical inference of network and matrix data. 
However, how to compute high-order cycle count statistics efficiently is still an open problem. In this paper, we aim to derive Computationally Efficient Equivalent Forms (CEEF) for cycle count statistics of any given order,   
where we express each cycle count statistic equivalently as a linear combination of finitely many terms. Using the CEEF,  
 we provide a much more efficient way to compute the cycle count statistics. 

The CEEF problem has no known general solution and requires delicate combinatorial arguments together with extensive calculations. While this task is hard to accomplish by humans alone,  it provides an ideal setting in which Artificial Intelligence (AI) can be useful. 
We solve the problem by combining several theorems we derive with powerful coding skills of modern AI systems. Our results leverage graph-theoretic arguments and yield new formulas for general cases that were previously unknown. We find that, although AI cannot solve the problem independently, it becomes highly effective when guided by humans through theorems we derive as well as a clear derivation strategy, step-by-step instructions, and carefully-written prompts. 

We consider several statistical applications, including spiked matrix testing, estimation of weak spike eigenvalues, and pairwise network comparison. For each problem, we demonstrate that optimal statistical performance is achieved by using high-order cycle count statistics, and our CEEF formulas make their computation feasible on large-scale data sets.  

\end{abstract}
	
%	
%\begin{IEEEkeywords}
%	Cycle counting, graph theory, network analysis, matrix statistics, 
%	combinatorial algorithms, AI-assisted mathematics, low-rank matrix detection.
%\end{IEEEkeywords}

\section{Introduction}\label{sec:intro}   

Cycle count statistics are fundamental tools in engineering, information theory, statistics, and machine learning.  
For instance, cycle counting is a special case of motif counting and plays an important role in biomelacular engineering \cite{milo2002network,alon2008biomolecular}, especially in the studies of protein-protein interaction networks. In channel coding, cycle counts in the Tanner graph are used as important performance metrics for the codes \cite{halford2005algorithm, karimi2012message}. In information theory and statistics, cycle counts and the general sub-graph counts have been used to establish information-theoretic bounds and derive optimal algorithms, such as in graph alignment \cite{mao2024testing,wang2025efficient}, network data inference \cite{bubeck2016testing, banerjee2017optimal}, and covariance matrix testing \cite{chen2010tests,qiu2012test}. 

%Cycle count statistics play an important role in network data analysis \cite{bubeck2016testing, banerjee2017optimal, gao2017testing, Jin2021optimal, EstK, GOF}, covariance matrix testing \cite{Chen}, and spectrum estimation from noisy low-rank matrices \cite{Kong}. These problems often exhibit intrinsic low-rank structure, and cycle count statistics provide accurate estimates of the corresponding spectral moments. In particular, higher-order cycle count statistics have been used to establish fundamental limits and optimality results in statistical inference \cite{bubeck2016testing, banerjee2017optimal, Jin2021optimal}.
%In Section~\ref{sec:applications}, we will present several concrete applications of cycle count statistics and demonstrate that incorporating higher-order cycle counts lead to substantially improved statistical performance.   

%\end{itemize}
%Summary 2: Existing cycle count algorithms in the engineering literature
%\begin{itemize}
%\item Color-coding
%\item BFS algorithm \cite{niu2025fast}
%\item DFS-XOR algorithm \cite{safar2009counting}
%\end{itemize}

Despite their importance, computing high-order cycle count statistics remains a major computational challenge.  
For any $n \times n$ matrix and $m \geq 3$, the order-$m$ cycle count statistics associated with the matrix is a big sum of $O(n^m)$ terms.  Seemingly, the brute-forth approach where we enumerate all $O(n^m)$ terms 
is infeasible when $m$ is large.  Such a challenge motivates the CEEF problem:  how to derive a \textbf{C}omputationally \textbf{E}fficient \textbf{E}quivalent \textbf{F}orm (CEEF) for cycle count statistics, that is,  
an exact formula for the order-$m$ cycle count statistic which allows  us to compute it with a complexity 
much lower than $O(n^m)$.  CEEF formulas of this kind only exist for small orders \cite{harary1969graph, Chang2003, Movarraei2016, Russian, Jin2021optimal}, and their derivations rely on case-by-case arguments rather than a general strategy applicable to an arbitrary order $m$.

In this paper, we develop a general combinatorial framework for solving the CEEF problem for arbitrary cycle orders. The framework introduces several new graph-theoretic concepts, including Fully Annotated Multigraphs (FAMs), root FAMs,  Succinct Expressive Algebraic (SEA) terms,  and Incompressible Full Sum (IFS) terms.  
Here, a FAM is a labeled multi-graph, and each SEA term or IFS term is associated with a specific FAM. 
In particular, fix $m \geq 3$ and consider all FAMs induced by a size-$m$ root FAM. 
We first divide them by size and then by isomorphism.  
This way, we divide all FAMs into a number of isomorphic classes. 
Picking a candidate from each class, we express the order-$m$ cycle count statistic as a linear combination of many terms,  where each term corresponds 
to a candidate FAM,  and each coefficient depends on a few simple characteristics 
of the candidate FAM and the isomorphic class. Moreover, to further reduce the computation 
complexity,  we show that each term can be reduced 
equivalently to either an SEA term or an IFS term.  

Following the above ideas, we can derive a symbolic formula for the order-$m$ cycle count statistic, given an arbitrary order $m$. The formula is mathematically correct, but to determine each term and each linear coefficient explicitly, 
we need to enumerate all candidate FAMs. Unfortunately,    such a task 
is merely too hard for human. Take $m = 12$ for example.  The resulting CEEF formula consists of 1,900 terms, each corresponding to a candidate FAM with a coefficient to be determined. 
Enumerating all 1,900 FAMs are extremely labor-intensive and prone to error, making the task nearly impossible  for human to accomplish.

%
%
%This is where the AI comes in. 
%The remaining challenge is to how to execute an explicit formula for cycle count statistics at each given order. To do so, we must enumerate all isomorphic classes and 
%select a candidate from each class.  
%Although each step of what we have described above is mathematically well defined, the execution requires complicated combinatorics and becomes increasingly tedious as the cycle order grows. For example, when the cycle order is 12, the resulting CEEF formula consists of 1,900 terms, each corresponds to a candidate FAM with a coefficient to be determined. 
%Carrying out these calculations manually is extremely labor-intensive and prone to error, making the overall procedure impractical in practice.

%
%The results above can also be viewed as a step-by-step strategy: 
%if we can enumerate all 
%
%
%
%If we are able to enumerate all underlying graphs, then we are able to find each 
%term precisely, 

%
%together with an algorithm 
%  %(which is in fact a step-by-step proof strategy) 
%  for deriving the CEEF formulas. We prove that this algorithm correctly produces a CEEF formula for any prescribed cycle order. 
  %, provided that each of its steps is carried out correctly.

Recent advances in Artificial Intelligence (AI) provide a promising alternative for carrying out this kind of task.  In particular, for problems involving delicate and tedious combinatorial arguments, AI may have certain advantages over humans. 
 However, a straightforward use of AI is insufficient: current AI systems are unable to solve the CEEF problem reliably through direct prompting alone. We therefore develop a human--AI collaboration workflow in which AI is not asked to solve the CEEF problem from scratch. Instead, humans first work out a list of theorems to lay out the theoretical foundation of solving the problem.   Using these theorems, humans then design a rigorous algorithm with multiple steps (which can also be viewed as a multi-step reasoning and derivation strategy),  and AI is then guided to understand and execute its individual steps. In summary, humans are responsible for the high-level architecture, 
 and AI is responsible for the specific steps; humans are responsible for mathematics, and AI is responsible for coding and execution.

This division of labor is essential. Current AI systems do not appear capable of independently solving a problem as technically demanding as CEEF, making the human contribution of a correct high-level reasoning and derivation strategy indispensable. Conversely, AI is well suited to executing many labor-intensive but well-defined subproblems once the high-level strategy is determined. We believe that this workflow is useful broadly for research-level problems in AI for Mathematics (AI4Math) \cite{Drori2022}.

The remaining part  of this section is organized as follows. We first formally define the CEEF problem in Section~\ref{subsec:CEEF}. We then give a high-level description of our combinatorial framework and human--AI collaboration workflow in Section~\ref{subsec:humAI-intro}, positioning our work in the literature, and emphasizing why the resulting mathematical and algorithmic contributions are nontrivial. Finally, we discuss the connection between our work and recent developments in AI4Math in Section~\ref{subsec:connection}.

%The use of AI tools to solve mathematical problems has attracted significant recent attention (e.g., \cite{hendrycks2021math, zheng2022minif2f, Davies2021}) and has motivated the emerging research area of  AI4Math \cite{Drori2022}.  However,  except a few notable works  (e.g., \cite{romera2024funsearch, alfarano2024lyapunov, bubeck2025early, dobriban2025solving, alexeev2026short}), most existing studies focus on mathematical problems with known solutions (e.g., \cite{Trinh, lample2022htps}). Their primary goal is to evaluate the performance of AI rather than to use AI to solve open research problems (see Section \ref{subsec:connection}). 
%%
%For this reason,  how to use AI tools to solve {\it difficult research-level problems without known solutions}  remains an open yet important question in AI4Math. Major progress in this direction would not only mark a milestone for the reasoning model, but could also fundamentally reshape mathematical practice. 
%	
%We focus on a long-standing open problem: deriving a {\bf C}omputationally {\bf E}fficient {\bf E}quivalent {\bf F}orm (CEEF) for the cycle count statistic. The CEEF problem involves delicate and extremely tedious combinatorial arguments, making it difficult to solve by humans alone.  However,  by using a novel {\it hugAI approach} that combines human and artificial intelligence, we are able to solve this problem satisfactorily.  
	
\subsection{The CEEF problem}  \label{subsec:CEEF} 	
For any symmetric matrix $A\in\mathbb{R}^{n\times n}$ and  integer $m \geq 3$, the order-$m$ cycle count (CC) statistic is defined by 
\begin{equation} \label{Defi:Cm}
C_m = \sum_{1\leq i_1,  \ldots, i_m (dist)\leq n} A_{i_1 i_2} A_{i_2 i_3} \ldots A_{i_m i_1},  \qquad  \mbox{where `dist' stands for `distinct'}. 
\end{equation}
Because of the restriction that $i_1, \ldots, i_m$ should be distinct, the above definition does not depend on the diagonal of $A$. Without loss of generality, we assume 
\begin{equation} \label{diagA} 
A_{ii} = 0,   \qquad 1 \leq i \leq n.   
\end{equation} 
In the special case where $A$ is the adjacency matrix of an undirected network (which we refer to as the {\it binary case}),  $C_m / (2m)$ is the number of $m$-cycles in the network \cite{Jin2021optimal}.  This is why we call $C_m$ the term {\it cycle count statistics}.   
	
%How to compute $C_m$ is known as a challenging problem.  A {\it brute-force} approach is to compute $C_m$ using $m$-layers of for-loops (see (\ref{Defi:Cm})), but the  computation cost is $O(n^m)$. Alternatively,  we may first derive a formula for $C_m$ where we  write $C_m$ as the linear combination of many easy-to-compute terms. Using such a formula, we can compute $C_m$ with much lower costs. 

%Computing $C_m$ is known to be challenging. 
As a brute-force approach to computing $C_m$, we may use $m$ nested for-loops, but this incurs a computational complexity of $O(n^m)$. Alternatively, we may derive a formula that expresses $C_m$ as a linear combination of many easy-to-compute terms. Such a formula can substantially reduce the computation cost.  
Below, we present two examples.  Let ${\bf 1}_n$ denote the $n$-dimensional vector of ones.  For any $n\times n$ matrices $B$ and $C$, let $d(B)$ denote the diagonal matrix whose diagonal entries are the same as those in $B$, and let  $B \circ C$ represent their Hadamard (or entry-wise) product, while  $B\cdot C$ denotes the matrix multiplication.

\begin{example}  \label{example1}
When $m = 4$, $C_4   =     {\bf 1}_n' \cdot (A \circ A \circ A \circ A) \cdot {\bf 1}_n  - 2  [{\bf 1}_n' \cdot (((A \circ A) \cdot {\bf 1}_n) \circ ((A \circ A) \cdot {\bf 1}_n))]   +   {\bf 1}_n' \cdot (A \circ ((A \cdot A) \cdot A)  \cdot {\bf 1}_n)$, 
where each term on the right hand side (RHS) can be computed with a computational complexity of $O(n^3)$.  
\end{example}
	
\begin{example} \label{example2}
When $m = 8$, $C_8$ is a linear combination of $44$ terms, as shown in Table \ref{tb:C8}. 
Among them,  $43$ ones are similar to those in Example~\ref{example1} and can be computed with a complexity of $O(n^3)$, 
and the remaining one (marked with $\star\star$) has the form:   
\begin{equation} \label{IFSterm} 
\sum_{1\leq i_1, i_2, i_3, i_4\leq n}  (A \circ A)_{i_1 i_2} A_{i_1 i_3} A_{i_1 i_4} A_{i_2 i_3} A_{i_2 i_4} (A \circ A)_{i_3 i_4}.  
\end{equation} 
This term can be computed with a complexity of $O(n^4)$.  
\end{example}

%%%%%%%%%%%%%%
%%%%%%%%%%%%%%
\begin{table}[tbp]
	\centering
	\caption{The formula of $C_8$ from our approach. The term marked by $\star\star$ is an IFS term, and the remaining terms are SEA terms.} \label{tb:C8} 	\small
	\scalebox{.72}{
		\begin{tabular}{>{\raggedleft\arraybackslash}p{0.06\textwidth}p{0.6\textwidth}}
			+36 & ${\bf 1}_n' \cdot ((A \circ A \circ A \circ A \circ A \circ A \circ A \circ A) \cdot {\bf 1}_n)$ \\
			-96 & ${\bf 1}_n' \cdot (((A \circ A) \cdot {\bf 1}_n) \circ ((A \circ A \circ A \circ A \circ A \circ A) \cdot {\bf 1}_n))$ \\
			-36 & ${\bf 1}_n' \cdot (((A \circ A \circ A \circ A) \cdot {\bf 1}_n) \circ ((A \circ A \circ A \circ A) \cdot {\bf 1}_n))$ \\
			-112 & ${\bf 1}_n' \cdot ((A \circ A \circ A \circ A \circ ((A \circ A) \cdot (A \circ A))) \cdot {\bf 1}_n)$ \\
			+32 & ${\bf 1}_n' \cdot ((A \circ A \circ A \circ A \circ A \circ ((A \cdot A) \cdot A)) \cdot {\bf 1}_n)$ \\
			+72 & ${\bf 1}_n' \cdot ((((A \circ A) \cdot {\bf 1}_n) \circ ((A \circ A) \cdot {\bf 1}_n)) \circ ((A \circ A \circ A \circ A) \cdot {\bf 1}_n))$ \\
			+16 & ${\bf 1}_n' \cdot (((A \circ A) \cdot ((A \circ A) \cdot {\bf 1}_n)) \circ ((A \circ A \circ A \circ A) \cdot {\bf 1}_n))$ \\
			+80 & ${\bf 1}_n' \cdot ((A \circ A \circ A \circ A \circ (A \cdot A) \circ (A \cdot A)) \cdot {\bf 1}_n)$ \\
			+32 & ${\bf 1}_n' \cdot (((A \circ A) \cdot {\bf 1}_n) \circ ((A \circ A \circ A \circ A) \cdot ((A \circ A) \cdot {\bf 1}_n)))$ \\
			+192 & ${\bf 1}_n' \cdot ((A \circ A \circ A \circ ((A \circ A \circ (A \cdot A)) \cdot A)) \cdot {\bf 1}_n)$ \\
			+32 & ${\bf 1}_n' \cdot ((A \circ A \circ A \circ ((A \cdot A) \cdot (A \circ A \circ A))) \cdot {\bf 1}_n)$ \\
			+4 & ${\bf 1}_n' \cdot ((A \circ A \circ A \circ ((A \cdot (A \circ A \circ A)) \cdot A)) \cdot {\bf 1}_n)$ \\
			+64 & ${\bf 1}_n' \cdot (((A \circ A) \cdot {\bf 1}_n) \circ ((A \circ A \circ ((A \circ A) \cdot (A \circ A))) \cdot {\bf 1}_n))$ \\
			+5 & ${\bf 1}_n' \cdot ((A \circ A \circ ((A \circ A) \cdot ((A \circ A) \cdot (A \circ A)))) \cdot {\bf 1}_n)$ \\
			+22 & $\Sigma_{i_1 i_2 i_3 i_4} (A \circ A)_{i_1 i_2} A_{i_1 i_3} A_{i_1 i_4} A_{i_2 i_3} A_{i_2 i_4} (A \circ A)_{i_3 i_4}$ $\qquad\boldsymbol{\star}\boldsymbol{\star}$ \\
			-16 & ${\bf 1}_n' \cdot (((A \circ A \circ A \circ A) \cdot {\bf 1}_n) \circ ((A \circ ((A \cdot A) \cdot A)) \cdot {\bf 1}_n))$ \\
			-64 & ${\bf 1}_n' \cdot (((A \circ (A \cdot A)) \cdot {\bf 1}_n) \circ ((A \circ A \circ A \circ (A \cdot A)) \cdot {\bf 1}_n))$ \\
			-8 & ${\bf 1}_n' \cdot ((A \circ A \circ A \circ (A \cdot \mathrm{d}((A \circ (A \cdot A)) \cdot {\bf 1}_n) \cdot A)) \cdot {\bf 1}_n)$ \\
			-64 & ${\bf 1}_n' \cdot ((A \circ A \circ A \circ ((A \cdot A) \cdot A)) \cdot ((A \circ A) \cdot {\bf 1}_n))$ \\
			-16 & ${\bf 1}_n' \cdot ((A \circ A \circ A \circ ((A \cdot A) \cdot \mathrm{d}((A \circ A) \cdot {\bf 1}_n) \cdot A)) \cdot {\bf 1}_n)$ \\
			+1 & ${\bf 1}_n' \cdot ((A \circ ((((((A \cdot A) \cdot A) \cdot A) \cdot A) \cdot A) \cdot A)) \cdot {\bf 1}_n)$ \\
			-16 & ${\bf 1}_n' \cdot ((((A \circ A) \cdot ((A \circ A) \cdot {\bf 1}_n)) \circ ((A \circ A) \cdot {\bf 1}_n)) \circ ((A \circ A) \cdot {\bf 1}_n))$ \\
		\end{tabular}
		\hspace{-4em} 
		\begin{tabular}{>{\raggedleft\arraybackslash} p{0.06\textwidth}p{0.6\textwidth}}
			-96 & ${\bf 1}_n' \cdot (((A \circ A) \cdot {\bf 1}_n) \circ ((A \circ A \circ (A \cdot A) \circ (A \cdot A)) \cdot {\bf 1}_n))$ \\
			-4 & ${\bf 1}_n' \cdot (((A \circ A) \cdot ((A \circ A) \cdot ((A \circ A) \cdot {\bf 1}_n))) \circ ((A \circ A) \cdot {\bf 1}_n))$ \\
			-24 & ${\bf 1}_n' \cdot ((A \circ A \circ (A \cdot \mathrm{d}((A \circ A) \cdot {\bf 1}_n) \cdot A) \circ (A \cdot A)) \cdot {\bf 1}_n)$ \\
			-32 & ${\bf 1}_n' \cdot ((A \circ A \circ ((A \circ ((A \cdot A) \cdot A)) \cdot (A \circ A))) \cdot {\bf 1}_n)$ \\
			-64 & ${\bf 1}_n' \cdot ((A \circ A \circ ((A \circ (A \cdot A)) \cdot (A \circ (A \cdot A)))) \cdot {\bf 1}_n)$ \\
			-16 & ${\bf 1}_n' \cdot ((A \circ A \circ (((A \cdot A) \circ (A \cdot A)) \cdot (A \circ A))) \cdot {\bf 1}_n)$ \\
			+8 & ${\bf 1}_n' \cdot ((A \circ A \circ A \circ ((((A \cdot A) \cdot A) \cdot A) \cdot A)) \cdot {\bf 1}_n)$ \\
			+16 & ${\bf 1}_n' \cdot ((((A \circ A) \cdot {\bf 1}_n) \circ ((A \circ A) \cdot {\bf 1}_n)) \circ ((A \circ ((A \cdot A) \cdot A)) \cdot {\bf 1}_n))$ \\
			+8 & ${\bf 1}_n' \cdot (((A \circ A) \cdot ((A \circ A) \cdot {\bf 1}_n)) \circ ((A \circ ((A \cdot A) \cdot A)) \cdot {\bf 1}_n))$ \\
			+16 & ${\bf 1}_n' \cdot ((((A \circ A) \cdot {\bf 1}_n) \circ ((A \circ (A \cdot A)) \cdot {\bf 1}_n)) \circ ((A \circ (A \cdot A)) \cdot {\bf 1}_n))$ \\
			+24 & ${\bf 1}_n' \cdot ((A \circ A \circ (((A \cdot A) \cdot A) \cdot A) \circ (A \cdot A)) \cdot {\bf 1}_n)$ \\
			+12 & ${\bf 1}_n' \cdot ((A \circ A \circ ((A \cdot A) \cdot A) \circ (A \cdot (A \cdot A))) \cdot {\bf 1}_n)$ \\
			+8 & ${\bf 1}_n' \cdot (((A \circ A) \cdot {\bf 1}_n) \circ ((A \circ ((A \cdot A) \cdot A)) \cdot ((A \circ A) \cdot {\bf 1}_n)))$ \\
			+16 & ${\bf 1}_n' \cdot (((A \circ (A \cdot A)) \cdot ((A \circ A) \cdot {\bf 1}_n)) \circ ((A \circ (A \cdot A)) \cdot {\bf 1}_n))$ \\
			+4 & ${\bf 1}_n' \cdot (((A \circ A) \cdot {\bf 1}_n) \circ ((A \circ ((A \cdot A) \cdot \mathrm{d}((A \circ A) \cdot {\bf 1}_n) \cdot A)) \cdot {\bf 1}_n))$ \\
			+4 & ${\bf 1}_n' \cdot (((A \circ A) \cdot ((A \circ (A \cdot A)) \cdot {\bf 1}_n)) \circ ((A \circ (A \cdot A)) \cdot {\bf 1}_n))$ \\
			+24 & ${\bf 1}_n' \cdot ((A \circ ((A \cdot A) \cdot A) \circ (A \cdot A) \circ (A \cdot A)) \cdot {\bf 1}_n)$ \\
			+2 & ${\bf 1}_n' \cdot ((A \circ (((A \cdot A) \circ (A \cdot A) \circ (A \cdot A)) \cdot A)) \cdot {\bf 1}_n)$ \\
			-8 & ${\bf 1}_n' \cdot (((A \circ A) \cdot {\bf 1}_n) \circ ((A \circ ((((A \cdot A) \cdot A) \cdot A) \cdot A)) \cdot {\bf 1}_n))$ \\
			-8 & ${\bf 1}_n' \cdot (((A \circ (((A \cdot A) \cdot A) \cdot A)) \cdot {\bf 1}_n) \circ ((A \circ (A \cdot A)) \cdot {\bf 1}_n))$ \\
			-4 & ${\bf 1}_n' \cdot (((A \circ ((A \cdot A) \cdot A)) \cdot {\bf 1}_n) \circ ((A \circ ((A \cdot A) \cdot A)) \cdot {\bf 1}_n))$ \\
			-12 & ${\bf 1}_n' \cdot (((((A \circ A) \cdot {\bf 1}_n) \circ ((A \circ A) \cdot {\bf 1}_n)) \circ ((A \circ A) \cdot {\bf 1}_n)) \circ ((A \circ A) \cdot {\bf 1}_n))$ \\
		\end{tabular}
	}
\end{table}

From these examples, $C_4$ can be computed with a cost at most $O(n^3)$, and 
$C_8$ can be computed with a cost at most $O(n^4)$.  In both cases, the complexity is much lower than $O(n^m)$---the complexity of the brute-forth enumeration.

To formalize the CEEF problem, we introduce a few definitions.

%%%%%%%%%%%%%
%%%%%%%%%%%%%
%%%%%%%%%%%%%
\begin{definition}[SEA term] \label{def:SEA}
We call a term a {\it Succinct Expressive Algebraic (SEA)} term if it is an elementary function 
of $A$ and ${\bf 1}_n$ constructed using only three operations: (i) multiplication (applied to two matrices or a matrix and a vector),  (ii) entry-wise product (applied to two matrices or two vectors), and (iii) diagonal extraction (applied to a matrix).  
\end{definition}

In Examples~\ref{example1}-\ref{example2}, all terms in $C_4$ are SEA terms; and in the formula of $C_8$,  43 out of the 44 terms are SEA terms. 
	
\begin{definition}[FS term] \label{def:FS}
We call a term a $K$-layer {\it Full Sum (FS)} term if it has the form of 
\[
\sum_{1\leq i_1, i_2, \ldots, i_K\leq n} X^{(1)}_{j_1, k_1} X^{(2)}_{j_2, k_2} \ldots  X^{(N)}_{j_N, k_N},    \qquad K  \geq 1, N \geq 1,  
\] 
where each $X^{(s)}$ is an $n\times n$ matrix that is an elementary function of $A$ constructed using only matrix multiplication, entry-wise product, and diagonal extraction, and   $j_1,  k_1, \ldots, j_N, k_N$ take values in $\{i_1, i_2,\ldots, i_K\}$ with $j_{\ell} \neq k_{\ell}$, 
$1 \leq \ell \leq N$.  
\end{definition} 
	
%%%%%%%%%%%%
%%%%%%%%%%%%
\begin{definition}[IFS term] \label{def:IFS}
We say a $K$-layer FS term is an  {\it Incompressible Full Sum (IFS) term} if it cannot be written equivalently as 
an $L$-layer FS term for any $L < K$.  
\end{definition}

In the example of $C_8$, the term in \eqref{IFSterm} is a 4-layer FS term, where $N=6$, $X^{(s)}=A$ for $s\in \{2,3,4,5\}$, $X^{(s)}=A\circ A$ for $s\in \{1,6\}$, and $(j_1, k_1, \ldots, j_6, k_6)=(i_1, i_2, i_1, i_3, i_1, i_4, i_2, i_3, i_2, i_4, i_3, i_4)$. 
It can be shown that this term cannot be written as an $L$-layer FS term with $L<4$, so it is an IFS term. 

Any SEA term can be computed with a cost of $O(n^3)$. Any $K$-layer FS term, when computed directly using for-loops,  has a cost of $O(n^K)$.  Therefore,  for $K \geq 4$,  it is desirable to convert a $K$-layer FS term into an SEA term or an IFS term, whichever possible.  
If such an FS term can be converted into an SEA term, the cost is reduced from $O(n^K)$ to $O(n^3)$. Alternatively, if a $K$-layer FS term can be converted into a $K'$-layer IFS term with $K' < K$, the cost is reduced from $O(n^K)$ to $O(n^{K'})$.  Moreover, by definition, an IFS term cannot be compressed into fewer layers, suggesting that its computational cost is difficult to reduce further. The discussion here motivates the CEEF problem we formulate as follows. 

%%%%%%%%%%%%
%%%%%%%%%%%%
\begin{definition}[CEEF] \label{def:CEEF}
For each $m \geq 3$,  a {\it Computationally Efficient Equivalent Form (CEEF)}  of $C_m$ refers to a formula that decomposes $C_m$ into a linear 
combination of finitely many SEA terms and IFS terms,  with explicit linear coefficients
\end{definition}

To solve the CEEF problem,  we seek an algorithm that takes an integer $m \geq 3$ as input and automatically outputs a formula of $C_m$ satisfying Definition~\ref{def:CEEF}. 
An example of the output of our algorithm with $m=8$ is given in Table \ref{tb:C8}. 

\smallskip	
	
\begin{remark}[CEEF is more than a cycle-counting algorithm] \label{rmk:value-of-CEEF}
Solving CEEF is fundamentally different from developing algorithms for counting fixed-length cycles, such as DFS/backtracking, Johnson's algorithm, BFS, color coding, and dynamic programming (e.g., see \cite{ribeiro2021survey} for a review). These algorithms take a given data matrix $A$ as input and compute the numerical value of $C_m(A)$. However, they do not derive an algebraically equivalent expression for $C_m(A)$ that is valid for arbitrary $A$. 
%Consequently, the value of a CEEF extends beyond computational efficiency: it yields a general closed-form representation of the cycle count statistic.
Moreover, some algorithms (e.g., color coding) provide only approximate values of $C_m(A)$. Even exact algorithms are primarily designed to reduce the computational cost, without establishing that the resulting complexity is optimal. In contrast, the CEEF framework explicitly requires every retained term to be computationally irreducible,  meaning that its computational complexity cannot be reduced further while preserving algebraic equivalence.
In summary, the value of CEEF extends beyond efficient computation: it provides a general closed-form representation of the cycle count statistic, which is also useful for analyzing the theoretical properties of $C_m(A)$, assessing the computational efficiency of existing and future cycle-counting algorithms, and designing new approximate computation formulas that have both high accuracy and low complexity (see Remark~\ref{rmk:CC-proxy}). 

%Solving CEEF is different from designing algorithms for counting fixed-length cycles \cite{alon2008biomolecular,karimi2012message}, such as the DFS/Backtracking algorithm, Johnson's algorithm, BFS algorithm, color coding, and dynamic programming. These algorithms can be applied to a given data matrix $A$, but they do not provide an algebraically equivalent form of $C_m(A)$ that is valid for arbitrary $A$. Therefore, the value of the CEEF formulas is beyond computation itself. 
%Furthermore, some of these algorithms (e.g., color coding) only output an approximation to $C_m(A)$, and most of these algorithms do not guarantee that the complexity is the lowest possible. In contrast, the definitions of SEA terms and IFS terms in CEEF require that the complexity cannot be further reduced. 
\end{remark}

\subsection{Literature review, our new combinatorial framework, and the human-AI collaboration workflow} \label{subsec:humAI-intro} 
In our setting, $A$ is an arbitrary $n \times n$ symmetric matrix, including the binary case as a special case. 
When $m$ is relatively small,    the CEEF problem can be {\it solved by hand}.  For example,  \cite{harary1969graph} solved the binary case for $m = 4, 5$, \cite{Chang2003} solved the binary case for $m=6$,  and \cite{Movarraei2016} solved the binary case for $m = 7$. The CEEF problem for general, non-binary matrices $A$ is more difficult and has received less attention; in this setting, \cite{Jin2021optimal} solved $m = 3, 4$.

Unfortunately,  this problem becomes much more challenging for larger $m$. 
%For instance, when $m = 12$,  our study shows that $C_m$ is a linear combination of 1900 terms.  
Deriving formulas for large $m$ requires  delicate and extremely tedious calculations, where mistakes are easy to make. Indeed, 
\cite{harary1969graph} contained an error for  $m = 7$, as later pointed out by \cite{Russian}. 
Consequently,  deriving CEEF formulas {\it by hand} quickly becomes infeasible as $m$ increases.  
%and we must develop delicate theory that use complicated combinatorics and graph theory.  
In a remarkable paper,  \cite{Russian} studied the case of binary $A$ and derived  
formulas for $C_m$ for $3 \leq m \leq 13$.  However, their work focused exclusively on 
the binary case and the range $3 \leq m \leq 13$. Moreover, because some important proof details were omitted, 
it remains unclear how to generalize their approach 
to the more general setting where $A$ is not binary and $m$ is arbitrary.   
For these reasons,  the CEEF problem is both technically difficult and  extremely tedious for humans, and therefore remains largely unsolved. 
	
We may tackle the CEEF problem with AI. 
However, as mentioned earlier,   if we directly ask AI to generate the formula for a given $m$, the output is usually far from correct.  
%One possible solution is 
We have also used multi-shot learning (e.g., \cite{Tom2020,Wei2022}), 
in which we provide AI with the formulas for some small $m$ and ask it to reason for the formulas for larger $m$. Unfortunately, because the formulas become rapidly more complicated as $m$ increases,  this approach also fails to produce satisfactory results.  

Our proposal is a novel {\it human-guided AI (hugAI)} approach: we first derive a  symbolic representation of $C_m$ using several theorems. Next, to  figure out a more explicit form for the formula, we design a clear derivation strategy, in the form of an ``algorithm," and provide step-by-step guidance to AI;  each step is then carried out by combining the strengths of humans and AI. 
The key to the success of our approach lies in ensuring two properties: First, the derivation strategy must be {\it theoretically rigorous}, meaning that for any given $m$, if a human or an AI system correctly understands and executes the strategy, then the resulting output is guaranteed to be  the desired formula of $C_m$. Second, the derivation strategy must be sufficiently {\it transparent and structured} for an AI system to understand and follow effectively.

Specifically, our algorithm has two steps. 
{\it (1) Merging step}: We fix a set of $m$ distinct indices $S=\{i_1, i_2, \ldots, i_m\}$ and let ${\cal G}$ be the graph with edges only between $i_1 \& i_2, i_2 \& i_3, \ldots,  i_m \& i_1$.   
The algorithm identifies all multi-graphs induced by ${\cal G}$ through node merging, and partitions these multi-graphs into equivalence classes, only retaining one representative in each class. 
%This produces a list of multi-graphs denoted by ${\cal L}$.  
Using the M\"obius function \cite{Stanley2011},  we show that $C_m$ is a linear combination of FS terms (Definition~\ref{def:FS}), each corresponding to a multi-graph, where all linear coefficients are explicitly computable. 
{\it (2) Pruning step}: We relate each FS term to a Fully Annotated Multi-graph (FAM), where each layer in the sum corresponds to  a node in FAM. We propose a recursive pruning procedure where 
`pruning a node' is equivalent to `reducing one summation layer without changing the value of the sum'. 
%This procedure always terminates in finite steps.  
We show that it converts each FS term equivalently to either an SEA term or an IFS term, thus providing a CEEF formula.   

This algorithm satisfies the two  properties we seek. First, it is theoretically rigorous, since we prove that its output yields a symbolic expression of $C_m$ as the sum of SEA terms and IFS terms.
Second, it is easy to follow by AI systems, because the algorithm is formulated entirely in the language of graph theory.  Since graph theory is a field with highly structured notation and abundant training data available to modern AI systems, communicating this algorithm with AI becomes effective.

We use AI to implement this algorithm through carefully designed prompts. While many AI tools are available, we focus primarily on DeepSeek-R1 (DS), which has been shown to be competitive in solving mathematical problems (e.g., \cite{DS-math}). For completeness and cross validation, we also investigate several other AI systems using the same prompts; see Table~\ref{tab:results}.

Guided by our prompts, DS is able to successfully understand and execute the proposed algorithm (see Section~\ref{sec:results} for details). 
%In the merging step, DS can (a) generate a script that correctly identifies all multi-graphs, (b) visualize each multi-graph, (c) determine graph isomorphism, and (d) suggest that the M\"obius function may be useful for deriving the linear coefficients. In the pruning step, DS can (a) generate a script to implement the pruning procedure---which would be difficult to realize without advanced coding skills---and (b) successfully convert each FS term into either an SEA term or an IFS term, as expected. See Section~\ref{sec:results}.
As the final output, DS produces Python code such that, for any input $m \geq 3$, the code generates an explicit formula that satisfies Definition~\ref{def:CEEF}. %The resulting formulas for $3 \leq m \leq 12$ are provided in the Appendix. 
This resolves the CEEF problem for general $m$ and arbitrary symmetric matrices $A$.

%In summary, we find that AI is unable to accomplish the task independently, but it is able to 
%do so if we provide  a clear strategy, a step by step guidance, 
%and carefully written prompts. Typically, AI is unable to output the results as expected in one shot:  
%we need to refine the prompt a few times before AI is able to output the correct 
%results. 
	
Our contribution is two-fold.     
First, we solve a long-standing open problem in combinatorics--the CEEF problem. 
Compared with existing works, we  use  novel graph theory which has not been 
used before, and  make a timely contribution by finding the formulas that are demanded in several research areas 
such as network analysis, estimating 
spiked eigenvalues, and matrix testing (see Section~\ref{sec:applications}).  
Second, we show that AI can be a powerful research assistant for humans in tackling 
difficult open math  problems. %There has been an increasing attention on using AI to tackle research-level math problems \cite{alfarano2024lyapunov}. 
For difficult research-level math problems, whether AI can be truly useful is largely unclear.  Our paper assures that AI can be useful with appropriate guidance and prompts, and we showcase how to use AI properly.

%%%%%%%%%%%%%%
%%%%%%%%%%%%%%
\subsection{Connection to the recent research of AI4Math}  \label{subsec:connection} 

Recently, AI for mathematics (AI4Math) has attracted increasing attention. Broadly speaking, existing research can be divided into two categories. The first category focuses on evaluating AI's ability to assist mathematical research, where the interaction between humans and AI is primarily through direct prompting. The target problems may either have known solutions (e.g., \cite{Trinh, romera2024funsearch}) or remain open (e.g., \cite{bubeck2025early, dobriban2025solving, alexeev2026short}).
The second category focuses on designing human--AI collaboration frameworks that go beyond direct prompting (e.g., \cite{lample2022htps, alfarano2024lyapunov}). This line of work typically considers mathematical problems that are difficult to describe precisely in a single prompt and lack well-established proof strategies. Our work belongs to this second category.

For the CEEF problem, direct prompting alone is insufficient. Instead, substantial human input is required to develop the high-level reasoning and derivation strategy, while AI is employed to solve the resulting subproblems independently. The key novelty of our approach is the design of a reasoning and derivation strategy that is mathematically rigorous, not directly implied by existing results in combinatorics, and readily interpretable by AI. Together, these properties enable an effective collaboration between humans and AI. We believe that this paradigm of human--AI collaboration has the potential to extend to many other research-level mathematical problems.

Recent advances in AI for Science (AI4Sci) have shifted the focus from improving the intrinsic capabilities of AI models to augmenting them with external tools for solving complex scientific problems~\cite{AI4SciSurvey,PINN,AlphaFold2,GNoME,AgenticSurvey,LLMAgentSurvey,Coscientist,ChemCrow,SWEbench}. Our work follows this emerging paradigm. Rather than expecting AI to solve the CEEF problem independently, we adopt a \emph{hugAI} approach in which humans develop the reasoning and derivation strategy and AI serves as a research assistant, executing well-defined subproblems upon carefully designed prompts.

{\bf Content and notations:}
Section \ref{sec:main} presents the human-side contribution of this work, in which we present our theorems and algorithm.  
Section \ref{sec:results} presents the AI-side contribution of this work, in which we describe how to use AI to implement our  algorithm; a numerical validation pipeline is also established to verify the correctness of the resulting CEEF formulas. Section~\ref{sec:applications} contains numerical experiments in three applications of the cycle count statistic, demonstrating the benefit of using higher-order $C_m$ and thus the importance of solving CEEF. 
Section~\ref{sec:Discu} contains a short discussion.  
All proofs can be found in the Appendix.
Throughout this paper, $\circ$ denotes the Hadamard (or entry-wise) product.  For $n > 1$,  $I_n\in\mathbb{R}^{n\times n}$ denotes the identity matrix and ${\bf 1}_n \in \mathbb R^{n}$ denotes the vector of ones. 
For any vector $v \in \mathbb{R}^d$, $d(v)$ or $\mathrm{diag}(v)$ represents the $n \times n$ diagonal matrix whose $i$-th diagonal entry is $v_i$; 
and for any matrix $M\in\mathbb{R}^{n\times n}$,   $d(M)$ or $\mathrm{diag}(M)$ denotes the diagonal matrix constructed from the diagonal entries of $M$.

\section{The Algorithm and Its Theoretical Justification} \label{sec:main}
This section contains the human-part contribution of the paper, including main theorems and algorithms.  
Discussion on the AI-part contribution is deferred to Section \ref{sec:results}. 
Our goal here is to develop an ``algorithm" that derives the CEEF formula of $C_m$ for any given $m$.
We begin in Section~\ref{subsec:FAM} by introducing a new concept in graph theory, FAM, which plays a key role in our results. The proposed algorithm consists of two steps, a merging step and a pruning step, described in Section~\ref{subsec:merging} and Section~\ref{subsec:pruning}, respectively. Both steps are formulated in the language of graph theory and are justified rigorously.

\subsection{The FAM and the associated full sum}  \label{subsec:FAM} 
\begin{figure}[htb!]
\centering
\includegraphics[width=0.9 \textwidth]{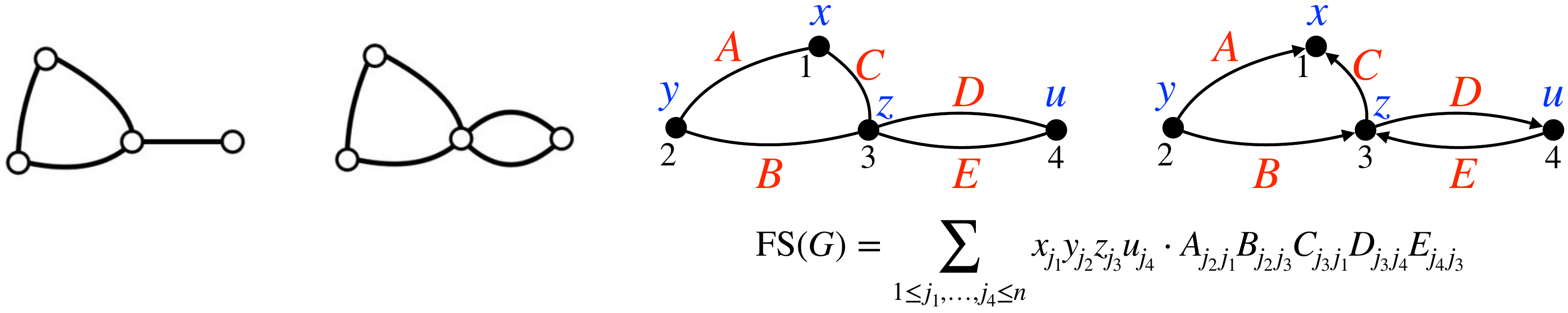}
\caption{A simple graph (left), a multi-graph (middle left), an undirected FAM (middle right), and a directed FAM (right). In the two FAMs, nodes are labeled by $x, y, z, u\in\mathbb{R}^n$, and edges are labeled by $A, B, C, D, E\in\mathbb{R}^{n\times n}$. The corresponding Full Sums (FSs) are displayed below these graphs. For an undirected FAM, the edge-label matrices are symmetric, so the value of the FS remains the same if we replace $A_{j_2j_1}$ by $A_{j_1j_2}$ (in comparison,  for a directed FAM, the value of the FS may change if we replace $A_{j_2 j_1}$ by $A_{j_1 j_2}$).}
\label{fig:fam1} 
\end{figure}

In graph theory \cite{harary1969graph,bondy1991graph},  a simple graph is a collection of nodes and edges such that,  
between any pair of distinct nodes, there is either no edge or exactly one undirected edge,  and  self-loops (i.e., edges connecting a node to itself)  are not allowed.  A multi-graph relaxes these restrictions by allowing multiple parallel edges between two nodes, as well as self-loops. In Figure~\ref{fig:fam1}, the left two panels illustrate the simple graph and multi-graph, respectively.  

Throughout this paper, simple graphs are assumed to be undirected, whereas multigraphs may be either undirected or directed. For a directed multigraph, each edge from node $a$ to node $b$ is represented by a triplet $(a,b,s)$, where $1 \le s \le w(a,b)$ and $w(a,b)$ denotes the number of edges from $a$ to $b$. For an undirected multigraph, edges between nodes $a$ and $b$ are likewise represented by triplets $(a,b,s)$ with $1 \le s \le w(a,b)$. In this case, the ordering of $a$ and $b$ does not matter, and $w(a,b)=w(b,a)$ always holds. We write $E(\mathcal{G})$ for the set of all triplets $(a,b,s)$ such that $w(a,b)\ge 1$.

We propose the {\it Fully Annotated Multi-graph (FAM)}, as well as the Full Sum and Distinct Sum associated with the FAM. 
These concepts play a central role in our results. 
	
%%%%%%%%%%
%%%%%%%%%%
\begin{definition}[Fully Annotated Multi-graph (FAM)] \label{def:LMG}
A FAM $\mathcal{G}$ is a multi-graph whose nodes are labeled by vectors $v^{(a)}\in\mathbb{R}^n$ and whose edges $(a,b,s)$ are labeled by matrices $M^{(a,b,s)}\in\mathbb{R}^{n\times n}$. 
For an undirected FAM, all edge-label matrices are required to be symmetric.
\end{definition}
%%%%%%%%%%%%%
%%%%%%%%%%%%%
%%%%%%%%%%%%%
\begin{definition}[FS and DS associated with a FAM] \label{def:LMG-FS}
Given an FAM $\mathcal{G}$, the Full-Sum (FS) associated with $\mathcal{G}$ is defined as
\[
\mathrm{FS}(\mathcal{G}):=\sum_{1\leq j_1,\ldots, j_k\leq n}\left( \prod_{1\leq a\leq k}v^{(a)}_{j_a}\prod_{(a,b,s)\in E(\mathcal{G})} M_{j_aj_b}^{(a,b,s)}\right),
\]
where $v_j$ denotes the $j$th entry of a vector $v\in\mathbb{R}^n$ and $M_{jj'}$ denotes the $(j, j')$th entry of a matrix $M\in\mathbb{R}^{n\times n}$. 
The Distinct-Sum (DS) associated with $\mathcal{G}$, $\mathrm{DS}(\mathcal{G})$, is defined similarly, except that $j_1, \ldots, j_k$ are required to be distinct in the sum. 
\end{definition} 
		
In Figure \ref{fig:fam1}, the right two panels illustrate an undirected FAM and a directed FAM, respectively. Both contain four nodes and six edges. Their associated FS expressions are nearly identical; the key difference is that, for an undirected FAM, terms such as $A_{j_2j_1}$ and $B_{j_2j_3}$ may be replaced by $A_{j_1j_2}$ and $B_{j_3j_2}$, respectively,  without changing the value of the FS, because the edge-label matrices are symmetric.  In contrast, for directed FAMs, such substitutions may change the value of the FS.   

Note that $\mathrm{FS}(\mathcal{G})$ can be viewed as a special case of the FS expression in Definition~\ref{def:FS}, in which certain matrices $X^{(s)}$ are restricted to be diagonal and the associated indices satisfy $j_s=k_s$.

The above definitions connect to our problem through a particular FAM constructed as follows. Fix $m \ge 3$ and a set of $m$ distinct indices $S=\{i_1,i_2,\ldots,i_m\}$, $1 \leq i_1, i_2, \ldots, i_m \leq n$.  Let $\mathcal{G}=\mathcal{G}(S)$ denote the simple graph where 
each index in $S$ is a node and we have a total of $m$ 
(undirected)  edges between $i_1 \& i_2$, $i_2 \&  i_3$, $\ldots$, $i_{m-1} \& i_m$, and $i_m  \& i_1$, respectively,  but nowhere else. We call $\mathcal{G}$  the root graph associated with $S$.

%%%%%%%%%%%%%%%%%%%%
\begin{definition}[The root-FAM] Fix $m \geq 3$ and $S = \{i_1, i_2, \ldots, i_m\}$, where $i_1, i_2, \ldots,  i_m$ are distinct indices in $\{1, 2, \ldots, n\}$.  Let $\mathcal{G}$ be the root graph as described above. We define the root FAM associated with $S$ to be the FAM obtained by labeling each node of $\mathcal{G}$ with $\mathbf{1}_n$ and each edge with the matrix $A$ (the original data matrix). By construction, every root FAM is undirected.

\end{definition}  
%%%%%%%%%%%%%%%%%%%%%%%%
We observe that the FS and DS associated the root-FAM are 
\begin{equation} \label{FS-DS-rootFAM}
	\mathrm{FS}({\cal G}) = \sum_{j_1,  \ldots, j_m} A_{j_1 j_2} A_{j_2 j_3} \ldots A_{j_m j_1} = \mathrm{trace}(A^m),\qquad \mathrm{DS}(\mathcal{G}) =  \sum_{j_1, \ldots, j_m (dist)} A_{j_1 j_2} A_{j_2 j_3} \ldots A_{j_m j_1} = C_m.
\end{equation}
This illustrates how these concepts are intrinsically connected to our problem. Note that although ${\cal G}$ depends on $S$,  the quantities $\mathrm{FS}({\cal G})$ and $\mathrm{DS}(\mathcal{G})$ 
do not depend on $S$.

\subsection{The merging process} \label{subsec:merging}  
	
In the merging process, we aim to express $C_m$ as  a linear combination of finitely many FS terms. In detail, 	let  $m \geq 3$ and $S = \{i_1, i_2, \ldots, i_m\}$, where $i_1, i_2, \ldots, i_m$ are distinct. 
Recall that ${\cal G} = {\cal G}(S)$ denotes the root-FAM associated with $S$.   
%%%%%%%%%%%%%%%%%%%%%%%%%%%
\begin{definition}[Partition] 
		Fix $k \leq m$. We call $\sigma = \{S_1, S_2, \ldots, S_k\}$ a $k$-partition of $S$ if $S_1, S_2, \ldots, S_k$ are pairwise disjoint non-empty subsets of $S$ satisfying $\cup_{j = 1}^k S_j = S$. We call $k$ the size of the partition (denoted by $|\sigma|$) and each $S_j$ a block. The set of all partitions for $S$ are denoted by $\Pi(S)$.   
\end{definition} 

Given each $\sigma\in \Pi(S)$, we can create a new FAM from the root FAM, by node merging 
according to the partition $\sigma$. 
%%%%%%%%%%%%%%%%%%%%%%%%%%
\begin{definition}[The induced FAM by a partition] 
Fix $1\leq k \leq m$ and a $k$-partition $\sigma = \{S_1, S_2, \ldots, S_k\}$. 
The induced FAM associated with $\sigma$  is the undirected FAM with node set $\{j_1, j_2,\ldots, j_k\}$. Its edges are induced by those of the root FAM as follows:  an edge $(a,b)$ in the root FAM induces an edge $(j_\ell, j_t)$ whenever $a\in S_\ell$ and $b\in S_t$. Consequently, the induced FAM may contain parallel edges and self-loops. Each node is still labeled by ${\bf 1}_n$, and each edge is still labeled by the data matrix $A$.   
We call $k = |\sigma|$ the size of the induced FAM. For $1 \leq \ell \leq k$, the quantity $|S_{\ell}|$ is called the multiplicity of node $j_{\ell}$. %By construction, the induced FAM is also undirected. 
\end{definition} 

By this definition, for each $1 \leq k \leq m$ and each $k$-partition, we obtain a size-$k$ induced FAM (when $k=m$,  it is the root FAM itself). We call such a process the {\it merging process}. 
%Note that during merging, the total number of edges remains unchanged, which is always equal to $m$.

\begin{example} 
Suppose $m=4$ and $S=\{i_1,i_2,i_3,i_4\}$. We consider a partition $\sigma=\{\{i_1\},\{i_2,i_4\},\{i_3\}\}$. The induced FAM has three nodes $\{j_1, j_2, j_3\}$, where $j_2$ is obtained from merging $i_2$ and $i_4$. 
After merging, there are two parallel edges between $j_1$ and $j_2$, and two
parallel edges between $j_2$ and $j_3$; all node labels are ${\bf 1}_n$ and
all edge labels are $A$. The associated FS term is
\[
\mathrm{FS}({\cal G}_{\sigma})=\sum_{j_1,j_2,j_3}(A_{j_1j_2})^2(A_{j_2j_3})^2 
=\sum_{j_1,j_2,j_3}(A\circ A)_{j_1j_2}(A\circ A)_{j_2j_3}
={\bf 1}_n'(A\circ A)^2{\bf 1}_n.
\]
There are a total of 9 partitions for $m=4$. However, many of the induced FAMs have self-loops. Since $A$ has a  zero diagonal, the associated FS term of such a FAM is always zero. 
Therefore, we restrict our attention to those induced FAMs that have no self-loop. 
Figure \ref{fig:merge-m4} illustrates two such induced FAMs. 
%The left most graph is the root FAM, while the two graphs in the middle are the two induced FAMs without self-loops.  
\end{example}
	
\begin{figure}[htbp]
\centering
\makebox[\textwidth][c]{\hspace*{1.5cm}%
\begin{tikzpicture}[x=1cm,y=1cm,scale=1.2,every node/.style={font=\footnotesize}]
			% Left: root FAM in the same style as Table \ref{tb:example-C4}
			\node at (-4.2,2.75) {root FAM};
			\node[dotnode] (t1) at (-5.2,1.2) {};
			\node[dotnode] (t2) at (-4.2,2.2) {};
			\node[dotnode] (t3) at (-3.2,1.2) {};
			\node[dotnode] (t4) at (-4.2,0.2) {};
			\draw[thick] (t1) -- (t2);
			\draw[thick] (t2) -- (t3);
			\draw[thick] (t3) -- (t4);
			\draw[thick] (t4) -- (t1);
			\node[text=black, scale=0.65] at (-4.82,1.80) {$A$};
			\node[text=black, scale=0.65] at (-3.58,1.80) {$A$};
			\node[text=black, scale=0.65] at (-3.58,0.60) {$A$};
			\node[text=black, scale=0.65] at (-4.82,0.60) {$A$};
			\node[left, scale=0.78] at (t1) {$i_1$};
			\node[right, scale=0.78] at (t2) {$i_2$};
			\node[right, scale=0.78] at (t3) {$i_3$};
			\node[left, scale=0.78] at (t4) {$i_4$};
			\node[align=left, scale=0.62, anchor=center] at (-4.2,-0.58)
			{$\mathrm{FS}=\sum_{j_1,j_2,j_3,j_4}
			A_{j_1j_2}A_{j_2j_3}A_{j_3j_4}A_{j_4j_1}$};
			
			% Right-top: size-3 induced FAM
			\node at (2.20,3.18) {size-3 induced FAM};
			\node[dotnode] (l1) at (0.95,2.05) {};
			\node[dotnode] (l2) at (2.55,2.80) {};
			\node[dotnode] (l3) at (2.55,1.30) {};
			\draw[thick] (l1) to[bend left=18] node[pos=.56, above=1pt, text=black, scale=0.62, inner sep=1pt] {$A$} (l2);
			\draw[thick] (l1) to[bend right=18] node[pos=.50, below=1pt, text=black, scale=0.62, inner sep=1pt] {$A$} (l2);
			\draw[thick] (l2) to[bend left=18] node[pos=.45, right=2pt, text=black, scale=0.62, inner sep=1pt] {$A$} (l3);
			\draw[thick] (l2) to[bend right=18] node[pos=.66, left=2pt, text=black, scale=0.62, inner sep=1pt] {$A$} (l3);
			\node[left, scale=0.78] at (l1) {$j_1$};
			\node[right, scale=0.78] at (l2) {$j_2$};
			\node[right, scale=0.78] at (l3) {$j_3$};
			\node[align=left, scale=0.78, anchor=west] at (3.45,2.05)
			{$\mathrm{FS}=\sum_{j_1,j_2,j_3}
			(A_{j_1j_2})^2(A_{j_2j_3})^2 $\\
			$={\bf 1}_n'(A\circ A)^2{\bf 1}_n$};
			
			% Right-bottom: size-2 induced FAM
			\node at (2.20,0.43) {size-2 induced FAM};
			\node[dotnode] (r1) at (0.95,-0.22) {};
			\node[dotnode] (r2) at (2.55,-0.22) {};
			\draw[thick] (r1) to[bend left=30] node[pos=.50, above=1pt, text=black, scale=0.58, inner sep=1pt] {$A$} (r2);
			\draw[thick] (r1) to[bend left=12] node[pos=.50, above=0pt, text=black, scale=0.58, inner sep=1pt] {$A$} (r2);
			\draw[thick] (r1) -- node[pos=.50, below=1pt, text=black, scale=0.58, inner sep=1pt] {$A$} (r2);
			\draw[thick] (r1) to[bend right=16] node[pos=.50, below=2pt, text=black, scale=0.58, inner sep=1pt] {$A$} (r2);
			\node[left, scale=0.78] at (r1) {$j_1$};
			\node[right, scale=0.78] at (r2) {$j_2$};
			\node[align=left, scale=0.78, anchor=west] at (3.45,-0.33)
			{$\mathrm{FS}=\sum_{j_1,j_2}
			(A_{j_1j_2})^4$\\
			$={\bf 1}_n'(A\circ A\circ A\circ A){\bf 1}_n$};
			
			% Merge arrows and partition labels
			\draw[-stealth, thick] (-3.15,1.55) -- (0.35,2.27);
			\node[align=center, scale=0.64] at (-1.35,2.42)
			{$\sigma_1=\{\{i_1\},\{i_2,i_4\},\{i_3\}\}$\\merge $i_2$ and $i_4$};
			
			\draw[-stealth, thick] (-3.15,0.85) -- (0.35,0.08);
			\node[align=center, scale=0.64] at (-1.35,-0.24)
			{$\sigma_2=\{\{i_1,i_3\},\{i_2,i_4\}\}$\\merge opposite nodes};
		\end{tikzpicture}}
		\caption{A partial illustration of the merging process for $m=4$. The process starts from the root FAM (a 4-cycle) and considers all partitions. Each partition yields an induced FAM. Only two induced FAMs that have no self-loops are shown here.}
		\label{fig:merge-m4}
	\end{figure}

\begin{definition} 
We say two induced FAMs are isomorphic if one can be obtained from the other by a permutation of the nodes.  
\end{definition}

The merging process identifies all induced FAMs for the root FAM. Some of these induced FAMs are isomorphic.
When two FAMs are isomorphic, it follows by Definition~\ref{def:LMG-FS} that their corresponding FS terms must be equal. We thus divide all these induced FAMs into different classes, first by size and then by isomorphism. In detail, we propose Algorithm~\ref{alg:merging}.

\begin{algorithm}[ht]
\caption{The merging process. } \label{alg:merging}
 Input: $S = \{i_1, i_2, \ldots, i_m\}$. 
\begin{enumerate}
\item[a.] List all partitions in $\Pi(S)$. For each partition, plot the induced FAM and record the size. 
\item[b.] Exclude all induced FAMs with self-loops (because any induced size-$1$ FAM must have self-loops and are excluded, the size of the retaining induced FAMs ranges from $2$ to $m$).  
\item[c.] Divide all remaining FAMs into $(m-1)$ groups, where the $k$-th group contains all FAMs with size $k$, for $2 \leq k \leq m$ (when $k = m$, group $k$ has only one member, which is the root-FAM ${\cal G}$). 
\item[d.] Fix $2 \leq k \leq m$. Partition the $k$-th group into equivalence classes such that 
two FAMs belong to the same class if and only if they are isomorphic. 
List all classes in some fixed order (e.g. lexicographical order), and let 
\begin{align} \label{def:b-and-d} 
b_{m, k}& =\mbox{the total number of classes in the $k$-th group}, \\
d_{m,k,t} &= \mbox{cardinality of class $t$ (i.e., the number of FAMs contained in that class)}, \quad 1 \leq  t \leq b_{m, k}. \nonumber
\end{align}
Now, all induced FAMs are partitioned into classes according first to their size  and then to graph isomorphism. Each class is indexed by a pair $(k, t)$, where $2 \leq k \leq m$ and $1 \leq t \leq b_{m, k}$. 
   
\item[e.]   For each class $(k, t)$, select a candidate FAM denoted by ${\cal G}_{m, k, t}$. 
Denote by $j_1, \ldots, j_k$ the nodes of ${\cal G}_{m,k,t}$ and by $g_i$ the degree of node $j_i$ (in this case, $g_i$ coincides with the number of original nodes grouped into $S_i$).  
Define $h_{m,k,t}$ by 
%This gives rise to the set of multi-graphs $\{{\cal G}_{m, k, t}: 1 \leq t \leq b_{m, k}, 2 \leq k \leq m\}$.
%When $k=m$, $h_{m,k,t}=1$; when $k<m$, let $g_1, g_2, \ldots, g_k$ denote the multiplicities  of all $k$ nodes of ${\cal G}_{m, k, t}$. Introduce 
\begin{equation} \label{def:h} 
h_{m,k,t} =   \Pi_{i = 1}^k (g_i - 1)! \quad \mbox{if }k<m, \qquad \mbox{and}\qquad h_{m,k,t}=1 \quad \mbox{if }k=m. 
\end{equation} 
%In the special case of $k = m$, the class contains only one FAM, which is the 
%we only have one FAM with size $k$ which is the root-FAM ${\cal G}$. In this case, $b_{m, k} = 1$, and for all $1 \leq t \leq b_{m, k}$ (where the only possible value for $t$ is $t = 1$),  
%\[
%d_{m, k, t} = 1,  \qquad h_{m, k, t} = 1, \qquad  {\cal G}_{m, k, t} = {\cal G}. 
%\] 
\end{enumerate} 
Output: $b_{m,k}$ for $2\leq k\leq m$, and $({\cal G}_{m,k,t}, d_{m,k,t}, h_{m,k,t})$ for $2\leq k\leq m$ and $1\leq t\leq b_{m,k}$.  
% FAM , the  $b_{m, k}$ and $d_{m, k, t}$,  $1 \leq t \leq b_{m, k}$ and $2 \leq k \leq m-2$. Output 
%\[
%\{: 1 \leq t \leq b_{m, k}, 2 \leq k \leq m\} 
%\] 
%and 
%\[
%\{\mathrm{FS}({\cal G}_{m, k,t}): 1 \leq t \leq b_{m, k}, 2 \leq k \leq m\}. 
%\] 
%These steps can be implemented with AI; see Section \ref{sec:results} for details. 
\end{algorithm}

We present our main theorem about the merging step.  
%Recall that $\mu$ is the M\"obius function and $\hat{0}$ is the finest partition.  
%%%%%%%%%%
%%%%%%%%%%
%%%%%%%%%%
\begin{theorem} \label{thm:Cm}
Fix $m \geq 3$. Let ${\cal G}_{m, k, t}$, $d_{m,k, t}$ and $h_{m, k, t}$ be as in Algorithm~\ref{alg:merging} for $2\leq k\leq m$ and $1\leq t\leq b_{m,k}$.  %and suppose  is the induced multi-graph from the partition 
Let $\sigma_{m, k, t}$ be the partition that induces the FAM ${\cal G}_{m, k, t}$. 
Define $\mu(\hat{0}, \sigma_{m, k, t}) = (-1)^{m - k}  \cdot  h_{m, k, t}$ and $a_{m, k, t} = d_{m, k, t}  \cdot  \mu(\hat{0}, \sigma_{m, k, t}) $. Then, 
\[
C_m = \sum_{k=2}^{m} \sum_{t=1}^{b_{m,k}} a_{m,k,t} \cdot \mathrm{FS}({\cal G}_{m, k, t}). 
\]  
\end{theorem} 

This theorem provides a rigorous computational formula of $C_m$: It expresses $C_m$ as a linear combination of finitely many FS terms, where all linear coefficients are explicitly computable. 
By Definition~\ref{def:LMG-FS}, $\mathrm{FS}({\cal G}_{m, k, t})$ involves $k$ layers of sum and can be computed with $O(n^k)$ cost. Since $k$ ranges from $2$ to $m$, this formula alone does not reduce the computational complexity compared to the brute-forth approach. To complete the story, we will develop a pruning process in Section~\ref{subsec:pruning}, which converts each $\mathrm{FS}({\cal G}_{m, k, t})$  equivalently to either an SEA term or an IFS term. 

\begin{remark}[Connection to the M\"obius function] \label{rmk:Mobius}
In Theorem~\ref{thm:Cm},  $\mu(\cdot, \sigma_{m, k, t})$ coincides with the M\"obius function on a partition lattice, and $\hat{0}$ refers to the finest partition. 
Initially, we were not aware that the linear coefficients in the decomposition of $C_m$ are related to the M\"obius function. However, when we guided the AI (specifically, DeepSeek-R1) to compute these coefficients, AI hinted us that the M\"obius inversion technique might be relevant. Indeed, our proof of Theorem~\ref{thm:Cm} involves deriving an explicit form of the M\"obius function. A more detailed discussion is contained in the Appendix. 
\end{remark}

\begin{remark}[Self-loops]
Algorithm~\ref{alg:merging} excludes induced FAMs that contain self-loops. Indeed, when a FAM has self-loops, each summand in the associated Full Sum (see Definition~\ref{def:LMG-FS}) involves diagonal entries of $A$, which are assumed to be zero in our setting (otherwise, we plug in $A-\mathrm{diag}(A)$). As a result, the Full Sum is zero and omitted in the computation circuit.   
\end{remark}

We note that $C_m$ is the Distinct-Sum associated with the root FAM ${\cal G}$; see \eqref{FS-DS-rootFAM}. In this sense, Theorem~\ref{thm:Cm} expresses a DS-term as a linear combination of FS-terms. One may wonder whether the opposite is true. The following corollary shows that $\tr(A^m)$, the Full-Sum associated with the root FAM, can be expressed as a linear combination of DS-terms on the induced FAMs.  
\begin{corollary} \label{cor:Cm}
For any $m \geq 3$,
\[
C_m = \tr(A^m) - \sum_{k=2}^{m-1} \sum_{t=1}^{b_{m,k}} d_{m,k,t}\cdot \mathrm{DS}({\cal G}_{m, k, t}). 
\]
\end{corollary} 
This provides an alternative decomposition of $C_m$ that may also of interest in theory and in practice.

\begin{example}  \label{example4}
When $m=4$, $b_{m, k} = 1$ for $k = 2, 3, 4$; i.e., each group has one equivalence class.  
Table \ref{tb:example-C4} shows the induced FAMs that represent the three classes. Suppose ${\cal G}_{m,k,t}$ has nodes $\{j_1, \ldots, j_k\}$. By Definition~\ref{def:LMG-FS}, there is a function $f_{m, k, t}(A, j_1, \ldots, j_k)$ such that $\mathrm{FS}({\cal G}_{m,k,t})=\sum_{j_1, \ldots, j_k}f_{m, k, t}(A, j_1, \ldots, j_k)$; see Table \ref{tb:example-C4}. 
For example, $f_{m, k, t}(A, j_1, \ldots, j_k)=(A_{j_1j_2})^2(A_{j_2j_3})^2$ for the second FAM. It follows that $\mathrm{FS}({\cal G}_{m,k,t})=\sum_{j_1, j_2, j_3} (A_{j_1j_2})^2(A_{j_2j_3})^2= {\bf 1}_n' (A\circ A)^2 {\bf 1}_n$, so this term is equivalent to an SEA term. A similar observation is made for the other two FAMs. This explains the CEEF formula in Example~\ref{example1}. 
\end{example}

		\begin{table}[htbp]
		\centering
		\small
		\caption{ The key quantities for $m = 4$, with Full-Annotated Multi-graphs (FAM) included for illustration. Combining the following table with Theorem \ref{thm:Cm}, we obtain  $C_4 = \tr(A^4) - 2[ {\bf 1}_n' (A \circ A)^2 {\bf 1}_n] + {\bf 1}_n' (A \circ A \circ A \circ A) {\bf 1}_n$.} \label{tb:example-C4}
		\scalebox{0.9}{ 
			\begin{tabular}{|c|c|c|c|c|c|c|c|c|}
				\hline
				$(k,t)$ & Partition & Multigraph &FAM& $(d_{m k t}, h_{mkt}, a_{mkt})$ & $f_{mkt}(A,j_1,\cdots,j_k)$ & 
				%$\sum_{j_1, \ldots, j_k}  f_{m, k, t}(A,j_1,\cdots,j_k)$ 
				SEA form \\
				\hline
				$(4,1)$ & \begin{tabular}{@{}c@{}}
					$\{\{i_1\},\{i_2\},\{i_3\},\{i_4\}\}$\\
					$j_k = i_k,\;1\le k\le4$\\
				\end{tabular} & 	\begin{tikzpicture}[graph layout={nodes=4}, scale=0.6]
					\newnode \newnode \newnode \newnode
					\draw[thick] (1) -- (2);
					\draw[thick] (2) -- (3);
					\draw[thick] (3) -- (4);
					\draw[thick] (4) -- (1);
					\node[above left, scale=0.6] at (1) {$j_1$};
					\node[above right, scale=0.6] at (2) {$j_2$};
					\node[below right, scale=0.6] at (3) {$j_3$};
					\node[below left, scale=0.6] at (4) {$j_4$};
				\end{tikzpicture}& 	\begin{tikzpicture}[graph layout={nodes=4}, scale=0.6]
					\newnode \newnode \newnode \newnode
					\draw[thick] (1) -- (2);
					\draw[thick] (2) -- (3);
					\draw[thick] (3) -- (4);
					\draw[thick] (4) -- (1);
					\node[above left, scale=0.6] at (1) {$j_1$};
					\node[above right, scale=0.6] at (2) {$j_2$};
					\node[below right, scale=0.6] at (3) {$j_3$};
					\node[below left, scale=0.6] at (4) {$j_4$};
					\node[text=black, scale=0.58] at (-0.56,0.58) {$A$};
					\node[text=black, scale=0.58] at (0.56,0.58) {$A$};
					\node[text=black, scale=0.58] at (0.56,-0.58) {$A$};
					\node[text=black, scale=0.58] at (-0.56,-0.58) {$A$};
				\end{tikzpicture} & $(1,1,1)$ &  $A_{j_1 j_2}A_{j_2 j_3}A_{j_3 j_4}A_{j_4 j_1}$ & $\tr(A^4)$ \\
				\hline
				$(3,1)$ &  \begin{tabular}{@{}c@{}}
					$\{\{i_1\},\{i_2,i_4\},\{i_3\}\}$\\
					$j_1=i_1,\;j_2=i_2,\;j_3=i_3$\\
				\end{tabular} & \begin{tikzpicture}[graph layout={nodes=3}, scale=0.6]
					\newnode \newnode \newnode
					\draw[thick] (1) -- (2) node[edge label, scale=0.7] {2};
					\draw[thick] (2) -- (3) node[edge label, scale=0.7] {2};
					\node[left, scale=0.6] at (1) {$j_1$};
					\node[right, scale=0.6] at (2) {$j_2$};
					\node[below, scale=0.6] at (3) {$j_3$};
					%					\node[above=6pt, scale=0.6] at (0,0.5) {Merge $j_2$-$j_4$};
				\end{tikzpicture} & \begin{tikzpicture}[graph layout={nodes=3}, scale=0.6]
					\coordinate (q1) at (-1.05,0);
					\coordinate (q2) at (0.62,0.72);
					\coordinate (q3) at (0.62,-0.72);
					\node[dotnode] at (q1) {};
					\node[dotnode] at (q2) {};
					\node[dotnode] at (q3) {};
					\draw[thick] (q1) to[bend left=15] node[pos=.55, above=1pt, text=black, scale=0.58, inner sep=1pt] {$A$} (q2);
					\draw[thick] (q1) to[bend right=15] node[pos=.50, below=1pt, text=black, scale=0.58, inner sep=1pt] {$A$} (q2);
					\draw[thick] (q2) to[bend left=15] node[pos=.50, right=2pt, text=black, scale=0.58, inner sep=1pt] {$A$} (q3);
					\draw[thick] (q2) to[bend right=15] node[pos=.50, left=2pt, text=black, scale=0.58, inner sep=1pt] {$A$} (q3);
					\node[left, scale=0.65] at (q1) {$j_1$};
					\node[right, scale=0.65] at (q2) {$j_2$};
					\node[right, scale=0.65] at (q3) {$j_3$};
				\end{tikzpicture}  & $(2, 1, -2)$  & $(A_{j_1j_2})^2 (A_{j_2j_3})^2$ & ${\bf 1}_n' (A \circ A)^2 {\bf 1}_n $ \\
				\hline
				$(2,1)$ & 	\begin{tabular}{@{}c@{}}
				\\
					$\{\{i_1,i_3\},\{i_2,i_4\}\}$\\
					$j_1=i_1,\;j_2=i_2$\\
				\end{tabular}& 	\begin{tikzpicture}[graph layout={nodes=2}, scale=0.6]
					\newnode \newnode
					\draw[thick] (1) -- (2) node[edge label, scale=0.7] {4};
					\node[left, scale=0.6] at (1) {$j_1$};
					\node[right, scale=0.6] at (2) {$j_2$};
					%					\node[above=6pt, scale=0.6] at (0,0.5) {Merge $j_1$-$j_3$, $j_2$-$j_4$};
				\end{tikzpicture}& \begin{tikzpicture}[x=.58cm,y=.58cm,baseline=-.3ex, transform canvas={yshift=5pt}] 
					\coordinate (p1) at (-1.08,0);
					\coordinate (p2) at (1.08,0);
					\node[dotnode] at (p1) {};
					\node[dotnode] at (p2) {};
					\draw[thick] (p1) to[bend left=34] node[pos=.36, above=1pt, text=black, scale=0.56, inner sep=1pt] {$A$} (p2);
					\draw[thick] (p1) to[bend left=13] node[pos=.66, above=0pt, text=black, scale=0.56, inner sep=1pt] {$A$} (p2);
					\draw[thick] (p1) to[bend right=13] node[pos=.34, below=0pt, text=black, scale=0.56, inner sep=1pt] {$A$} (p2);
					\draw[thick] (p1) to[bend right=34] node[pos=.64, below=1pt, text=black, scale=0.56, inner sep=1pt] {$A$} (p2);
					\node[left, scale=0.65] at (p1) {$j_1$};
					\node[right, scale=0.65] at (p2) {$j_2$};
				\end{tikzpicture}
				& $(1,1,1)$ & $(A_{j_1j_2})^4$ & ${\bf 1}_n^{\prime}(A \circ A \circ A \circ A){\bf 1}_n$ \\
				\hline
			\end{tabular}
		} 
		\smallskip
	\end{table}

Our results in this merging step are new, and the proofs are non-trivial. First,  by introducing two new notions (FAM and the associated Full Sum for the FAM) in graph theory,  we are able decompose $C_m$ as a linear combination of the Full Sum of many induced FAMs. This allows us to develop sophisticated graph theory and use the results  to analyze $C_m$. In particular, the two notions play an important role in the pruning step below. Second, to figure out the linear coefficients in Theorem \ref{thm:Cm}, we need to combine careful analysis with existing results on the M\"obius function (see Remark~\ref{rmk:Mobius}). 
%(at first, we do not know how to figure out these coefficients, but fortunately, AI hinted that the M\"oblus inversion technique may be relevant. We solve the problem by following the hint. This demonstrates how valuable AI can be for research).  

%%%%%%%%%%
%%%%%%%%%%
%%%%%%%%%%  
\subsection{The pruning process} 
	\label{subsec:pruning}

In the merging process, $C_m$ has been expressed as a linear combination of finitely many FS-terms, where all linear coefficients are explicitly computable. Our goal now is to convert each $\mathrm{FS}({\cal G}_{m, k, t})$  equivalently to either an SEA term or an IFS term. 
To the best of our knowledge, this is the first work to formally introduce the notions of SEA terms and IFS terms. In particular, the concept of SEA terms based on three fundamental matrix-vector operations has never appeared in previous studies (e.g., \cite{harary1969graph, Chang2003, Movarraei2016, Russian, Jin2021optimal}). Consequently, no existing methodology is available for this task; and the formulas previously derived for relatively small values of $m$ rely largely on case-by-case analysis rather than on a principled procedure. 
In contrast, we develop a general and principled procedure applicable to arbitrary values of $m$.

Fix an FAM ${\cal G}={\cal G}_{m, k, t}$. By Definition~\ref{def:LMG-FS}, there is a one-to-one 
correspondence between the number of nodes in this FAM and the number of summation layers in the associated Full Sum (the latter is directly connected to computational complexity). 
We propose a recursive pruning algorithm, in which ``pruning a node''
means reducing the number of summation layers by one, while preserving the value of the sum.  
%{\color{red}Throughout the pruning process, every edge label is read in its stored orientation as in Definition~\ref{def:LMG-FS}; the arbitrary initial choices do not affect the value of the FS because the initial labels are symmetric, but after pruning the generated matrix labels are generally not symmetric.} 
To achieve this goal, we develop a scheme to prune the {\it Type-I and Type-II pendants} defined as follows.   
%%%%%%%%%%%%%%
%%%%%%%%%%%%%%
%%%%%%%%%%%%%%
\begin{definition}[Pendants]
In an undirected multi-graph, a node is called a {\it Type-I pendant} node if it has exactly $1$ immediate neighbor (node $b$ is an immediate neighbor of $a$ if there exists at least one edge between them), and a node is called a {\it Type-II pendant} node if it has exactly $2$ immediate neighbors. In either case, each immediate neighbor of a {\it pendant} node is referred to as a {\it hinge} node. For a FAM, 
a node is called a Type-I (respectively, Type-II) pendant  if it is a Type-I (respectively, Type-II) pendant in the underlying undirected multi-graph obtained by removing all edge directions and all node/edge labels.  
\end{definition}
	
%%%%%%%%%%%%%%
%%%%%%%%%%%%%%
%%%%%%%%%%%%%%
\begin{figure}[htb!]
\centering
\includegraphics[width=.55\textwidth]{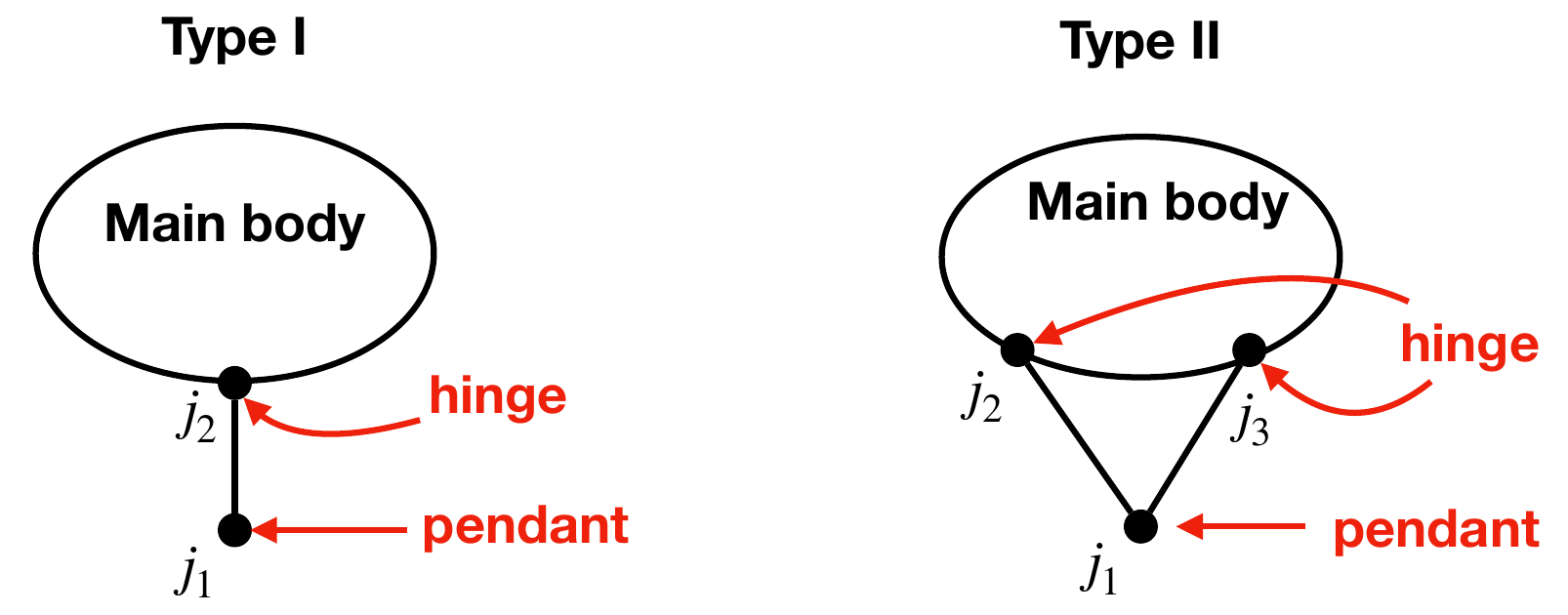}
\caption{Type I pendant (left) and Type II pendant (right). No matter whether the FAM is directed or undirected, pendants are defined based on the underlying undirected multi-graph.}
\label{fig:2types} 
\end{figure}

A FAM ${\cal G}$ obtained from the previous merging step is initially undirected. However, the pruning process requires ${\cal G}$ to be represented as a directed FAM. To this end, we arbitrarily assign an orientation to each edge in ${\cal G}$. Because the edge-label matrices of ${\cal G}$ are symmetric at this stage, the assigned orientations do not affect the corresponding full sum $\mathrm{FS}({\cal G})$. As pruning proceeds, pendant nodes are successively removed and the FAM is updated accordingly. During this process, some edge-label matrices may become asymmetric, making it necessary to keep track of the edge orientations.

\begin{definition}[Orientation alignment] \label{def:re-orientation}
Given a directed FAM ${\cal G}$, an orientation alignment from node $a$ to node $b$, denoted by $\mathrm{OA}_{a\to b}({\cal G})$, re-orients every edge between $a$ and $b$ so that it points from $a$ to $b$. Specifically, edges already directed from $a$ to $b$ are left unchanged, while edges directed from $b$ to $a$ are reversed and their corresponding edge-label matrices are replaced by their transposes.
\end{definition}

Combining Definition~\ref{def:re-orientation} with Definition~\ref{def:LMG-FS}, an orientation alignment does not alter the FS-term associated with a FAM. 

We now describe the pendant pruning procedure. For both Type-I and Type-II pendants, the procedure consists of two steps: first performing an orientation alignment, and then pruning the pendant node and updating the node labels and edge labels of the resulting FAM. The essential requirement is that each of these operations preserves the corresponding FS-term.

\begin{figure}[htb!]
\centering
\includegraphics[height=.18\textwidth]{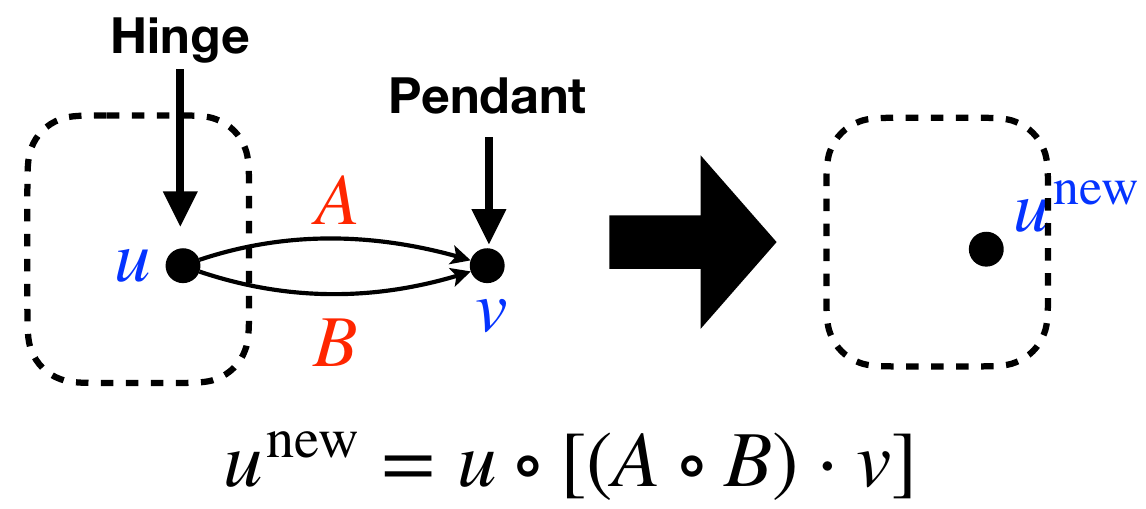}
\hspace{2 em} 
\includegraphics[height=.18\textwidth]{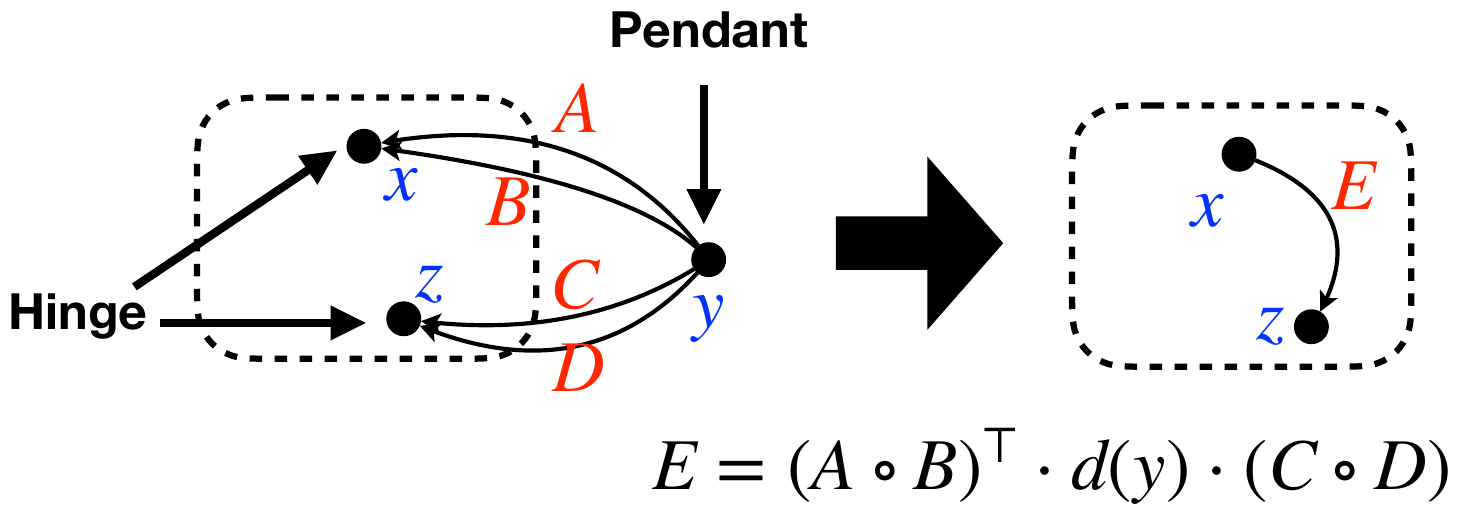}
\caption{The pruning process and corresponding updating rule. Left: Type I. Right: Type II.}
\label{fig:pendant}
\end{figure}

Consider a Type-I pendant node $j_1$, with $j_2$ as the hinge node. Suppose there are $s$ parallel edges between them. We conduct orientation alignment from the hinge node to the pendant node, i.e., $\mathrm{OA}_{j_2\to j_1}({\cal G})$. Let $M^{(1)},\ldots,M^{(s)}\in\mathbb{R}^{n\times n}$ be the edge-labels after orientation alignment. Let $v\in\mathbb{R}^n$ and $u\in\mathbb{R}^n$ be the respective node labels for $j_1$ and 
	$j_2$.  
	When we prune the pendant node, we remove $j_1$ and all edges between $j_1$ and $j_2$, 
	and update the node label of {\color{black}the hinge node $j_2$} by 
	\begin{equation} \label{Pruning1}
		u^{\text{new}} = u\circ \bigl[ (M^{(1)}\circ \ldots\circ M^{(s)})\cdot v\bigr],     
	\end{equation}
	where $\circ$ is the entry-wise product between two vectors or two matrices.   
	Aside from the above modifications,  all other nodes, edges, and their labels remain unchanged. An illustration of this updating rule is provided in the left panel of Figure~\ref{fig:pendant}. 
	
Consider a Type-II pendant node $j_1$, with $j_2$ and $j_3$ as the hinge nodes.  We perform $\mathrm{OA}_{j_1\to j_2}({\cal G})$ and $\mathrm{OA}_{j_1\to j_3}({\cal G})$, which means that all edges between the pendant and two hinges are re-oriented to point from the pendant to the hinges. Suppose there are $s$ parallel edges between $j_1$ and $j_2$, and their labels are $Q^{(1)}, \ldots, Q^{(s)}\in\mathbb{R}^{n\times n}$ after orientation alignment, and there are $t$ parallel edges between $j_1$ and $j_3$,  with labels $R^{(1)}, \ldots, R^{(t)}\in\mathbb{R}^{n\times n}$ after orientation alignment.
Let {\color{black}$y,x,z\in\mathbb{R}^n$} be the node labels for $j_1, j_2, j_3$, respectively.  
When we prune the pendant node, we remove $j_1$, all edges between $j_1$ and $j_2$, and all edges between 
	$j_1$ and $j_3$; and we add a new directed edge from $j_2$ to $j_3$ with the following label
	\begin{equation} \label{Pruning2}
		E = (Q^{(1)}\circ\ldots \circ Q^{(s)})^\top \cdot \mathrm{diag}(y)  \cdot (R^{(1)}\circ \ldots \circ R^{(t)}). 
	\end{equation}
Here, $\circ$ denotes entry-wise product, $\cdot$ denotes matrix multiplication, and $\mathrm{diag}(y)$ is a diagonal matrix whose diagonal entries are from $y$. 
All other nodes, edges, and their labels remain unchanged. An illustration is provided in the right panel of Figure~\ref{fig:pendant}.

%%%%%%%%%%%%%%%
%%%%%%%%%%%%%%%
%%%%%%%%%%%%%%%

\begin{lemma}[Invariance of the Full-Sum under pruning] \label{lemma:LMG} 
Consider a FAM ${\cal G}$ with at least $1$ pendant node which is either Type-I or Type-II.  Suppose we pick any pendant node and prune it according to the two rules in \eqref{Pruning1}-\eqref{Pruning2}, and let ${\cal G}^{\text{pruned}}$ be the FAM after pruning. Then, ${\cal G}^{\text{pruned}}$ has one fewer nodes than ${\cal G}$, while the associated Full-Sum remains unchanged: 
\[
\mathrm{FS}(\mathcal{G}) = \mathrm{FS}(\mathcal{G}^{\text{pruned}}).
\] 
\end{lemma} 
	
Lemma~\ref{lemma:LMG} justifies that pruning reduces the number of summation layers while keeping the associated Full-Sum unchanged. Furthermore, once the edge orientations have been aligned, the updating rules in \eqref{Pruning1}-\eqref{Pruning2} only use three operations: multiplication, entry-wise product, and diagonal extraction. 
These are exactly the operations allowed in SEA terms; see Definition~\ref{def:SEA}.   
	
	We now propose the following algorithm; see Algorithm~\ref{alg:pruning}. 
\begin{algorithm}[htbp]
\caption{The pruning process.} \label{alg:pruning}
Input: an induced FAM ${\cal G}_{m, k, t}$ from Algorithm~\ref{alg:merging}, with arbitrarily assigned edge orientations.  
	\begin{enumerate} 
		\item[a.] Check whether the FAM contains any pendant nodes. If not, terminate the algorithm. Otherwise, select an arbitrary pendant node and remove it together with all edges connecting it to its hinge node(s). 
		\item[b.] Update the node and edge labels using either \eqref{Pruning1} or \eqref{Pruning2}, {\color{black}with the edge orientations updated as described above,} depending on whether the pendant node is of Type I or Type II.
		\item[c.] Repeat the above two steps, until either only one node remains or no pendant node exists. 
	\end{enumerate}
Output: the resultant FAM ${\cal G}_{m, k, t}^{\mathrm{final}}$ when the algorithm terminates.  
\end{algorithm}

We make a few  observations. First, because ${\cal G}_{m, k, t}$ has $k$ nodes, and each pruning step reduces the number of nodes by one,  Algorithm~\ref{alg:pruning} must terminate in at most $(k-1)$ steps. 
Second, by Lemma~\ref{lemma:LMG}, $\mathrm{FS}({\cal G}_{m, k, t}) = \mathrm{FS}({\cal G}_{m, k, t}^{\mathrm{final}})$. As a result, we can replace $\mathrm{FS}({\cal G}_{m, k, t})$ by $\mathrm{FS}({\cal G}_{m, k, t}^{\mathrm{final}})$ in the decomposition of $C_m$ presented in Theorem~\ref{thm:Cm}. 
Finally, we consider how to compute $\mathrm{FS}({\cal G}_{m, k, t}^{\mathrm{final}})$. 
When the algorithm terminates, if there is only one node left, $\mathrm{FS}({\cal G}_{m, k, t}^{\mathrm{final}})={\bf 1}_n'v$, where $v\in\mathbb{R}^n$ is the label of this retaining node. By \eqref{Pruning1}-\eqref{Pruning2}, $v$ is obtained from $A$ and ${\bf 1}_n$ by using only three operations: multiplication, entry-wise product, and diagonal extraction; therefore, ${\bf 1}_n'v$ satisfies the requirement of an SEA term (Definition~\ref{def:SEA}). 
If the algorithm terminates with $\ell>1$ nodes, then $\mathrm{FS}({\cal G}_{m, k, t}^{\mathrm{final}})$ is an $\ell$-layer Full-Sum. The termination criteria requires that no pendant exists, which means that the number of summation layers cannot be further reduced by applying the pruning rules in \eqref{Pruning1}-\eqref{Pruning2}. 

The above observations are formalized in the following theorem:  
%%%%%%%%%%
%%%%%%%%%%
\begin{theorem} \label{thm:prune} 
Fix $m \geq 3$. Let $b_{m,k}$, ${\cal G}_{m, k, t}$ and $a_{m,k, t}$ be the same as in Theorem~\ref{thm:Cm}, and let ${\cal G}_{m, k, t}^{\mathrm{final}}$ be the output by applying Algorithm~\ref{alg:pruning} to ${\cal G}_{m,k,t}$. Then, 
\[
C_m = \sum_{k=2}^{m} \sum_{t=1}^{b_{m,k}} a_{m,k,t} \cdot \mathrm{FS}({\cal G}^{\mathrm{final}}_{m, k, t}). 
\]  
Furthermore, there are only two possible cases for each ${\cal G}^{\mathrm{final}}_{m,k,t}$:
\begin{itemize}
\item Case 1: there is only one node in this FAM. Let $v\in\mathbb{R}^n$ denote the label of this node. Then,  $\mathrm{FS}({\cal G}_{m, k, t}^{\mathrm{final}})={\bf 1}_n'v$, and it is an SEA term. 
\item Case 2: there is no pendant node. Then, there exist at least 4 nodes in this FAM, each with at least 3 immediate neighbors. In addition, $FS({\cal G}_{m, k, t}^{\text{final}})$ is an $\ell$-layer FS term with    $4\leq \ell \leq \lfloor m/2 \rfloor$. 
\end{itemize}
\end{theorem} 

Comparing Theorem~\ref{thm:prune} with Definition~\ref{def:CEEF}, we see that the above decomposition of $C_m$ satisfies the requirement of CEEF for every $m \geq 3$. In this sense, we have successfully solved the CEEF problem.

However,  at a practical level, for each specific $m$, one must still execute Algorithms~\ref{alg:merging}-\ref{alg:pruning} to derive the explicit expression of each $\mathrm{FS}({\cal G}_{m,k,t}^{\text{final}})$ (such as those shown in Examples~\ref{example1}-\ref{example2}). Although this process remains tedious, it no longer requires new theoretical insights; rather, it only requires a research assistant to correctly understand and follow Algorithms~\ref{alg:merging}-\ref{alg:pruning} step by step. For tasks of this nature, AI has important advantages over humans. In the next section, Section \ref{sec:results}, we describe how AI can be used to automatically implement the algorithm.

\begin{example}
We use the FAM in the third row of Table~\ref{tb:example-C4} to illustrate the pruning process. This FAM is visualized in the left panel of Figure~\ref{fig:pruning-example}.  Its associated FS term is $\sum_{j_1, j_2, j_3} (A \circ A)_{j_1 j_2} (A \circ A)_{j_2 j_3}$. We first prune   $j_3$ and then prune  $j_1$. The output is a one-node graph with node label $v = (A \circ A)^2{\bf 1}_n$. The original FS term has been converted equivalently to ${\bf 1}_n' v = {\bf 1}_n' (A \circ A)^2 {\bf 1}_n$, which is clearly an SEA term. 
\end{example}
%%%%%%%%%%%%%%%%%%%%%%%
	\begin{figure}[htbp]
		\begin{center}
			\begin{tikzpicture}[baseline, xscale=1.1, yscale=1.8,
				blacknode/.style={circle, fill=black, inner sep=1.5pt},
				every edge/.append style={thick}]
				% First diagram - Original graph
				\node[blacknode] (1) at (0,0) {};
				\node[blacknode] (2) at (1.5,0) {};
				\node[blacknode] (3) at (3,0) {};
				\node[scale=0.8] at (0,-0.4) {$j_1$};  % 从0.6改为0.8
				\node[scale=0.8] at (1.5,-0.4) {$j_2$};
				\node[scale=0.8] at (3,-0.4) {$j_3$};
				
				% Labels without dotted lines
				\node[scale=0.7] at (0,0.4) {$\textbf{1}_n$};  % 从0.5改为0.7
				\node[scale=0.7] at (1.5,0.4) {$\textbf{1}_n$};
				\node[scale=0.7] at (3,0.4) {$\textbf{1}_n$};
				
				% Double edges between nodes
				\draw (1) edge[bend left=20] node[above, scale=0.7] {$A$} (2);  % 从0.5改为0.7
				\draw (1) edge[bend right=20] node[below, scale=0.7] {$A$} (2);
				\draw (2) edge[bend left=20] node[above, scale=0.7] {$A$} (3);
				\draw (2) edge[bend right=20] node[below, scale=0.7] {$A$} (3);
				
				% Single arrow
				\draw[-stealth, thick] (4.3,0) -- (5.8,0);
				\node[scale=0.7] at (5,0.35) {Pruning $j_3$};  % 从0.5改为0.7
				
				% Second diagram - After shrinking C
				\node[blacknode] (4) at (7,0) {};
				\node[blacknode] (5) at (8.5,0) {};
				\node[scale=0.8] at (7,-0.4) {$j_1$};
				\node[scale=0.8] at (8.5,-0.4) {$j_2$};
				
				% Labels without dotted lines
				\node[scale=0.7] at (7,0.4) {$\textbf{1}_n$};
				\node[scale=0.7] at (9,0.4) {$(A\circ A) \cdot \textbf{1}_n$};
				
				% Double edges between nodes
				\draw (4) edge[bend left=20] node[above, scale=0.7] {$A$} (5);
				\draw (4) edge[bend right=20] node[below, scale=0.7] {$A$} (5);
				
				% Single arrow
				\draw[-stealth, thick] (10.5,0) -- (11.9,0);
				\node[scale=0.7] at (11.2,0.35) {Pruning $j_1$};
				
				% Third diagram - Final scalar
				\node[blacknode] (6) at (13.5,0) {};
				\node[scale=0.8] at (13.5,-0.4) {$j_2$};
				
				% Label without dotted line
				\node[scale=0.7] at (13.5,0.4) {$(A\circ A)^2 \textbf{1}_n$};
			\end{tikzpicture}
			\caption{An illustration of the pruning process for a FAM with three nodes. In this example, it happens that edge-label matrices of the original, intermediate, and final FAMs are all symmetric. Thus, edge orientations do not matter and are omitted in the plots.}  \label{fig:pruning-example}
		\end{center}
	\end{figure}

Our results are novel and nontrivial. We are the first to introduce the notions of SEA terms and IFS terms, which provide a rigorous framework for formulating the CEEF problem. Moreover, the pruning process proposed in this work has not appeared in the existing literature. We also formulate the entire algorithm exclusively in the language of graph theory, making it naturally interpretable and teachable to AI systems.

\section{Implementation with AI: Prompts, Results, Validation, and Comparison} \label{sec:results}

In Section~\ref{sec:main}, we have proposed an two-stage algorithm (Algorithms~\ref{alg:merging}-\ref{alg:pruning}) for deriving the CEEF formula of $C_m$ for each given $m$. However, the execution is labor-extensive. For example, the formula for $C_{12}$ has $1900$ terms, and to derive the explicit expression of each individual term is challenging for humans.

	To overcome the challenge, we use AI, but (not surprisingly) a {\it straightforward use} of AI won't work. 
	In particular, we have conducted experiments where we ask large language models (LLMs)  to either reason in natural language or write a code to derive a formula $C_m$ as desired (the full prompt is given in the Appendix). Such an approach only works 
	for small values of $m$ (e.g., $m=3,4,5$), where correct answers exist in the literature. When $m \geq 6$, 
	the desired formula either is not well-known or does not exist in the literature,   and LLMs fail to  output  a 
	satisfactory result:  It often guesses a few terms with incorrect coefficients and lacks a correct reasoning approach. 
	This suggests that AI is unable to solve the CEEF problem independently. We propose a hugAI approach which combines the strengths of human and AI.

\subsection{The human-guided AI (hugAI) approach} \label{subsec:humAI}

\begin{figure}[htb]
		\centering
		\includegraphics[width=1\linewidth]{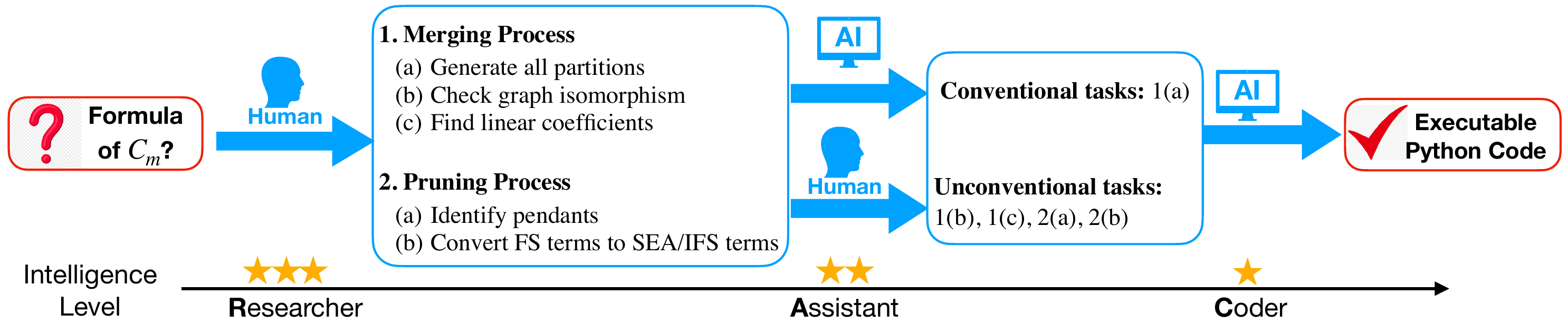}
		\caption{ The pipeline of our hugAI approach for counting cycles.} 
		\label{fig:pipelinedemoi}
	\end{figure}

	 In this approach,  first, we divide the CEEF problem into many steps, and outline a clear strategy for each step; this is mostly done by us (humans), as given in Section~\ref{sec:main}. Next, for each step, we either design an algorithm or formulate a clear algorithmic idea, and then ask AI to write a code for implementation.  Finally, we ask AI to generate executable Python code that takes $m$ as input and outputs the exact formula for $C_m$	in LaTeX format. The three steps outlined here require different levels of intelligence, which we denote as {\it level-R (Researcher)}, {\it level-A (Assistant)}, and {\it level-C (Coder)}, respectively (see Figure \ref{fig:pipelinedemoi}). It is of interest to assess the level of intelligence exhibited by AI; we report the results below.

		%%%%%%%%%%%%
	%%%%%%%%%%%%
	%%%%%%%%%%%%
	
For simplicity, we focus on DeepSeek-R1 (DS-R1), a recent reasoning model that demonstrates competitive performance on benchmark tasks \cite{guo2025deepseek}. We also include additional LLMs in our study; further details are provided in Section~\ref{sec:model-comparison}.

First, we prescribe the format for input and output.  For input, we teach DS-R1 to express graphs in strings:
	\smallskip
	\begin{quote}\small \itshape
%		Graph Representation: 
%		A naive loop with m vertices is represented as:
%		{1 [1,2]; 1 [1,m]; 1 [2,3]; 1 [3,4]; ... 1 [m-1,m]}. 
%		Each c [a,b] means an edge between vertices a and b with multiplicity c. {\color{red}(orientation?)}
An FAM is represented by:\\
- A dictionary node\_label[v], where each node v has a vector label. Initially every node label is ones\_n.\\ 
- A list of oriented edge records (src, dst, matrix\_label). The edge contributes matrix\_label[i\_src, i\_dst] to the Full Sum.\\
- Parallel edges are allowed.
	\end{quote}
	\smallskip
	For output, we ask DS-R1 to format the output as latex code that prints each term as either a  SEA term or an IFS term. The following prompt describes the string format with an example:
	\smallskip
	\begin{quote}\small \itshape
		If the final graph has at least 4 vertices, denote the vertices count as v, output the graph and the corresponding string in the following form: First, output $\sum_{i_{1} i_{2} ... i_{v}}$.
		Then, start from the smallest index to the biggest index, for each remaining vertex with assigned\_string\_vertex not being ${\bf 1}_n$, output $\langle$assigned\_string\_vertex of this vertex$\rangle$\_\{i\_\{vertex index\}\} . Then, use dictionary order, start from the smallest, for each two vertices with $t \geq 1$ remaining edge between them, output  $\langle$assigned\_string\_edge of edge 1 $\circ$ ... $\circ$ assigned\_string\_edge of edge t$\rangle$\_\{i\_\{vertex index 1\} i\_\{vertex index 2\}\}.  Join the above three parts to get the final expression, which should be in the form of a big sum.
	\end{quote}
	\smallskip

	Next, we divide our project into five major tasks, three for the merging process (1(a), 1(b), 1(c)), 
	two for the pruning process (2(a), 2(b)), as described in Figure \ref{fig:pipelinedemoi} and the caption of Table~\ref{tab:results}.  Based on the difficulty level,  1(a) is a {\it conventional task} for AI, and others are {\it unconventional tasks} 
	for AI. 
	
	Task 1(a) is well-studied in the literature and AI is able to accomplish it without much struggle.  For this task, we only provide a verbal description in the prompt (the full prompt is provided in Section~\ref{app:prompt} of the Appendix) and  DS-R1 is able to find an appropriate algorithm on its own. This demonstrates that DS-R1 has {\it level-A (Assistant) intelligence}.  
	
	The other $4$ tasks are unconventional for AI, and for each of them, AI needs detailed guidance and carefully written prompts. 
	For reasons of space, we only briefly discuss these tasks. 
	
	%are more challenging for AI.  For example, 
	%
	%An unconventional task is not well-defined in the literature and requires more human guideline. We consider three example tasks. (a) Graph isomorphism (merging partitions into isomorphic classes of multi-graphs). (b) M\"{o}bius coefficient calculation (computing the coefficients $a_{m,k,t}$ in Theorem~\ref{thm:Cm}). (c) Multi-graph conversion: Converting each multi-graph to either an SEA or IFS term. 
	%

	For Task 1(b), DS-R1 does not know what {\it isomorphism} means in the current setting, so we need to inform 
	it with the precise definition first.  After that, DS-R1 knows how to check isomorphism in a brute-force fashion, but the algorithm is very slow. To address the problem, we  inform DS-R1 that for checking isomorphism,  
	we only need to consider degree-preserved permutations. 
	This significantly speeds up the computing. Such kind of communication between DS-R1 and us is very common during  
	our study:   many times, DS-R1 struggles to understand some concepts, but after several  rounds of revision in the prompts, 
	it is able to write a code and produce the right answer.  
	
	For Task 1(c),  despite that the M\"{o}bius function %(Definition~\ref{def:Mobius})
	(i.e., the function $\mu(\hat{0}, \sigma_{m, k, t})$ in Theorem~\ref{thm:Cm})
	is well-known, we must teach DS-R1 how to relate it to our problem. As an example, we inform DS-R1 by prompts (see Section~\ref{app:prompt})  that we only need consider the pairs of partitions $\mu(\pi,\sigma)$ with $\pi=\hat{0}$ (the finest partition).  With a step-by-step guidance, DS-R1 is able to 
	produce the right results. 
	
	For Tasks 2(a)-2(b), since SEA and IFS terms are new concepts, we need to inform DS-R1 with the precise definitions first. 
	%We deliberately restrict DS-R1 to the `Coder' role, providing explicit step-by-step algorithmic specifications. 
	Our prompt starts with the prescribed input format (see the text above), followed by detailed instructions on the two types of pruning. 
	%A small part of the prompt for Type I pendant pruning is given below, and 
	The full prompt is contained in the Section~\ref{app:prompt}.

In summary,  first,  DS-R1 is able to accomplish conventional tasks, with correct answers. This demonstrates that DS-R1 has {\it level-A (Assistant) intelligence}. 
Second, DS-R1 is able to accomplish unconventional tasks, provided with a 
step-by-step guidance and carefully written prompts. This demonstrates that DS-R1 
has {\it level-C (Coder) intelligence} for these unconventional tasks: 
it is able to accomplish the task but requires a detailed step-by-step guidance. 
What is exciting is that DS-R1 also has preliminary {\it level-R (Researcher)} intelligence.   
One example is that, when we initially tried to prove Theorem \ref{thm:Cm}, we didn't realize it was connected to the  M\"{o}bius function. DS-R1 discovered this connection independently, and we followed its hint to identify the precise linear coefficients in our formula. 
Another example is,  DS-R1 is able to detect inconsistency in our prompt.  In one of our prompts, 
we provided DS-R1 with both a definition and an example of the linear coefficients, but 
unfortunately, the coefficients in the examples are not consistent with those in the definition. 
DS-R1 detected such an inconsistency. Using its hint, we revised the prompt, and 
this time,  DS-R1 fully  understood our point and provided the correct answer for our question.

\subsection{Results} 
	The final outcome of our hugAI approach is a Python code, which takes an integer $m \geq 3$ as input and output the desirable formula as in Theorem~\ref{thm:Cm} and Theorem~\ref{thm:prune}, where each FS term is converted to either a SEA term or an IFS term. Running this code for a given $m$ permits us to obtain the CEEF for $C_m$. 
	
	\begin{table}[htbp] 
	\centering
		\caption{Term count for the combinatorial formula of $C_m$ for $m=3,\cdots,12$. } 		\label{table:complexity} 
	\scalebox{.95}{ 
		\begin{tabular}{lcccccccccc} 
			\toprule
			$m$ & 3 &  4 &  5 & 6 & 7 & 8 & 9 & 10 & 11 & 12 \\ 
			\midrule
			\#Terms ($a_m$) & $1$  & $3$ & $3$ & $10$ & $15$ & $44$ &  $89$ & $254$ & $633$ & $1900$    \\ 
			\#SEA Terms & 1 & 3 & 3 & 10 & 15 & 43 & 86 & 239 & 581 & 1666 \\
			\#$O(n^4)$ IFS Terms & & & & & & 1 & 3 & 13 & 46 & 190 \\
			\#$O(n^5)$ IFS Terms & & & & & & & & 2 & 6 & 37 \\
			\#$O(n^6)$ IFS Terms & & & & & & & & & & 7 \\ 
			\bottomrule
		\end{tabular}
	}
	%\smallskip
	\end{table}

In Section~\ref{app:formula} of the Appendix, we present the resultant formulas of $C_m$ by running this code for $m=3, 4, \ldots, 12$. Table~\ref{table:complexity} presents the time complexity of computing $C_m$ using such a formula. Note that the time complexity to compute each SEA term is dominated by the cost of calculating matrix multiplications. The time complexity of each IFS term is $O(n^\ell)$, where $\ell$ is the number of summation layers, i.e. the number of nodes left after the final pruning process. IFS terms only emerge when $m \geq 8$. In this table, we have grouped the IFS terms according to their calculation time complexity. 

\begin{remark}[CEEF-inspired approximate formulas] \label{rmk:CC-proxy}
These CEEF results are valuable not only for the direct computation of $C_m$, but also for motivating new approximation methods. For example, in ongoing work, \cite{CC2} develop an approximate formula for each $C_m$ with $m \geq 8$ whose computational complexity is only $O(n^3)$. The key idea is to approximate each IFS term by a linear combination of SEA terms, yielding an approximation whose error is asymptotically negligible under the random spiked Wigner model. Simulation results show that this CEEF-inspired approximation is both more accurate and computationally more efficient than existing approximate cycle-counting algorithms, such as color coding.
%These CEEF formulas not only can be used for computation directly but also can motivate new approximate formulas of $C_m$. For example, in an ongoing work, \cite{CC2} developed an approximate formula for each $C_m$ for $m\geq 8$ that has only a complexity of $O(n^3)$. The idea is to inspect each IFS term and approximate it by a linear combination of SEA terms, so that the approximation error is negligible under a random spiked Wigner model. In their simulations, this CEEF-inspired approximate formula has higher accuracy and lower computational cost than other approximate cycle-counting algorithms such as color coding. 
\end{remark}

\subsection{Validation of the results} \label{subsec:validation}
	
	We now validate the correctness of these formulas. As established in Section~\ref{sec:main}, our algorithm (i.e., the high-level reasoning and derivation strategy) has already been rigorously justified; as a result, any correct execution of this strategy is guaranteed to yield a valid formula for $C_m$. Therefore, the purpose of this validation is not to re-examine the strategy itself, but rather to assess whether the AI correctly understands and executes the prescribed procedure.

\paragraph{Manual validation}
For $m \leq 8$, the formulas contain relatively few terms (see Table~\ref{table:complexity}) and can be derived manually. We therefore use these hand-derived expressions as the ground truth and find that the formulas produced by our method are correct for all $m \leq 8$. For larger values of $m$, there are few known results in the literature, except when $A$ is binary (e.g., \cite{Russian}). In this setting, the formulas are much simpler, as many terms are identical and can be combined (for example, $A \circ \cdots \circ A = A$). We apply our method and simplify the resulting expressions accordingly, and find that they agree with those reported in \cite{Russian} for $m \leq 13$.

\paragraph{Numerical validation} Manual validation becomes intractable for large $m$ or the general (non-binary) case. To address this limitation, we  introduce a numerical validation pipeline that enables verification of the formulas for both large $m$ and the general case. In this approach, we generate many $n\times n$ matrices $A$, compute the true $C_m$, and compare them with the values based on our formulas.

	One question is how to compute the true value of $C_m$. We may search ${n \choose m}$ indices in a brute-forth way, but a better solution is to use dynamic programming (DP);  
	see the Supplementary Material for details. 
	The DP method is guaranteed to output the true $C_m$, but its complexity is $O(n^22^n)$ for a fixed $m$. As a consequence, the DP-benchmark is feasible to compute only for small $n$. We thereby restrict our experiments to $n\leq 16$. 
	
		\begin{figure}[tb!]
	\centering
	\includegraphics[width=\textwidth]{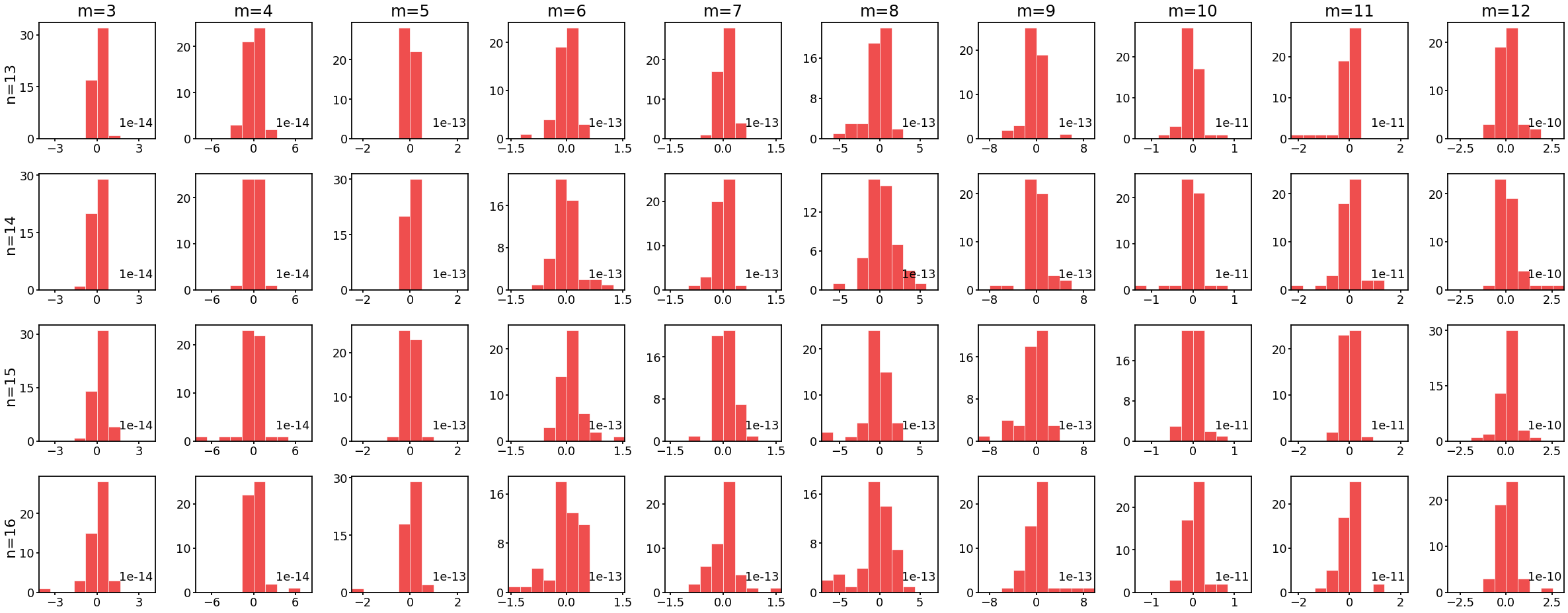}
	\vspace{-1.4em}
	\caption{Numerical validation (the histogram of discrepancies between the values by our formulas and the DP-benchmarks in the scale of $10^{-13}$ or $10^{-14}$, where $n=13,14,15,16$ and $m=3,\dots,12$).}
	\label{fig:compact-gallery}
	\end{figure}

	Another question is the following: if we observe matching $C_m$ values across a large number of small-size matrices $A$, how can we conclude that this provides evidence for the correctness of the formulas? An important observation is that the space of all possible formulas produced by our method is {\it finite} (see Definitions~\ref{def:SEA}-\ref{def:FS} and note that the coefficients in our formulas can only take integer values), whereas the data matrices $A$ we generate take {\it continuous} values. As a result, the probability that an incorrect formula generates correctly matching values is exactly zero, as formally established in the following lemma: 
	%%%%%%%%%%%%%%%%%%%%%%%%%%%%%%%%%%%%%%
	\begin{lemma}\label{lem:formula-verification}
	Fix $m\geq 3$. 
	Let $C^{\mathrm{DS}}_m(A)$ be the CEEF of $C_m(A)$ produced by the Python code output by DS-R1. If $C^{\mathrm{DS}}_m(A)$ is not algebraically equivalent to $C_m(A)$, then for any $n\geq 2$ and random matrix
	$A \in \mathbb{R}^{n \times n}$ whose entries have a jointly continuous distribution,
	$\mathbb{P}(C^{\mathrm{DS}}_m(A)=C_m(A))=0$.
	\end{lemma}
	
	%In light of this lemma, %binary matrices $A$ are not suitable for validation purposes.
	In our experiments, we sample $A$ from the Gaussian Orthogonal Ensemble (GOE), where $A_{ij}=A_{ji}\sim\mathcal{N}(0,1/n)$ for $i<j$ and
	$A_{ii}\sim\mathcal{N}(0,2/n)$; and for each setting, we use $t=50$ samples of $A$ with a
	fixed random seed. An important issue is the numerical precision required to declare two values as ``matching.'' For large $m$, individual terms in the formula may be small in magnitude, so that discrepancies can become effectively indistinguishable at finite precision. Fortunately, when $A$ is generated from the GOE, we can characterize the asymptotic order of each term in the formula, allowing us to calibrate an appropriate tolerance used in validation. Figure~\ref{fig:compact-gallery} presents the histograms of discrepancies between values from our formulas and the DP-benchmark, for a set of $(n,m)$.  
We observe that the values are well-matched up to tolerable numerical precision, confirming the correctness of our formulas. 
	
\begin{table}[tb!]
\centering
\caption{Average runtime (in seconds) of each $C_m$ for different values of $n$. Since this experiment is to validate our formula using the computationally expensive DP algorithm, we only consider small values of $n$. Experiments with much larger values of $n$ are deferred to Section~\ref{sec:applications}.}
\label{tab:runtime}
\scriptsize
\setlength{\tabcolsep}{1.5pt}
	\begin{tabular}{c cc cc cc cc}
		\toprule
		$m$ & \multicolumn{2}{c}{$n=13$} & \multicolumn{2}{c}{$n=14$} & \multicolumn{2}{c}{$n=15$} & \multicolumn{2}{c}{$n=16$} \\
		\cmidrule(lr){2-3}\cmidrule(lr){4-5}\cmidrule(lr){6-7}\cmidrule(lr){8-9}
		& Our Approach & DP & Our Approach & DP & Our Approach & DP & Our Approach & DP \\
		\midrule
		\rule{0pt}{2.4ex}
		3  & $2.84 \times 10^{-4}$ & $2.30 \times 10^{-2}$ & $2.93 \times 10^{-4}$ & $3.89 \times 10^{-2}$ & $1.61 \times 10^{-4}$ & $8.14 \times 10^{-2}$ & $1.87 \times 10^{-4}$ & $1.70 \times 10^{-1}$ \\
		4  & $1.41 \times 10^{-4}$ & $3.97 \times 10^{-2}$ & $2.51 \times 10^{-4}$ & $5.77 \times 10^{-2}$ & $1.04 \times 10^{-4}$ & $1.07 \times 10^{-1}$ & $1.18 \times 10^{-4}$ & $2.04 \times 10^{-1}$ \\
		5  & $1.05 \times 10^{-4}$ & $1.03 \times 10^{-1}$ & $2.30 \times 10^{-4}$ & $1.33 \times 10^{-1}$ & $8.08 \times 10^{-5}$ & $2.21 \times 10^{-1}$ & $8.75 \times 10^{-5}$ & $3.75 \times 10^{-1}$ \\
		6  & $3.01 \times 10^{-4}$ & $2.50 \times 10^{-1}$ & $3.68 \times 10^{-4}$ & $3.32 \times 10^{-1}$ & $2.23 \times 10^{-4}$ & $5.56 \times 10^{-1}$ & $2.34 \times 10^{-4}$ & $9.25 \times 10^{-1}$ \\
		7  & $4.43 \times 10^{-4}$ & $4.86 \times 10^{-1}$ & $4.76 \times 10^{-4}$ & $6.95 \times 10^{-1}$ & $3.38 \times 10^{-4}$ & $1.24 \times 10^{0}$ & $3.44 \times 10^{-4}$ & $2.18 \times 10^{0}$ \\
		8  & $4.13 \times 10^{-3}$ & $7.48 \times 10^{-1}$ & $3.75 \times 10^{-3}$ & $1.17 \times 10^{0}$ & $3.98 \times 10^{-3}$ & $2.27 \times 10^{0}$ & $3.96 \times 10^{-3}$ & $4.33 \times 10^{0}$ \\
		9  & $1.14 \times 10^{-2}$ & $9.47 \times 10^{-1}$ & $9.40 \times 10^{-3}$ & $1.62 \times 10^{0}$ & $9.18 \times 10^{-3}$ & $3.42 \times 10^{0}$ & $9.50 \times 10^{-3}$ & $7.09 \times 10^{0}$ \\
		10 & $4.49 \times 10^{-2}$ & $1.06 \times 10^{0}$ & $4.04 \times 10^{-2}$ & $1.92 \times 10^{0}$ & $3.77 \times 10^{-2}$ & $4.35 \times 10^{0}$ & $4.94 \times 10^{-2}$ & $9.73 \times 10^{0}$ \\
		11 & $1.26 \times 10^{-1}$ & $1.08 \times 10^{0}$ & $1.21 \times 10^{-1}$ & $2.06 \times 10^{0}$ & $1.22 \times 10^{-1}$ & $4.88 \times 10^{0}$ & $1.23 \times 10^{-1}$ & $1.15 \times 10^{1}$ \\
		12 & $5.13 \times 10^{-1}$ & $1.11 \times 10^{0}$ & $4.95 \times 10^{-1}$ & $2.10 \times 10^{0}$ & $5.01 \times 10^{-1}$ & $5.10 \times 10^{0}$ & $5.02 \times 10^{-1}$ & $1.26 \times 10^{1}$ \\
		% 13 & -- & -- & $1.81 \times 10^{0}$ & $2.11 \times 10^{0}$ & $1.80 \times 10^{0}$ & $5.14 \times 10^{0}$ & $1.82 \times 10^{0}$ & $1.28 \times 10^{1}$ \\
		% 14 & -- & -- & -- & -- & $7.08 \times 10^{0}$ & $5.11 \times 10^{0}$ & $7.16 \times 10^{0}$ & $1.28 \times 10^{1}$ \\
		% 15 & -- & -- & -- & -- & -- & -- & $2.76 \times 10^{1}$ & $1.28 \times 10^{1}$ \\
		\bottomrule
\end{tabular}
\end{table}

This small-scale experiment also supports the computational value of having a CEEF formula. In Table~\ref{tab:runtime}, we report the runtime: For small $n$, our method is already substantially faster than DP. Note that for large $n$, DP is infeasible to implement, while our method is still implementable.

\subsection{Comparison of representative LLMs} \label{sec:model-comparison}

While our main experiments focus on DS-R1, it is also of interest to evaluate how other LLMs perform under the same prompts. To this end, we test 11 additional models from seven providers (see Table~\ref{tab:models} in the Appendix for model details). For all models, web search is disabled, and reasoning (or thinking) mode is enabled whenever supported by the model interface.

Recall that the AI implementation of our method comprises two stages, Merging and Pruning, which together involve five tasks (1a-1c and 2a-2b; see the caption of Table~\ref{tab:results}x). To facilitate a more systematic comparison across LLMs, we decompose the prompting-and-response pipeline into two components, rather than requiring a single monolithic response:
\begin{itemize}
\item Stage 1 (Merging). An LLM receives Stage-1-User-Prompt (see Section~\ref{app:prompt})
and is asked to produce a Python function \texttt{step1(m)} that enumerates all 
partitions of the $m$-cycle, filters self-loops, groups the resulting
multi-graphs by isomorphism, and computes the aggregated M\"obius coefficient
for each canonical type. 
\item Stage~2 (Pruning). An LLM receives Stage 2 User Prompt (see Section~\ref{app:prompt})  together with its own verified Stage~1 code as context.  It is asked to produce a function \texttt{C\_m(A,\,m)} that applies pendant pruning to each canonical type and constructs the corresponding SEA or IFS terms. 
\end{itemize}
This two-component design avoids a potential output-length truncation issue: a single prompt frequently yields an output that exceeds the 32\,k-token limit before completing all tasks. It also enables a better diagnosis---when the formula is incorrect, we can identify whether it is from Stage~1 or Stage~2. 

The Stage~1 performance is validated by comparing the canonical multi-graphs and their coefficients with the Stage~1 output of DS-R1, which has been validated extensively in Section~\ref{subsec:validation} and is regarded the ground truth. 
%generated by the reference implementation. 
The Stage~2 performance is validated similarly as in Section~\ref{subsec:validation}, where we apply the Python function $\texttt{C\_m(A,\,m)}$ to continuous-valued random matrices and compare them with the DP benchmarks (with a numeric tolerance of $10^{-8}$). 
The results are summarized in Table~\ref{tab:results}. 

%%%%%%%%%%%%%%%%%%%%%%%%%%%%%%%%%%%%%%%
	\begin{table}[htb]
		\centering
		\caption{Evaluation of 12 LLMs (DS-R1 and 11 other models) for the two-stage implementation task.
			Stage~1 consists of four steps:
			\textbf{Syntax(1)}~(code runs),
			\textbf{1a}~(valid partition enumeration),
			\textbf{1b}~(canonical graph types),
			\textbf{1c}~(M\"obius coefficients).
			Stage~2 consists of two steps:
			\textbf{Syntax(2)}~(code runs) and
			\textbf{2a/2b}~(the generated formula matches the numerical benchmark).
			A~\cmark\ indicates success throughout the reported validation range;
			\xmark\ indicates at least one failure.}
		\label{tab:results}
		\scalebox{1}{
			\renewcommand{\arraystretch}{1.15}
			\setlength{\tabcolsep}{4pt}
			\begin{tabular}{@{}l | cccc | cc | c@{}}
				\toprule
				\multirow{2}{*}{\textbf{Model}} & \multicolumn{4}{c|}{\textbf{Stage 1: Merging}} & \multicolumn{2}{c|}{\textbf{Stage 2: Pruning}} & \multirow{2}{*}{\textbf{Summary}}\\
				\cmidrule(lr){2-5} \cmidrule(lr){6-7}
				& Syntax(1) & 1a & 1b & 1c
				& Syntax(2) & 2a/2b & \\
				\midrule
				Claude Opus~4.8     & \cmark & \cmark & \cmark & \cmark & \cmark & \cmark & \multirow{10}{*}{Successful} \\
				Claude Sonnet~4.6   & \cmark & \cmark & \cmark & \cmark & \cmark & \cmark &  \\
				DeepSeek R1     & \cmark & \cmark & \cmark & \cmark & \cmark & \cmark &  \\
				DeepSeek~V4 Pro     & \cmark & \cmark & \cmark & \cmark & \cmark & \cmark &  \\
				GPT-5.5             & \cmark & \cmark & \cmark & \cmark & \cmark & \cmark &  \\
				Gemini~3.1 Pro      & \cmark & \cmark & \cmark & \cmark & \cmark & \cmark &  \\
				Gemini~3.5 Flash    & \cmark & \cmark & \cmark & \cmark & \cmark & \cmark &  \\
				GLM~5.1             & \cmark & \cmark & \cmark & \cmark & \cmark & \cmark &  \\
				Kimi~K2.6           & \cmark & \cmark & \cmark & \cmark & \cmark & \cmark &  \\
				MiniMax-M3          & \cmark & \cmark & \cmark & \cmark & \cmark & \cmark &  \\
				\midrule
				MiniMax-M2.7        & \cmark & \cmark & \cmark & \cmark & \xmark & \xmark & Stage~1 only \\
				\midrule
				DeepSeek~V3.2       & \xmark & \xmark & \xmark & \xmark & \xmark & \xmark & Unsuccessful\\
				\bottomrule
		\end{tabular}}\label{table:comparisonresult}
	\end{table}

We have some findings. First, ten of the eleven models successfully complete the Stage~1 tasks. For example, these models correctly identify 41 surviving partitions and 10 canonical multigraphs at $m = 6$, and 715 surviving partitions and 44 canonical multigraphs at $m = 8$. This suggests that the combinatorial enumeration task, while nontrivial, lies within the reliable capabilities of most frontier models when provided with a precise algorithmic specification.
Second, Stage~2 further differentiates model performance. Nine of the models that pass Stage~1 also produce numerically correct formulas over the reported validation range. MiniMax-M2.7 passes the Stage~1 structural checks, but its Stage~2 code cannot be validated because the generated program fails to import. DeepSeek~V3.2 fails the Stage~1 structural checks, and its Stage~2 output likewise fails the numerical validation. This pattern highlights the value of separating the Merging and Pruning stages: a failure in Stage~2 has a different interpretation depending on whether the Stage~1 output has first been verified.

\begin{remark}[Variability across LLM runs]
We observe some variability in LLM performance across repeated runs. For example, GPT-5.5 occasionally fails in Stage~2, whereas DeepSeek~V3.2 occasionally succeeds in both stages. The results reported in Table~\ref{table:comparisonresult} correspond to the majority outcome across multiple runs of each model.
\end{remark}

\section{Applications} \label{sec:applications} 

In this section, we present several applications of the cycle count statistic $C_m$,
and illustrate the advantage of using a large value of $m$.  
Since these applications require large values of $m$, solving the CEEF problem is particularly important.

%\begin{remark}[Cycle count statistics versus trace statistics]
%The cycle count statistic $C_m$ defined in \eqref{Defi:Cm} is closely related to the statistic $T_m= \mathrm{trace}(A^m)$, but $C_m$ has several appealing properties that $T_m$ does not satisfy. For example, in network settings, it is common to assume that, for some low-rank matrix $\Omega$, $\mathbb{E}[A] = \Omega - \mathrm{diag}(\Omega)$. Under a sparse network regime and a mild regularity condition, one has 
%$\mathbb{E}[C_m] \approx \mathrm{trace}(\Omega^m)$ and $\mathrm{Var}(C_m) \approx C_m$.  
%These properties make cycle count statistics particularly useful in problems such as network testing and network goodness-of-fit, 
%among others \cite{Jin2021optimal, EstK, GOF}.  
%\end{remark}

\subsection{Detection of weak spikes using cycle count statistics} \label{subsec:spike-detection}

The spiked Wigner models \cite{capitaine2009largest, knowles2014outliers,perry2018optimality,chi2019nonconvex,cheng2021tackling,fan2022asymptotic} are widely used in random matrix theory, applied mathematics, statistics, and information theory. Specifically, let $Z\in\mathbb{R}^{n\times n}$ be a symmetric random matrix comprising independent variables in its upper triangle, where $Z_{ij}\sim {\cal N}(0, 1/n)$ when $i<j$ and $Z_{ii}\sim {\cal N}(0, 2/n)$. In the literature, $Z$ is called a standard Wigner matrix or a Gaussian Orthogonal Ensemble (GOE). 
The data matrix $A$ satisfies that  
\begin{equation}\label{spiked-model}
A = \Omega + Z=\text{`signal'} + \text{`noise'},\qquad \mbox{where $\Omega$ has only $K$ nonzero eigenvalues $\lambda_1,\ldots, \lambda_K$}.  
\end{equation}
Each $\lambda_k$ is called a spiked eigenvalue. In the spike detection problem, we test between two hypotheses: 
\[
H_0: K=0 \qquad \mbox{versus}\qquad H_1: K\geq 1. 
\] 
In this framework,  
\cite{JinKeSuiWang2025weakspikes} showed that $C_m\approx {\cal N}(0, 2m)$ when $H_0$ holds, and $C_m$ is substantially nonzero when $H_1$ holds (with mild regularity conditions). 
Therefore, for each $m$, we may use $C_m/\sqrt{2m}$ as the test statistic. 
\cite{JinKeSuiWang2025weakspikes} also found that when spiked eigenvalues are only slightly larger than $1$ (i.e., weak spikes), the testing power increases rapidly with the order of the cycle count statistic. They suggested to use a weighted linear combination of $C_m$ as the test statistic: For an integer $N\geq 3$, let 
\begin{equation} \label{App1-test-stat}
\phi_N = \frac{1}{\sqrt{B_N}}\sum_{m=1}^N\frac{C_m}{2m}, \qquad\mbox{with}\qquad B_N = \sum_{m=1}^N \frac{1}{2m}. 
\end{equation}
Here, $C_m$ denotes the cycle count statistic when $m\geq 3$. For $m<3$, they defined $C_1=\mathrm{trace}(A)$, and $C_2 = \mathrm{trace}(A^2)-(n-1)$.  
\cite{JinKeSuiWang2025weakspikes} showed that, in the extremely weak spike regime, $N=\Omega(\log(n))$ is necessary for fully distinguishing the two hypotheses. 
Their work not only demonstrates the effectiveness of cycle count statistics for weak spike detection, but also suggests that incorporating high-order $C_m$ leads to improved statistical performance. 

\begin{figure}[tbp]
\centering
\begin{minipage}[t]{0.31\textwidth}
\vspace{0pt}
\centering
\includegraphics[width=\linewidth,height=0.35\textheight,keepaspectratio]{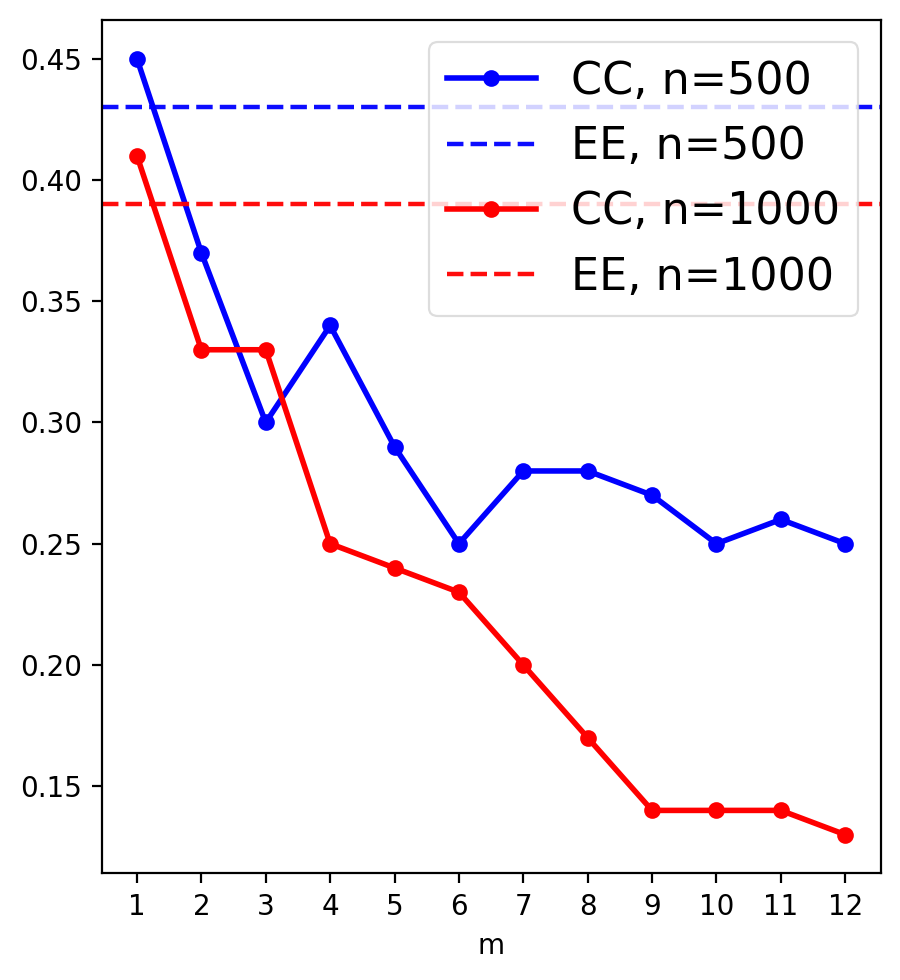}
\end{minipage}\hspace{.5cm}
\begin{minipage}[t]{0.63\textwidth}
\vspace{0pt}
\centering
\includegraphics[width=\linewidth,height=0.35\textheight,keepaspectratio]{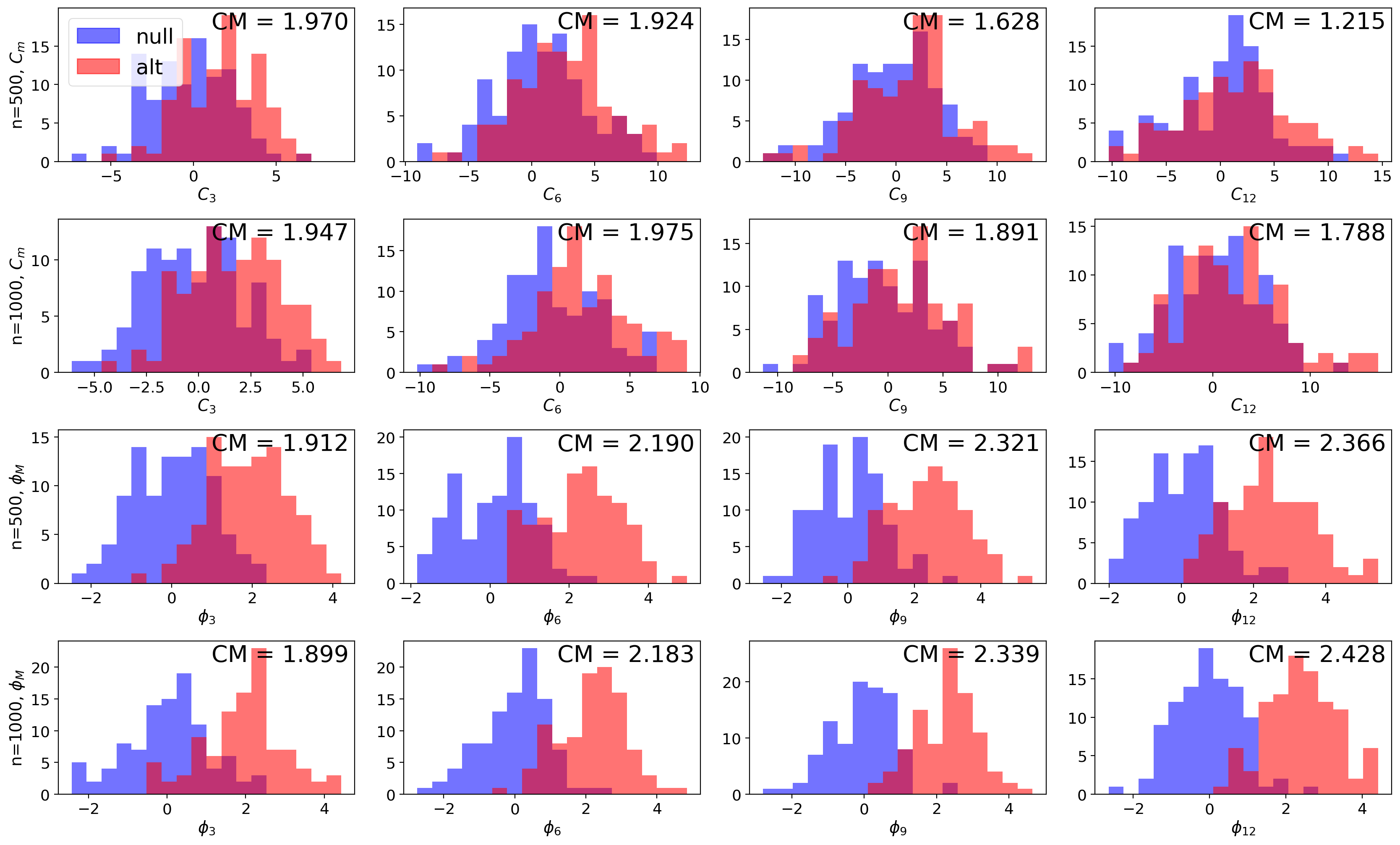}
\end{minipage}
\caption{Detection of weak spikes using cycle count statistics. Left: ideal testing error of $\phi_m$ versus $m$ (the test statistic $\phi_m$ is a weighted linear combination of cycle count statistics until order $m$), with a comparison to the testing approach by using empirical eigenvalues (EE). Right: histograms of $C_m$ and $\phi_m$ under the null and alternative hypotheses, for $m\in \{3,6,9,12\}$, where the numbers on the top right of these plots are the Contrast Mean (CM).}
\label{fig:testing-results}
\end{figure}

To this end, we conduct a simulation study with $K=2$ and $n\in \{500, 1000\}$. The two spiked eigenvalues are $\lambda_1=\lambda_2=1$. We sample 100 data matrices $A$ from the null hypothesis and $100$ from the alternative hypothesis. The ideal testing error is the smallest sum of type-I and type-II errors on these 200 samples, when the rejection threshold is optimized. 
In the left panel of Figure~\ref{fig:testing-results}, we report the ideal testing error of $\phi_m$ for $m=1,2,\ldots,  12$, and compare it with the empirical eigenvalues (EE) approach which uses $|\lambda_1|$ as the test statistic. In the right panel of Figure~\ref{fig:testing-results}, we report the histograms of $C_m$ and $\phi_m$ under two hypotheses, as well as the difference of means under two hypotheses (Contrast Mean), for $m\in \{3,6,9,10\}$. 
All the cycle count statistics $C_m$ involved in this experiment are computed using the formulas output by our approach in Sections~\ref{sec:main}-\ref{sec:results}.   

The results suggest that the cycle-count (CC) testing approach substantially outperforms the EE approach, and the testing error of the CC approach decays evidently as $m$ increases, confirming the benefit of computing high-order $m$ in this application. 
When $n=500$, the brute-forth approach or the dynamic programming is only practically feasible for $m\leq 4$. In contrast, our CEEF formula enables computation for all $m\leq 12$. For example, it takes only a few seconds to compute $C_{12}$. Even when $n$ increases to $1000$, the computing time of $C_{12}$ is still less than a minute. 

%Compared to using $C_m$ only, the weighted combination $\phi_m$ has a better performance.  

%The numerical trend is fully consistent with the conclusions of the testing
%paper: CC is substantially more effective than EE in the
%weak-signal regime. For $n=1000$, the ideal error decreases from about $0.410$
%at $m=1$ to about $0.130$ at $m=12$, while EE remains at about
%$0.390$. For $n=500$, the best CC error is about $0.250$, attained
%at $m=12$, whereas the EE error is about $0.430$. The main qualitative message
%is that higher-order cycle information materially improves testing performance:
%as more cycle statistics are aggregated, the CC error decreases steadily and
%its advantage over EE becomes increasingly pronounced. This trend is especially
%clear at the larger dimension, where the gain from moving to higher-order
%statistics is the most substantial. Figure~\ref{fig:testing-results} summarizes the testing evidence in
%two complementary ways. The left panel shows the ideal testing error as a
%function of the truncation order $m$, while the right panel displays the
%histograms of both the raw cycle quantities $C_m$ and the aggregated
%statistics $\phi_m$ for $m=3,6,9,12$. In both $n=500$ and $n=1000$, the
%separation between the null and alternative distributions becomes clearer after
%aggregation into $\phi_m$, and the separation becomes more visible for larger
%values of $m$, which is consistent with the sharper testing performance
%observed in the error curves.

\subsection{Estimation of weak spikes using cycle count statistics} \label{subsec:Newton-Prony}
Cycle count statistics are useful not only for detecting spikes but also for estimating them. As shown in \cite{JinKeSuiWang2026spikes}, $C_m$ provides an estimate of the $m$th power sum of the spiked eigenvalues, because 
\begin{equation} \label{Cm-expectation}
\mathbb{E}[C_m]\approx \mathrm{trace}(\Omega^m) = \lambda_1^m + \lambda_2^m + \ldots + \lambda_K^m:=c_m. 
\end{equation}
The problem of estimating the spiked eigenvalues reduces to recovering $K$ numbers from their power sums. Newton's identities \cite{Newton1707} and Prony's method \cite{Prony1795} are classical approaches to this problem. For convenience, we combine them to a single matrix identity as follows: Let $f(x) = \prod_{k=1}^K (x - \lambda_k) = x^K + \prod_{j=0}^{K-1}a_jx^j$ be the characteristic polynomial associated with $\lambda_1,\ldots,\lambda_K$. Write $a = (a_0, a_1, \ldots, a_{K-1})^{\top}$ and $c_m=\sum_{k=1}^K\lambda_k^m$. For any $N \geq 1$, $a$ satisfies the following linear equation:
%%%%%%%%%%
%%%%%%%%%%
%%%%%%%%%%
\begin{equation} \label{NP-identity} 
  {\renewcommand{\arraystretch}{0.7}
\begin{bmatrix}
0 & 0 & \cdots & 0 & 1  \\
0 & 0 & \cdots & 2 & c_1  \\
 \vdots & \vdots& \ddots & \vdots & \vdots   \\
 K & c_1  &  \cdots &  c_{K-2}  & c_{K-1}  \\ 
 c_1 & c_2 & \ldots & c_{K-1} & c_K  \\ 
  \vdots  & \vdots & \vdots & \vdots  \\
  c_{N-K} &  c_{N-K+1}   &  \cdots & c_{N -2}  & c_{N-1}  
\end{bmatrix} a = - \begin{bmatrix} c_1\\ c_2\\\vdots \\ c_K\\c_{K+1}\\\vdots\\c_N\end{bmatrix}
}. 
\end{equation}
Combining the insights in \eqref{Cm-expectation}-\eqref{NP-identity}, \cite{JinKeSuiWang2026spikes} proposed the Newton-Prony (NP) method for estimating $\lambda_1,\ldots,\lambda_K$. This method first estimates $c_m$ by the corresponding cycle count statistic, then solves $\hat{a}=(\hat{a}_0, \hat{a}_1, \ldots, \hat{a}_{K-1})^{\top}$ from the above linear equation, and finally obtains $\hat{\lambda}_1,\ldots,\hat{\lambda}_K$ from solving $x^K + \prod_{j=0}^{K-1}\hat{a}_j x^j =0$.

\begin{figure}[tbp]
\centering
\begin{minipage}[t]{0.32\textwidth}
\vspace{0pt}
\centering
\includegraphics[width=\linewidth,height=0.31\textheight,keepaspectratio]{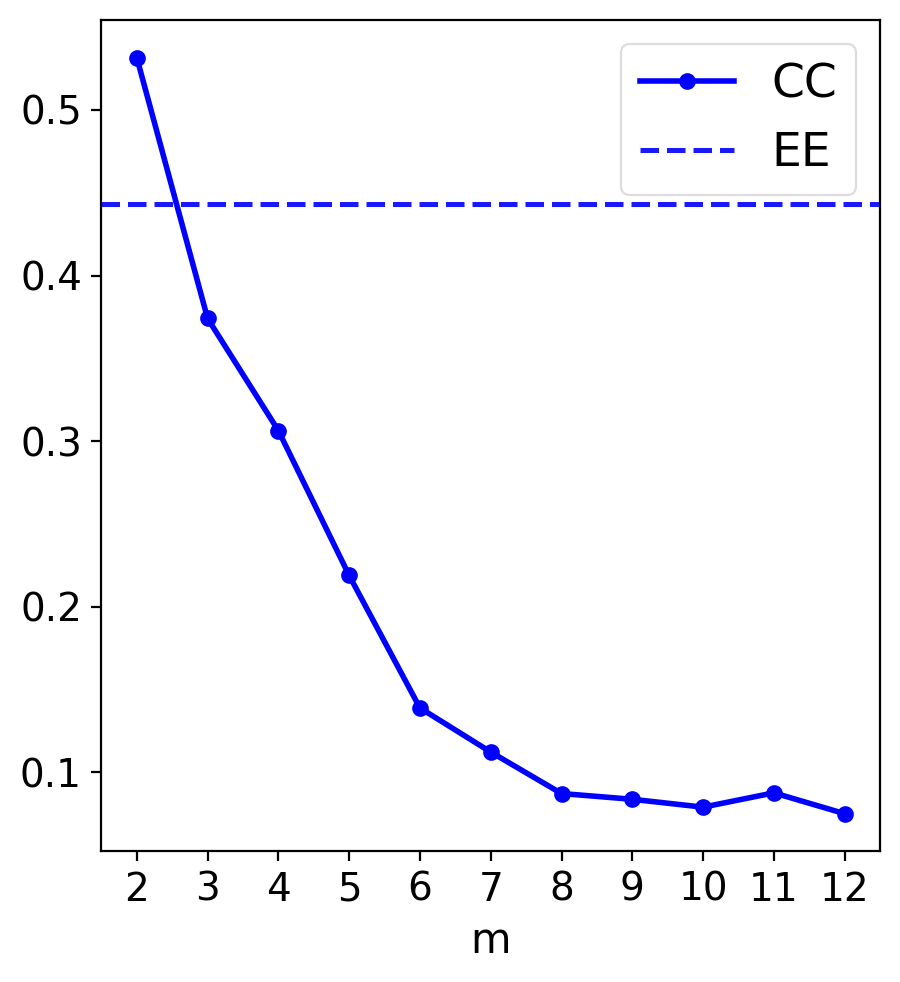}
\end{minipage}\hspace{.5cm}
\begin{minipage}[t]{0.62\textwidth}
\vspace{0pt}
\centering
\includegraphics[width=\linewidth,height=0.31\textheight,keepaspectratio]{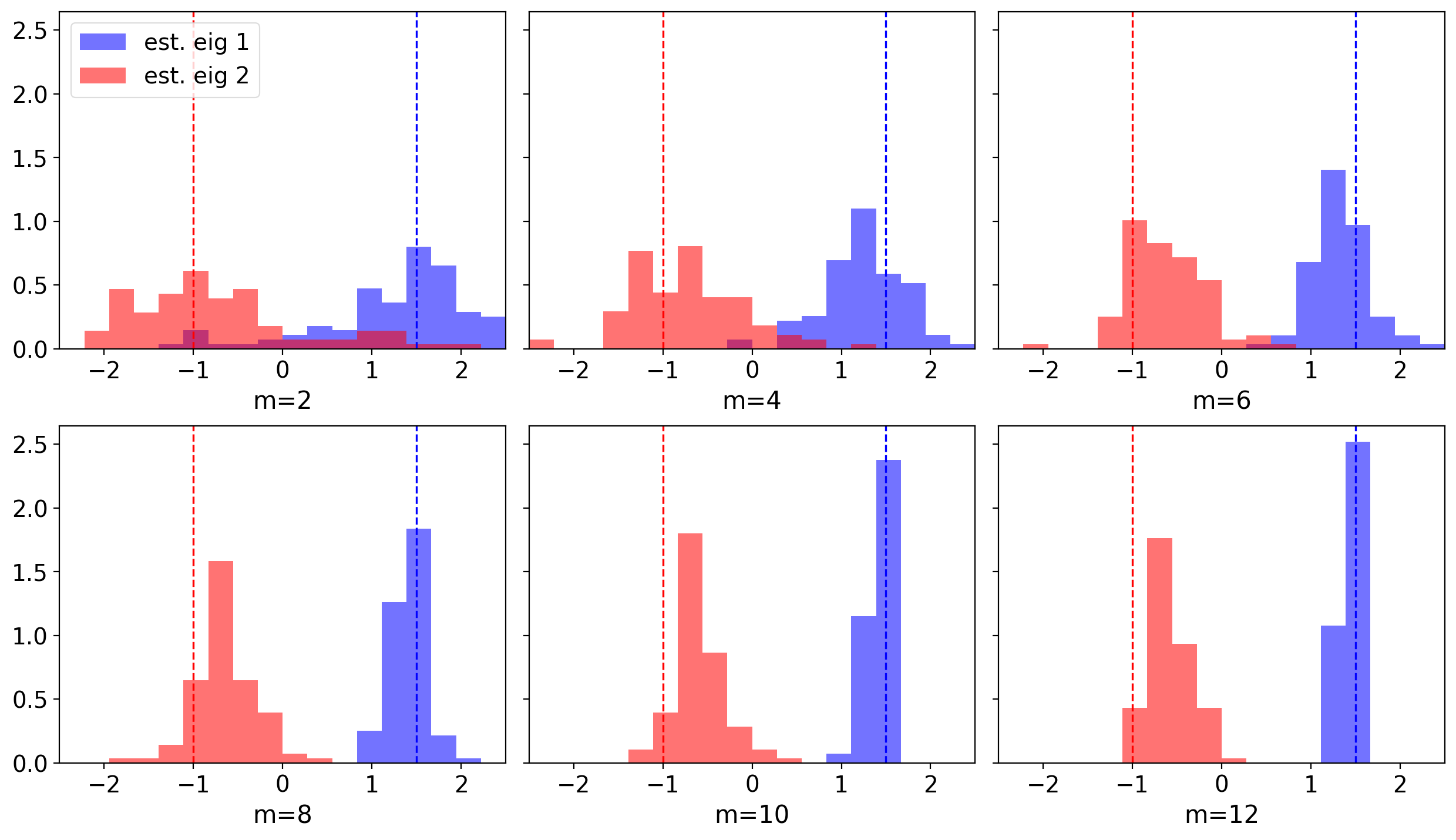}
\end{minipage}
\caption{Estimation of weak spikes using cycle count statistics ($n=1000$, $\lambda_1=1.5$, and   $\lambda_2 = -1$). Left: normalized mean-squared error (NMSE) versus $m$. Right: histograms of $\hat{\lambda}_1$ and $\hat{\lambda}_2$ for $m\in \{2,4,6,8,10,12\}$.}
\label{fig:estimation-results}
\end{figure}

We conduct a simulation study and compare this NP estimator with the empirical eigenvalue (EE) estimator. We fix $n=1000$ and $K=2$, and generate data from the model in \eqref{spiked-model}. The left panel of Figure~\ref{fig:estimation-results} reports the normalized mean-squared error $\mathrm{NMSE}=[\sum_{k=1}^K (\hat{\lambda}_k-\lambda_k)^2]/(\sum_{k=1}^K\lambda_k^2)$ based on 100 repetitions, for $m=2,3,\ldots,12$. 
The results suggest that the cycle-count-based NP estimator substantially outperforms the empirical eigenvalue estimator. Moreover, the estimation error decreases rapidly as $m$ increases. The right panel of Figure~\ref{fig:estimation-results} further breaks down the performance by individual eigenvalue, showing that the estimation error decreases with $m$ for both $\lambda_1$ and $\lambda_2$.  

This experiment confirms that cycle count statistics are effective for estimating spiked eigenvalues, and the estimation accuracy improves as $m$ increases, motivating the computation of high-order $C_m$. 
As noted in Section~\ref{subsec:spike-detection}, our CEEF formula enables the efficient computation of $C_m$ for $m\leq 12$, whereas the brute-forth approach is feasible only for $m\leq 4$.

\subsection{Network pairwise comparison using cycle count statistics}
We consider an application of cycle count statistics on network data. Let $A\in\mathbb{R}^{n\times n}$ be the adjacency matrix of an undirected network. We assume there is no self-loop, so all diagonal entries of $A$ are zero. Suppose the edges are independently generated, and let $\Omega\in\mathbb{R}^{n\times n}$ be a symmetric low-rank matrix such that $\Omega_{ij}=\mathbb{P}(A_{ij}=1)$. It follows that
\begin{equation} \label{network-model}
A = \Omega - \mathrm{diag}(\Omega) + W=\text{`main signal'}- \text{`secondary signal'}+\text{`noise'}, \qquad\mbox{where}\quad W = A-\mathbb{E}[A].  
\end{equation}
Here, $W$ is a generalized Wigner matrix (compared to the standard Wigner matrix $Z$ in \eqref{spiked-model}, the entries of $W$ are non-Gaussian and can have unequal variances). 
We call \eqref{network-model} the low-rank network model. It includes well-known models such as the stochastic block model, degree-corrected stochastic block model, and random dot product model as special cases.  
 
Network pairwise comparison \cite{ghoshdastidar2020two} has various applications in multiple-network analysis, such as change-point detection \cite{xie2013sequential} for dynamic networks. This problem is formulated as follows: Given two adjacency matrices $A_1$ and $A_2$ on the same set of nodes, suppose they both satisfy the low-rank network model in \eqref{network-model}, with parameter matrices $\Omega_1$ and $\Omega_2$, respectively. The aim is to test between two hypotheses: 
\[
H_0: \Omega_1=\Omega_2 \qquad\mbox{versus}\qquad H_1: \Omega_1\neq \Omega_2.
\]
In this framework, cycle count statistics can be computed from $A_1$, $A_2$, or their difference $\Delta = A_1 - A_2$. We use $C_m(X)$ to denote the cycle count statistic computed from a data matrix $X$. The IBM test statistic proposed in \cite{JinKeLuoMa2025pairwise} is given by
\begin{equation} \label{IBM-test-stat}
\psi_m:= \frac{C_m(\Delta)}{\sqrt{2^{m}m[C_m(A_1)+C_m(A_2)]}}.
\end{equation}
Here, $C_m(\Delta)$ provides an estimate of $\mathrm{trace}((\Omega_1-\Omega_2)^m)$, which contains the signal for distinguishing the two hypotheses. The quantities $C_m(A_1)$ and $C_m(A_2)$ are combined to estimate $\mathrm{Var}(C_m(\Delta))$, thereby normalizing the test statistic. 

To avoid signal cancellation caused by positive and negative eigenvalues of $\Omega_1-\Omega_2$, \cite{JinKeLuoMa2025pairwise} recommended using even values of $m$. Accordingly, we write 
$U_r=\phi_{2r}$ for all $r\geq 2$. The studies in \cite{JinKeLuoMa2025pairwise} were restricted to $r\in \{2,3\}$, where it was noted that larger values of $r$ could potentially improve the testing power but at the cost of substantially heavier computation. 
Our CEEF formula enables the computation of $U_r$ for larger values of $r$. To this end, we conduct a simulation study to examine whether increasing $r$ yields improved performance. 

Fix $(n, K)=(500, 2)$. We sample $(A_1, A_2)$ from the degree-corrected mixed membership (DCMM) model \cite{Jin2021optimal} with $K$ communities. Let $P=(1-b_n)I_2+b_n \mathbf 1_2 \mathbf 1_2^\top$, for some $b_n\in (0,1)$, and 
$\Theta=\mathrm{diag}(\theta_1, \theta_2,\ldots,\theta_n)$ be a diagonal matrix consisting of the degree heterogeneity parameters, and $\Pi=[\pi_1, \pi_2, \ldots,\pi_n]^{\top}$ be the membership matrix so that each $\pi_i$ describes node $i$'s mixed membership across two communities. 
The DCMM model is a special low-rank network model with $\Omega=\Theta\Pi P\Pi^{\top}\Theta$. 
In our experiments, we generate $\Theta$ as follows: draw $\theta^*_{i}\sim \mathrm{Uniform}(2,3)$ and then set 
$\theta_i=\beta_n \theta^*_i/\|\theta^*\|_2$, so that $\|\theta\|_2=\beta_n$, for some $\beta_n>0$. 
We also draw $\pi_i \sim \mathrm{Dirichlet}(1.6,0.4)$ and
$\widetilde{\pi}_i \sim \mathrm{Dirichlet}(1,1)$. Under the null hypothesis, $A_1$ and $A_2$ are independently drawn from Model~\eqref{network-model} with the same $\Omega=\Theta\Pi P\Pi^{\top}\Theta$. Under the alternative hypothesis, $A_1$ and $A_2$ are drawn with $\Omega$ and $\widetilde{\Omega}=\Theta\widetilde{\Pi} P\widetilde{\Pi}^{\top}\Theta$, respectively. 
Following \cite{JinKeLuoMa2025pairwise}, we fix $\beta_n=6$ and select the value of $b_n$ so that the signal-to-noise ratio (SNR) defined in \cite{JinKeLuoMa2025pairwise} is equal to a pre-specified number. We consider two scenarios:  SNR $=2$, and SNR $=1.5$.

\begin{figure}[tbp]
\centering
\begin{minipage}[t]{0.4\textwidth}
\vspace{0pt}
\centering
\includegraphics[width=\linewidth, height=1.35\linewidth]{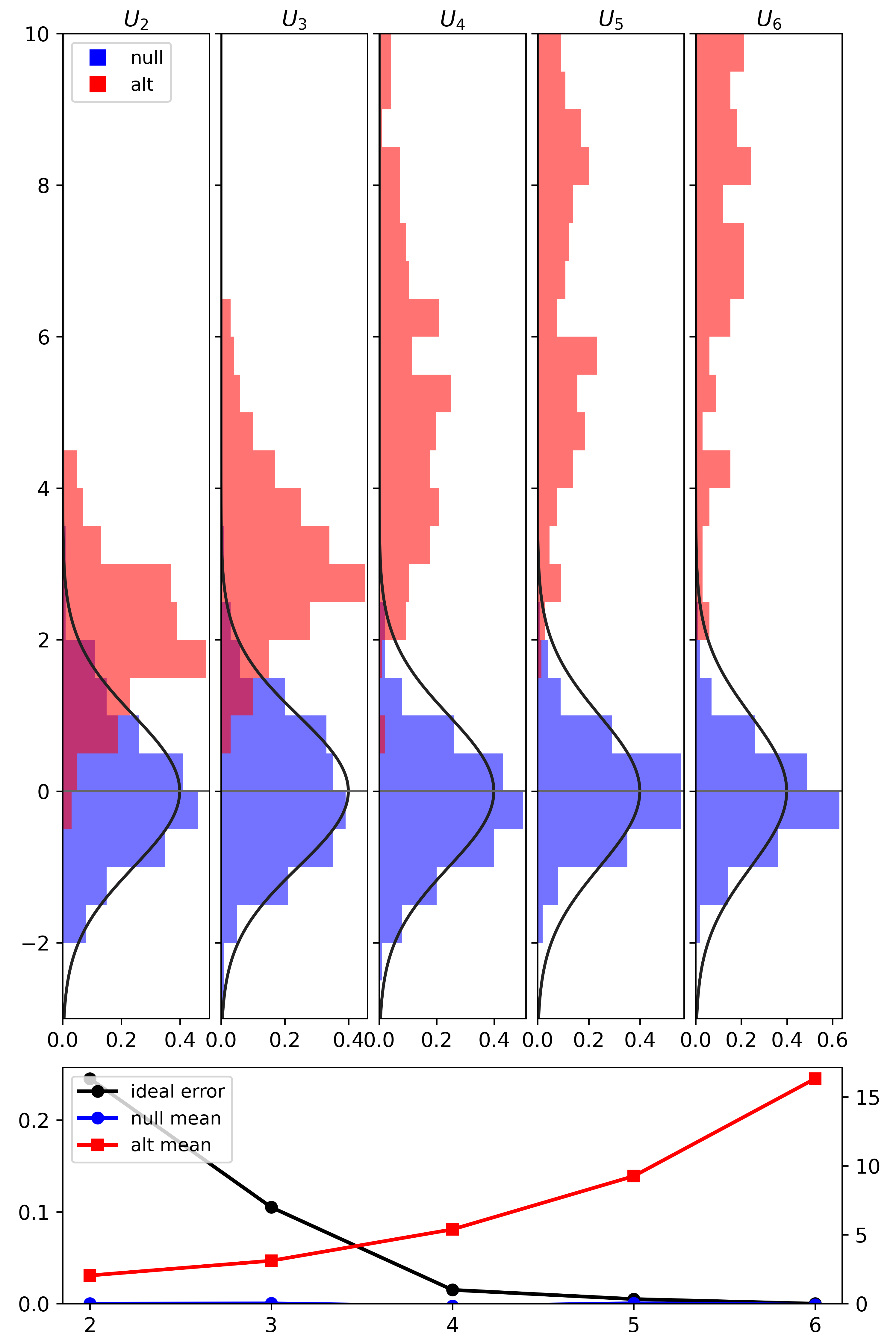}
\end{minipage}\hspace{1cm}
\begin{minipage}[t]{0.4\textwidth}
\vspace{0pt}
\centering
\includegraphics[width=\linewidth, height=1.35\linewidth]{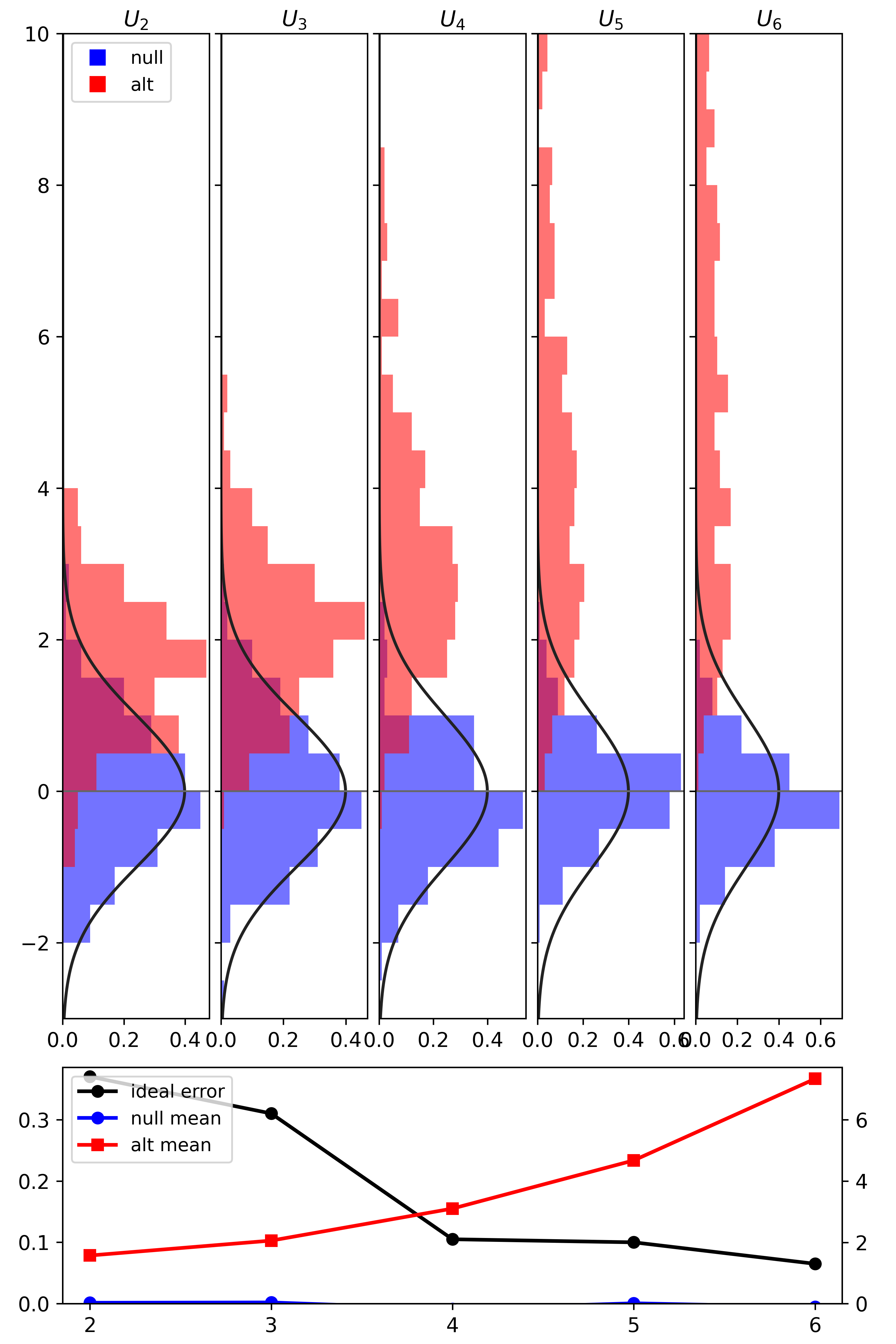}
\end{minipage}
\caption{Network pairwise comparison using cycle count statistics. %with $n=500$, $K=2$, $\beta_n=6$, Dirichlet memberships $\mathrm{Dir}(1.6,0.4)$ versus $\mathrm{Dir}(1,1)$, and $t=200$. 
Left: target SNR $=2$. Right: target SNR $=1.5$.}
\label{fig:pairwise-ibm}
\end{figure}

Figure~\ref{fig:pairwise-ibm} compares the resulting behavior at these two
signal levels. In both panels, the upper portion displays the empirical
distributions of the standardized statistics $U_2, U_3, \dots,U_6$ under the
null and alternative, together with the reference $N(0,1)$ law, while the
lower portion reports the corresponding ideal classification error and the
empirical means under the two hypotheses. The same qualitative pattern appears
in both settings: after standardization, the null distributions remain
concentrated around zero, whereas the alternative distributions move farther
and farther to the right as $r$ increases. Correspondingly, the ideal error
decreases with the cycle order in both SNR regimes, so the highest reported
orders produce the strongest separation. When the target SNR is reduced from
$2$ to $1.5$, the separation becomes weaker and the ideal errors are uniformly
larger, but the same monotone improvement with increasing order remains clearly
visible. The main takeaway is therefore that higher-order cycle statistics are
consistently more informative, even after the overall signal level is reduced.

\section{Discussion} \label{sec:Discu}
We have developed a general framework for deriving Computationally Efficient Equivalent Forms (CEEFs) for cycle count statistics. The resulting solution is built on new combinatorial concepts and techniques, including Fully Annotated Multigraphs (FAMs), Succinct Expressive Algebraic (SEA) terms, Incompressible Full Sum (IFS) terms, and the merging and pruning procedures developed in Section~\ref{sec:main}. These ideas do not follow directly from existing results in the literature \cite{harary1969graph, Chang2003, Movarraei2016, Jin2021optimal, Russian}; in particular, the notions of SEA terms, FAMs, and their associated full sums appear to be new.

A distinctive feature of this work is the way in which the mathematical framework was executed. For a technically demanding problem such as CEEF, current AI systems should not be expected to solve the problem independently through direct prompting. Instead, our human-guided AI (\emph{hugAI}) approach relies on humans to design the high-level reasoning and derivation strategy and decompose the overall task into many well-formulated subproblems. AI is then used to carry out these subproblems through coding, symbolic manipulation, and systematic enumeration. This division of labor is essential: the human contribution provides the theoretical structure, while AI helps execute the labor-intensive steps that would be impractical to complete manually.

This setting differs from many common uses of AI in theoretical research. Often, researchers present an AI system with a well-formulated problem arising from a particular step of a proof, and both the problem statement and the desired output are relatively concise. In the CEEF problem, by contrast, the output itself becomes increasingly large and intricate as the cycle length $m$ grows; for example, the formula for $m=12$ contains 1900 terms. This makes it unrealistic to expect current frontier AI systems to generate the full answer correctly in a single interaction. Our approach instead uses AI as an active assistant within a human-designed derivation strategy.

Our approach is also distinct from formal proof assistants such as \texttt{Lean}. In formal theorem proving, the mathematical ideas are typically developed by humans, while the proof assistant is used to formalize and verify logical correctness. In contrast, our framework uses AI to carry out nontrivial components of the derivation after the overall strategy has been established. Thus, AI contributes not merely to proof checking, but also to the execution of substantial symbolic and combinatorial tasks.

The validation of AI-generated mathematical output remains an important and largely open problem. Existing validation approaches are most effective when the final output or proof is sufficiently concise to permit manual inspection. In our setting, however, the CEEF formulas become too lengthy to verify manually in their entirety. To address this challenge, we developed a systematic validation pipeline that exploits the intrinsic discreteness of the solution space. Specifically, we validate the symbolic CEEF formulas by evaluating them on simulated continuous-valued data, where numerical agreement provides strong evidence for correctness.

Finally, the value of the CEEF problem extends beyond its combinatorial formulation. Cycle count statistics play an important role in network analysis, covariance matrix testing, and low-rank spectral inference. In Section~\ref{sec:applications}, we study three representative applications: spike detection, spiked eigenvalue estimation, and pairwise network comparison. Our simulations show that higher-order cycle count statistics can yield substantial gains in statistical performance. For the data scales considered in our experiments, brute-force enumeration is feasible only for small $m$, whereas the proposed CEEF formulas enable efficient computation of $C_m$ for $m$ as large as 12. These results highlight the statistical importance of efficient high-order cycle count computation and suggest several directions for future work.

\bibliographystyle{plain}
\bibliography{AI-refer}

%\newpage
	
\appendix 

\section{Proofs}\label{app:proofs}
	
In this section, we provide complete proofs for all lemmas and theorems presented in the main manuscript.
	
\subsection{Properties of the induced multi-graphs}
	
Recall that $S = \{i_1, i_2, \ldots, i_m\}$ and ${\cal G}$ is the undirected simple graph whose node set is $S$ and  there are edges between $i_1 \& i_2$, $i_2 \& i_3$, $\ldots$, $i_m \& i_1$, but nowhere else.  
	We use $\sigma = \{S_1, S_2, \ldots, S_k\}$ to denote a general size-$k$ partition on $S$, where $S_1,S_2,\ldots,S_k$ are non-overlapping subsets whose union is equal to $S$, and $k$ is the size of this partition, for which we often write $|\sigma|=k$. 
	As described in Section~\ref{subsec:merging}, each partition $\sigma$ induces a multi-graph
	${\cal G}_{\sigma}$: It modifies ${\cal G}$ by merging all nodes in each $S_j$ to one, while keeping the edges unchanged (note that ${\cal G}_{\sigma}$ has $|\sigma|$ nodes in this case). 
	
	\begin{definition} 
		We call ${\cal G}_{\sigma}$  non-trivial if it does not have self-loops, and we call a partition $\sigma$ non-trivial 
		if ${\cal G}_{\sigma}$ is non-trivial. 
	\end{definition} 
	
	The next lemma gives some useful properties for these induced ${\cal G}_\sigma$: 
	%%%%%%%%%%%%%%%
	%%%%%%%%%%%%%%%
	%%%%%%%%%%%%%%%
	\begin{lemma}\label{lem:graphrequirementdetails}
		A non-trivial induced multi-graph ${\cal G}_{\sigma}$ has the following properties: (a) It  is connected. (b) Its size satisfies that $k \geq 2$. (c) The degree of each node is even. 
	\end{lemma}
	
	{\it Proof of Lemma~\ref{lem:graphrequirementdetails}: } For (a), we recall that ${\cal G}_\sigma$ is obtained from merging nodes in ${\cal G}$. If a graph is connected before the merging process, it will also be connected afterwards. Since the original graph ${\cal G}$ is connected, each ${\cal G}_\sigma$ must also be connected. 
	For (b),  we note that a 1-node multi-graph can only have self-loops, which contradict the definition of a non-trivial multi-graph. 
	
	To prove (c),  we study how merging two nodes $a$ and $b$ in a multi-graph affects node degrees. Suppose the degree of each node is even before merging. For all nodes other than $a$ and $b$, their degrees are unchanged during merging. For $a$ and $b$, let $d_a$ and $d_b$ be their respective degrees before merging, and let 
	$w_{ab}$ be the number of edges between $a$ and $b$. Here, both $d_a$ and $d_b$ are even, and $w_{ab}$ can be either even or odd. 
	After these two nodes are merged, the degree of the new node $a\&b$ becomes 
	\[
	d^*=(d_a-w_{ab})+(d_b-w_{ab}), \qquad\mbox{which must be an even number}. 
	\]
	So far, we have shown the following: If each node has an even degree before merging, the same is true after merging. Therefore, the claim in (c) follows immediately by noting that each node has an even degree in the original graph ${\cal G}$. \qed

\subsection{Proof of Theorem \ref{thm:Cm}}\label{app:mainmaththm}
	
Write $S=\{i_1,i_2,\ldots,i_m\}$ and let $\Pi(S)$ be the collection of all partitions over $S$. 
For any $\sigma\in \Pi(S)$, let $\{j_1, j_2,\ldots, j_k\}$ be the nodes in the induced multi-graph ${\cal G}_\sigma$, and let $w(a,b)$ be the edge weight between any two nodes $j_a$ and $j_b$. Let $f_\sigma(A, j_1,j_2,\ldots, j_k):=\prod_{1\leq a<b\leq k}A_{j_aj_b}^{w(a,b)}$. We introduce two quantities: 
	\begin{equation}\label{main-proof-0}
		R^{\text{DS}}_\sigma: = 
		\sum_{\substack{1\leq j_1, j_2, \ldots, j_k\leq n\\\text{(distinct)}}} f_\sigma(A, j_1,j_2,\ldots, j_k), 
		\qquad 
		R^{\text{FS}}_\sigma: = 
		\sum_{1\leq j_1, j_2, \ldots, j_k\leq n} f_\sigma(A, j_1,j_2,\ldots, j_k). 
	\end{equation}
	We call them the (merged) $\sigma$-DS term and $\sigma$-FS term, respectively.

	Given partitions $\sigma$ and $\tau$, we use $\tau\preceq \sigma$ to denote that $\tau$ is obtained by splitting blocks in $\sigma$, and we use $\tau\prec \sigma$ to denote that $\tau\preceq\sigma$ but $\tau\neq \sigma$. 
	Our proof is based on the following key lemma:
	%%%%%%%%%%%%%%%%%%%%%%%%%%
	\begin{lemma} \label{lem:FStoDS}
		Let ${\cal D}(\sigma)$ be an arbitrary mapping that maps each $\sigma \in \Pi(S)$ to a real value. Then, 
		\[
		\sum_{\sigma\in \Pi(S)}{\cal D}(\sigma)R^{\text{FS}}_\sigma = \sum_{\sigma\in \Pi(S)} {\cal F}_1(\sigma) R^{\text{DS}}_\sigma, \qquad\mbox{with}\quad  {\cal F}_1(\sigma):=\sum_{\tau: \tau \preceq \sigma} {\cal D}(\tau), 
		\] 
		and 
		\[
		\sum_{\sigma\in \Pi(S)}{\cal D}(\sigma)R^{\text{DS}}_\sigma = \sum_{\sigma\in \Pi(S)} {\cal F}_2(\sigma) R^{\text{FS}}_\sigma, \quad\mbox{with}\quad {\cal F}_2(\sigma):=
		\begin{cases}
			{\cal D}(\sigma), &\mbox{if }\sigma=\hat{0}, \cr
			{\cal D}(\sigma ) -   \sum_{\tau: \tau \prec \sigma} {\cal F}_2(\tau), &\mbox{if }\sigma\neq \hat{0}. 
		\end{cases}
		\] 
	\end{lemma}
	
	{\it Proof of Lemma~\ref{lem:FStoDS}: } 
	Before proving this lemma, we argue that ${\cal F}_2(\cdot)$ is well-defined. 
	For each $\sigma\in \Pi(S)$, let $\mathrm{Child}(\sigma)=\{\tau: \tau\prec \sigma\}$. 
	We use $\Pi^*$ to denote the collection of $\sigma$ for which ${\cal F}_2(\sigma)$ has been defined. This set initially only contains one element, which is $\hat{0}$. Next, for any $\sigma$ such that $\mathrm{Child}(\sigma)\subset \Pi^*$, we can use the formula to obtain ${\cal F}_2(\sigma)$. We then add these $\sigma$ into $\Pi^*$. After $\Pi^*$ is updated, we search again those $\sigma$ such that $\mathrm{Child}(\sigma)\subset \Pi^*$, use the formula to obtain ${\cal F}_2(\sigma)$, and add them to $\Pi^*$. This procedure is repeated until every $\sigma$ is in $\Pi^*$. Therefore, ${\cal F}_2(\cdot)$ is well-defined.

	Consider the first claim. 
	Let $\tau_0=\hat{0}$. We introduce a mapping ${\cal M}$ that maps each $(i_1,i_2,\ldots,i_m)\in \{1,2,\ldots,n\}^m$ to a partition $\sigma$, depending on which of these $m$ indices are equal to each other (this mapping has also been used in \eqref{cm-1}, with a more careful explanation there). 
	We can easily see that
	\[
	R^{\text{FS}}_{\tau_0} = \sum_{1\leq i_1,\ldots,i_m\leq n} f_{\tau_0}(A, i_1,\ldots,i_m) = \sum_{\sigma\in \Pi(S)}\sum_{\substack{1\leq i_1,\ldots,i_m\leq n,\\ {\cal M}(i_1,\ldots,i_m)=\sigma}} f_{\tau_0}(A, i_1,\ldots,i_m). 
	\] 
	Additionally, in \eqref{cm-2}, we have proved the following result: For each $\sigma$, letting $j_1,j_2,\ldots,j_k$ denote the nodes in ${\cal G}_\sigma$, it holds that 
	\[
	\sum_{\substack{1\leq i_1,\ldots,i_m\leq n,\\ {\cal M}(i_1,\ldots,i_m)=\sigma}} f_{\tau_0}(A, i_1,\ldots,i_m) = \sum_{\substack{1\leq j_1, j_2, \ldots, j_k\leq n\\\text{(distinct)}}} f_\sigma(A, j_1,j_2,\ldots, j_k)=R^{\text{DS}}_\sigma. 
	\]
	Combining the above two equations and noticing that $\tau_0\preceq \sigma$ for all $\sigma\in \Pi(S)$, we obtain:
	\[
	R^{\text{FS}}_{\tau_0} =\sum_{\sigma\in \Pi(S)} R_\sigma^{\text{DS}} =  \sum_{\sigma: \tau_0\preceq \sigma}R_\sigma^{\text{DS}}, \qquad\mbox{for }\tau_0 = \hat{0}. 
	\]
	For $\tau\neq \tau_0$, we can mimic the above proof and consider all partitions over the nodes in ${\cal G}_\tau$. We can similarly show that
	\begin{equation} \label{FStoDS-1}
		R^{\text{FS}}_{\tau} =\sum_{\sigma: \tau\preceq \sigma}R_\sigma^{\text{DS}}, \qquad\mbox{for all }\tau\in \Pi(S).  
	\end{equation}
	It follows by \eqref{FStoDS-1} that
	\[
	\sum_{\tau\in \Pi(S)}{\cal D}(\tau)R^{\text{FS}}_\tau = \sum_{\tau\in \Pi(S)}{\cal D}(\tau) \Bigl(\sum_{\sigma: \tau\preceq \sigma}R_\sigma^{\text{DS}}\Bigr) = \sum_{\sigma\in \Pi(S)}\Bigl[   \sum_{\tau: \tau \preceq \sigma} {\cal D}(\tau) \Bigr] R^{\text{DS}}_\sigma. 
	\]
	This proves the first claim. 
	
	Consider the second claim. When there exists a mapping ${\cal F}_2(\sigma)$ such that $\sum_{\sigma\in \Pi(S)} {\cal F}_2(\sigma) R^{\text{FS}}_\sigma=\sum_{\sigma\in \Pi(S)}{\cal D}(\sigma)R^{\text{DS}}_\sigma$, we can apply the first claim of this lemma to conclude that 
	\begin{equation} \label{FStoDS-2}
		{\cal D}(\sigma)=\sum_{\tau: \tau\preceq \sigma} {\cal F}_2(\tau), \qquad\mbox{for all }\sigma\in \Pi(S). 
	\end{equation}
	We can rewrite \eqref{FStoDS-2} as  ${\cal D}(\sigma) =  {\cal F}_2(\sigma) + \sum_{\tau: \tau \prec \sigma}{\cal F}_2(\tau)$, which implies that 
	\[
	{\cal F}_2(\sigma) = {\cal D}(\sigma) - \sum_{\tau: \tau \prec \sigma}{\cal F}_2(\tau), \qquad\mbox{for all }\sigma\in \Pi(S).  
	\]
	This proves the second claim of Lemma~\ref{lem:FStoDS}. \qed

	We now proceed to prove Theorem~\ref{thm:Cm}. In the special case where $\sigma=\hat{0}$ (the finest partition), the induced ${\cal G}_\sigma$ has $m$ nodes and a single edge between each of $i_1\& i_2$, $i_2\&i_3$, \ldots, $i_m\& i_1$. It follows that $f_{\hat{0}}(A, j_1,j_2,\ldots, j_k)=A_{i_1i_2}A_{i_2i_3}\ldots A_{i_mi_1}$. As a result, 
	\[
	C_m = \sum_{\substack{1\leq i_1, i_2, \ldots, i_m\leq n\\\text{(distinct)}}} f_{\hat{0}}(A, i_1,i_2,\ldots,i_m)=R^{\text{DS}}_{\hat{0}}. 
	\]
	We can write it equivalently as
	\begin{equation} \label{main-proof-1}
		C_m = \sum_{\sigma \in \Pi(S)} \mathcal{D}(\sigma)  R^{\text{DS}}_\sigma, \qquad\mbox{with}\quad \mathcal{D}(\sigma):=
		\begin{cases}
			1 & \text{if } \sigma = \hat{0}, \\
			0 & \text{otherwise}.\\
		\end{cases}
	\end{equation}
	Using the second claim of Lemma~\ref{lem:FStoDS}, we immediately obtain that
	\begin{equation} \label{main-proof-2}
		C_m = \sum_{\sigma \in \Pi(S)} \mathcal{F}(\sigma)  R^{\text{FS}}_\sigma, \qquad\mbox{with}\quad \mathcal{F}(\sigma)=
		\begin{cases}
			1 & \text{if } \sigma = \hat{0}, \\
			-\sum\limits_{\hat{0}\preceq \tau \prec \sigma}\mathcal{F}(\tau) & \text{if } \sigma \neq \hat{0}.\\
		\end{cases}
	\end{equation}
\begin{definition}[The M{\"o}bius function]\label{def:Mobius}
For any $\pi, \sigma \in \Pi(S)$ with $\pi \prec \sigma$, $\mu(\pi,\pi)=1$, and $\mu(\pi,\sigma)$ is defined recursively by
\[
\mu(\pi,\sigma)=-\sum_{\pi\preceq\tau\prec\sigma}\mu(\pi,\tau).
\]
\end{definition}

Comparing ${\cal F}(\sigma)$ in \eqref{main-proof-2} with Definition~\ref{def:Mobius}, we conclude that 
\begin{equation} \label{main-proof-3}
{\cal F}(\sigma) = \mu(\hat{0}, \sigma), \qquad \mbox{for }\sigma\in \Pi(S). 
\end{equation}
The M{\"o}bius function $\mu(\hat{0},\sigma)$ has an explicit expression, which is well-known in the literature and presented in the following lemma (e.g., see  \cite{Stanley2011} for a proof): 
%%%%%%%%%
\begin{lemma}[Explicit formula of Möbius functions on partition lattices] \label{lem:Mobius}
	For any partition $\sigma$ with $k$ blocks $\{S_1, S_2, \ldots, S_k\}$ of $\{1,2,\ldots,m\}$, we have $\mu(\hat{0}, \sigma) = (-1)^{m-k} \prod_{i=1}^k (|S_i|-1)!$. 
\end{lemma}

	We plug \eqref{main-proof-3} into \eqref{main-proof-2} and let ${\cal G}_{m,k,t}$ be the same isomorphic class as in \eqref{cm-6}. It yields that 
	\begin{equation}\label{main-proof-4}
		C_m = \sum_{\sigma \in \Pi(S)} \mu(\hat{0},\sigma)R^{\text{FS}}_\sigma = \sum_{k=2}^{m}\sum_{t=1}^{b_{m,k}}\sum_{\sigma:\, {\cal G}_\sigma\in {\cal G}_{m,k,t}}\mu(\hat{0},\sigma)R^{\text{FS}}_\sigma,
	\end{equation}
	For partitions $\sigma$ and $\tilde{\sigma}$ such that their corresponding multi-graphs are isomorphic, we have  $R_\sigma^{\text{FS}}=R_{\tilde{\sigma}}^{\text{FS}}$ (the derivation for this equality is similar to that for \eqref{cm-5}, except for a minor difference that the distinct sum there is replaced by the full sum here).  
	Additionally, from the expression in Lemma~\ref{lem:Mobius}, we also observe that $\mu(\hat{0}, \sigma)=\mu(\hat{0}, \tilde{\sigma})$. 
	Let $\sigma_{m,k,t}$ be the canonical partition in ${\cal G}_{m,k,t}$. It is seen that
	\[
	\mu(\hat{0},\sigma)R^{\text{FS}}_\sigma = \mu(\hat{0},\sigma_{m,k,t})R^{\text{FS}}_{\sigma_{m,k,t}}, \qquad\mbox{for all $\sigma$ such that } {\cal G}_\sigma \in {\cal G}_{m,k,t}. 
	\]
	Recall that $d_{m,k,t}=|{\cal G}_{m,k,t}|$. We plug the above equality into \eqref{main-proof-4} to obtain:
	\begin{equation}\label{main-proof-5}
		C_m = \sum_{k=2}^{m-1}\sum_{t=1}^{b_{m,k}}d_{m,k,t}\cdot \mu\bigl(\hat{0},\sigma_{m,k,t}\bigr)\cdot  R^{\text{FS}}_{\sigma_{m,k,t}}. 
	\end{equation}
	Finally, we let $f_{m,k,t}(A, j_1,j_2,\ldots, j_k):=f_{\sigma_{m,k,t}}(A, j_1, j_2,\ldots, j_k)$ and use the expression of $R_\sigma^{\text{FS}}$ in \eqref{main-proof-3}. It follows that
	\[
	R^{\text{FS}}_{\sigma_{m,k,t}} = \sum_{1\leq j_1,j_2,\ldots,j_k\leq n}f_{m,k,t}(A, j_1,j_2,\ldots, j_k). 
	\]
	Let $a_{m,k,t}: = d_{m,k,t} \cdot \mu\bigl(\hat{0}, \sigma_{m,k,t}\bigr)$. 
	We plug the above equality into \eqref{main-proof-5} to get
	\begin{equation} \label{main-proof-6}
		C_m = \sum_{k=2}^{m-1} \sum_{t=1}^{b_{m,k}} a_{m,k,t} \sum_{j_1, j_2, \ldots, j_k} f_{m,k,t}(A, j_1, j_2, \ldots, j_k).   
	\end{equation}
	Let $h_{m,k,t} = \prod_{i=1}^k (|S_i| - 1)!$ as defined in (\ref{def:h}), with $\{S_1, S_2, \ldots, S_k\}$ being the blocks of partition $\sigma_{m,k,t}$. It follows by Lemma~\ref{lem:Mobius} that $\mu(\hat{0}, \sigma_{m,k,t}) = (-1)^{m-k} \cdot h_{m,k,t}$. It follows that 
	\begin{equation} \label{main-proof-7}
		a_{m,k,t} = d_{m,k,t} \cdot (-1)^{m-k} \cdot h_{m,k,t}. 
	\end{equation}
	The claim follows by combining \eqref{main-proof-7} with \eqref{main-proof-6}. \qed

\subsection{Proof of Corollary~\ref{cor:Cm}}\label{app:permgraph}
	
Recall that $C_m=\sum_{i_1,\ldots,i_m (\text{distinct})}A_{i_1 i_2} A_{i_2 i_3} \ldots A_{i_m i_1}$. Without the requirement of having distinct indices, we can easily deduce $\sum_{i_1,\ldots,i_m}A_{i_1 i_2} A_{i_2 i_3} \ldots A_{i_m i_1}=\mathrm{tr}(A^m)$. 
	As a result, 
	\begin{equation}\label{cm-0}
		C_m = \mathrm{tr}(A^m) - R_m, \qquad\mbox{with}\quad R_m: = \sum_{\substack{1\leq i_1, i_2, \ldots, i_m\leq n\\\text{(not distinct)}}} A_{i_1 i_2} A_{i_2 i_3} \ldots A_{i_m i_1}. 
	\end{equation}
	We study $R_m$. When $i_1,\ldots,i_m$ are non-distinct, there can be $k\leq m-1$ distinct indices. When $k=1$, since the diagonal entries of $A$ are zero, we have $A_{i_1 i_2} A_{i_2 i_3} \ldots A_{i_m i_1}=0$. It suffices to consider $2\leq k\leq m-1$, and we can write
	\[
	R_m =\sum_{k=2}^{m-1}\; \sum_{\substack{1\leq i_1, i_2, \ldots, i_m\leq n\\|\{i_1,i_2,\ldots,i_m\}|=k}} A_{i_1 i_2} A_{i_2 i_3} \ldots A_{i_m i_1}. 
	\]
	When $|\{i_1,i_2,\ldots,i_m\}|=k$, these $m$ indices are divided into $k$ non-overlapping sets and correspond to a size-$k$ partition. To this end,  we define a mapping ${\cal M}$ that maps each $(i_1,i_2,\ldots,i_m)$ to a partition $\sigma$. Let $\Pi(S)$ denote the collection of all partitions with $2\leq |\sigma|\leq m-1$. We can further write
	\begin{equation} \label{cm-1}
		R_m =\sum_{k=2}^{m-1}\sum_{\sigma\in \Pi(S): |\sigma|=k}\; \sum_{\substack{1\leq i_1, i_2, \ldots, i_m\leq n\\{\cal M}(i_1,i_2,\ldots,i_m)=\sigma}} A_{i_1 i_2} A_{i_2 i_3} \ldots A_{i_m i_1}. 
	\end{equation}
	For each $\sigma$ with $|\sigma|=k$, let $j_1, j_2,\ldots,j_k$ denote the nodes in the corresponding multi-graph ${\cal G}_\sigma$. 
	When ${\cal M}(i_1,i_2,\ldots,i_m)=\sigma$, we map each of $i_1,i_2, \ldots,i_m$ to one of $j_1,j_2,\ldots,j_k$. 
	For example, when $m=4$ and $\sigma=\{\{i_1,i_3\}, \{i_2\}, \{i_4\}\}$, the original indices $i_1, i_2, i_3, i_4$ are mapped to the new indices $j_1, j_2, j_1, j_3$, respectively. Consequently, 
	\[
	A_{i_1 i_2} A_{i_2 i_3}A_{i_3i_4}A_{i_4i_1}=A_{j_1j_2}A_{j_2j_1}A_{j_1j_3}A_{j_3j_1}= A^2_{j_1j_2}A_{j_1j_3}^2. 
	\]
	More generally, let $w(a,b)$ denote the edge weight between nodes $j_a$ and $j_b$ in ${\cal G}_\sigma$. We observe that each $A_{j_aj_b}$ is repeated $w(a,b)$ times in $A_{i_1 i_2} A_{i_2 i_3} \ldots A_{i_m i_1}$, which implies that  
	\[
	A_{i_1 i_2} A_{i_2 i_3} \ldots A_{i_m i_1} = \prod_{1\leq a<b\leq k}A_{j_aj_b}^{w(a,b)}, \qquad\mbox{when }{\cal M}(i_1,i_2,\ldots,i_m)=\sigma. 
	\]
	Furthermore, when we sum over $1\leq i_1,\ldots,i_m\leq n$ under the constraint that ${\cal M}(i_1,\ldots,i_m)=\sigma$, it is equivalent to summing over $1\leq j_1,\ldots, j_k\leq n$ under the constraint that $j_1,\ldots,j_k$ are distinct. 
	Combining the above arguments gives 
	\begin{equation} \label{cm-2}
		\sum_{\substack{1\leq i_1, i_2, \ldots, i_m\leq n\\{\cal M}(i_1,i_2,\ldots,i_m)=\sigma}} A_{i_1 i_2} A_{i_2 i_3} \ldots A_{i_m i_1} = \sum_{\substack{1\leq j_1, j_2, \ldots, j_k\leq n\\\text{(distinct)}}}\biggl( \prod_{1\leq a<b\leq k}A_{j_aj_b}^{w(a,b)}\biggr).  
	\end{equation}
	We plug \eqref{cm-2} into \eqref{cm-1} to obtain: 
	\begin{equation} \label{cm-3}
		R_m =\sum_{k = 2}^{m-1} \; \sum_{\sigma \in \Pi(S): |\sigma| = k} \;  \sum_{\substack{1\leq j_1, j_2, \ldots, j_k\leq n\\\text{(distinct)}}} \Bigl(\prod_{1\leq a<b\leq k}A_{j_aj_b}^{w(a,b)} \Bigr). 
	\end{equation}
	
	Write $f_\sigma(A, j_1,j_2,\ldots, j_k):=\prod_{1\leq a<b\leq k}A_{j_aj_b}^{w(a,b)}$ for brevity. 
	We can express \eqref{cm-3} equivalently as
	\[
	R_m = \sum_{k = 2}^{m-1} \; \sum_{\sigma \in \Pi(S): |\sigma| = k} R_{\sigma}, \qquad \mbox{with}\quad R_\sigma: = 
	\sum_{\substack{1\leq j_1, j_2, \ldots, j_k\leq n\\\text{(distinct)}}} f_\sigma(A, j_1,j_2,\ldots, j_k). 
	\]
	We group all multi-graphs ${\cal G}_\sigma$ first by size and then by isomorphism. Let $b_{m,k}$ be the total number of isomorphic classes for size-$k$ multi-graphs, and let
	${\cal G}_{m, k, t}$ denote the $t$-th isomorphic class, for $1\leq t\leq b_{m,k}$. It follows that 
	\begin{equation} \label{cm-4}
		R_m = \sum_{k = 2}^{m-1} \sum_{t=1}^{b_{m,k}}\sum_{\sigma:\;  {\cal G}_\sigma\in {\cal G}_{m,k,t}}  R_{\sigma}. 
	\end{equation}

Consider two partitions $\sigma$  and $\tilde{\sigma}$ such that ${\cal G}_\sigma$ and ${\cal G}_{\tilde{\sigma}}$ are  isomorphic multi-graphs. The definition of isomorphism implies that there exists a permutation ${\cal P}$ on $\{j_1,j_2,\ldots,j_k\}$ such that 
	\[
	f_{\tilde{\sigma}} (A,  j_1,j_2,\ldots, j_k) = f_\sigma\bigl(A,  {\cal P}(j_1), {\cal P}(j_2),\ldots, {\cal P}(j_k)\bigr). 
	\]
	When we sum over $1\leq j_1,\ldots, j_k\leq n$ and require them to be distinct, the above permutation does not matter. As a result, 
	\begin{equation} \label{cm-5}
		R_\sigma= R_{\tilde{\sigma}}, \qquad \mbox{whenever ${\cal G}_\sigma$ and ${\cal G}_{\tilde{\sigma}}$ are isomorphic}. 
	\end{equation}
	To this end, we pick the canonical partition  $\sigma_{m,k,t}\in {\cal G}_{m,k,t}$ and define $f_{m, k, t}(A, j_1, j_2, \ldots, j_k):=f_{\sigma_{m,k,t}}(A, j_1, j_2,\ldots, j_k)$. Combining \eqref{cm-5} with \eqref{cm-4}, we obtain:
	\begin{align} \label{cm-6}
		R_m & = \sum_{k = 2}^{m-1} \sum_{t=1}^{b_{m,k}} d_{m,k,t} \cdot  R_{\sigma_{k,m,t}}, \qquad \mbox{where}\quad d_{m,k,t}= |{\cal G}_{m,k,t}|,\cr
		& = \sum_{k = 2}^{m-1} \sum_{t=1}^{b_{m,k}} d_{m,k,t} \sum_{\substack{1\leq j_1, j_2, \ldots, j_k\leq n\\\text{(distinct)}}} f_{m,k,t}(A, j_1, j_2,\ldots, j_k).  
	\end{align}
	Plugging \eqref{cm-6} into \eqref{cm-0}, we obtain:
	\begin{equation}\label{cm-7}
		C_m=\mathrm{tr}(A^m)-\sum_{k = 2}^{m-1} \sum_{t=1}^{b_{m,k}} d_{m,k,t} \sum_{\substack{1\leq j_1, j_2, \ldots, j_k\leq n\\\text{(distinct)}}} f_{m,k,t}(A, j_1, j_2,\ldots, j_k),
	\end{equation}
	which completes the proof.
	\qed

\subsection{Proof of Lemma~\ref{lemma:LMG}}
	
	It is sufficient to show that the value of the Full-Sum remains the same before and 
	after either types of the pruning. 
	
	Consider the Type I pruning first. Without loss of generality, suppose node 1 is the Pendant and node 2 is the Hinge. Let $v$ and $u$ be their respective node labels, and $M^{(1)}, M^{(2)}, \ldots, M^{(s)}$ be the labels of the edges between them (after orientation alignment).
	We can express the Full-Sum of the original FAM $\mathcal{G}$ as:
	\begin{equation*}
		\begin{aligned}
			\mathrm{FS}(\mathcal{G}) &= \sum_{j_1, j_2, \ldots, j_k} f^{(1,2)}(A, j_3, \ldots, j_k) \cdot u_{j_2} \cdot \prod_{l=1}^{s} M^{(l)}_{j_2 j_1} \cdot v_{j_1}\\
			&= \sum_{j_2, \ldots, j_k} f^{(1,2)}(A, j_3, \ldots, j_k) \cdot u_{j_2} \cdot \sum_{j_1} \left(\prod_{l=1}^{s} M^{(l)}_{j_2 j_1}\right) \cdot v_{j_1}
		\end{aligned}
	\end{equation*}	
	where $f^{(1,2)}(A, j_3, \ldots, j_k)$ captures all other terms in the product that are independent of $j_1$ and $j_2$.
	
	Let $M = M^{(1)} \circ M^{(2)} \circ \cdots \circ M^{(s)}$ denote the Hadamard product of edge matrices. Then:
	\begin{equation}
		\begin{aligned}
			\mathrm{FS}(\mathcal{G}) &= \sum_{j_2, \ldots, j_k} f^{(1,2)}(A, j_3, \ldots, j_k) \cdot u_{j_2} \cdot \sum_{j_1} M_{j_2 j_1} \cdot v_{j_1}\\
			&= \sum_{j_2, \ldots, j_k} f^{(1,2)}(A, j_3, \ldots, j_k) \cdot [u \circ (M \cdot v)]_{j_2}
		\end{aligned}
	\end{equation}
	We can see that after pruning, the new FAM $\mathcal{G}^{\text{new}}$ has node 1 removed and the hinge node 2 now has an updated label $u^{\text{new}} = u \circ (M \cdot v)$. Its Full-Sum is:
	\begin{equation}
		\begin{aligned}
			\mathrm{FS}(\mathcal{G}^{\text{new}}) &= \sum_{j_2, \ldots, j_k} f^{(1,2)}(A, j_3, \ldots, j_k) \cdot u^{\text{new}}_{j_2}\\
			&= \sum_{j_2, \ldots, j_k} f^{(1,2)}(A, j_3, \ldots, j_k) \cdot [u \circ (M \cdot v)]_{j_2}
		\end{aligned}
	\end{equation}
	Therefore, $\mathrm{FS}(\mathcal{G}) = \mathrm{FS}(\mathcal{G}^{\text{new}})$.
	
	Consider the Type II pruning. Without loss of generality, suppose node 1 is the Pendant, and nodes 2 and 3 are its Hinges. Let $y$, $x$, and $z$ be their respective node labels. Let $Q^{(1)}, \ldots, Q^{(s)}$ be the labels of edges between nodes 1 and 2, and $R^{(1)}, \ldots, R^{(t)}$ be the labels of edges between nodes 1 and 3, both after orientation alignment. 
	
	We can express the Full-Sum of the original FAM $\mathcal{G}$ as:
	$$\begin{aligned}
		\mathrm{FS}(\mathcal{G}) &= \sum_{j_1, \ldots, j_k} f^{(1,2,3)}(A, j_4, \ldots, j_k) \cdot x_{j_2} \cdot \left(\prod_{l=1}^{s} Q^{(l)}_{j_1 j_2}\right) \cdot y_{j_1} \cdot \left(\prod_{l=1}^{t} R^{(l)}_{j_1 j_3}\right) \cdot z_{j_3}
	\end{aligned}$$
	
	where $f^{(1,2,3)}(A, j_4, \ldots, j_k)$ captures all other terms independent of $j_1$, $j_2$, and $j_3$.
	
	Let $Q = Q^{(1)} \circ \cdots \circ Q^{(s)}$ and $R = R^{(1)} \circ \cdots \circ R^{(t)}$. Then:
	\begin{equation}
		\begin{aligned}
			\mathrm{FS}(\mathcal{G}) &= \sum_{j_2, j_3, \ldots, j_k} f^{(1,2,3)}(A, j_4, \ldots, j_k) \cdot x_{j_2} \cdot z_{j_3} \cdot \sum_{j_1} Q_{j_1 j_2} \cdot y_{j_1} \cdot R_{j_1 j_3}\\
			&= \sum_{j_2, j_3, \ldots, j_k} f^{(1,2,3)}(A, j_4, \ldots, j_k) \cdot x_{j_2} \cdot z_{j_3} \cdot [Q^{\top} \cdot d(y) \cdot R]_{j_2 j_3}
		\end{aligned}
	\end{equation}
	where $d(y)$ is the diagonal matrix with entries $y_{j_1}$ on the diagonal.
	
	After pruning, the new FAM $\mathcal{G}^{\text{new}}$ has node 1 removed and a new edge with label $E = Q^{\top} \cdot d(y) \cdot R$ between nodes 2 and 3. Its Full-Sum is:
	\begin{equation}
		\begin{aligned}
			\mathrm{FS}(\mathcal{G}^{\text{new}}) &= \sum_{j_2, j_3, \ldots, j_k} f^{(1,2,3)}(A, j_4, \ldots, j_k) \cdot x_{j_2} \cdot E_{j_2 j_3} \cdot z_{j_3}\\
			&= \sum_{j_2, j_3, \ldots, j_k} f^{(1,2,3)}(A, j_4, \ldots, j_k) \cdot x_{j_2} \cdot [Q^{\top} \cdot d(y) \cdot R]_{j_2 j_3} \cdot z_{j_3}
		\end{aligned}
	\end{equation}
	Therefore, we also have $\mathrm{FS}(\mathcal{G}) = \mathrm{FS}(\mathcal{G}^{\text{new}})$.
	
	In both cases, the pruning operation reduces the number of nodes by exactly 1 while preserving the value of the Full-Sum. This completes the proof. \qed

\subsection{Proof of Theorem~\ref{thm:prune}}
	Before we prove Theorem~\ref{thm:prune}, we show that the pruning algorithm eventually stops.
	
	Consider each pruning operation. Since we remove exactly 1 node in each pruning operation, the number of nodes reduces by exactly 1 before and after each pruning operation.
	Consider the termination condition. The algorithm terminates either when only one node remains or no prunable nodes (pendants) remain. In either case, when the algorithm terminates, there is at least one node remaining.
	Therefore, since the algorithm starts with a finite number of nodes $k$, we conclude that the algorithm must terminate after at most $(k-1)$ pruning steps, which shows that the pruning algorithm eventually stops. 
	
	We now proceed to prove Theorem~\ref{thm:prune}. Consider the first case. Let $v$ be the node label for the only one node remaining in the final FAM. 
	There is no other node for the only remaining node to connect to, therefore there is no edge in the final FAM. 
	Hence, the Full-Sum simplifies to:
	\begin{align}
		\mathrm{FS}(\mathcal{G}) &= \sum_{j_1} v_{j_1}= {\bf 1}_n' v
	\end{align}
	
	This is a SEA term by definition, since it is expressible in a succinct algebraic form without any remaining sums over indices.
	
	Consider the second case. In this case, no pendant node of either type is left. First we state a useful lemma.
	
	\begin{lemma}\label{lem:connectivityofLMG}
		The pruning algorithm preserves the connectivity of the FAM. That is, if a FAM is connected, then after one step of either Type I pruning or Type II pruning, the pruned FAM is still connected.
	\end{lemma}
	
	{\it Proof of Lemma~\ref{lem:connectivityofLMG}: } 
	Consider Type I pruning. Type I pruning deletes a node with 1 immediate neighbor, which does not affect the connectivity of the remaining nodes. This shows that after one step of Type I pruning, the pruned FAM is still connected.
	
	Consider Type II pruning. Type II pruning deletes a node with 2 immediate neighbors, which does not affect the connectivity of the remaining nodes. This shows that after one step of Type II pruning, the pruned FAM is still connected. \qed
	
	We now proceed to prove Theorem~\ref{thm:prune}.
	We first show that each of the remaining node in the final FAM has at least 3 immediate neighbors. We prove this by contradiction.
	Assume that there exists a node with fewer than 3 immediate neighbors. 
	If it has 2 neighbors, it would be a Type II pendant, contradicting our assumption that no pendant nodes remain.
	If it has 1 neighbor, it would be a Type I pendant, contradicting our assumption that no pendant nodes remain. 
	If it has 0 neighbor, then the FAM would be disconnected. 
	By Lemma~\ref{lem:graphrequirementdetails}, the original multi-graph is connected. 
	By Lemma~\ref{lem:connectivityofLMG}, the pruning process preserves connectivity of the FAM.
	Therefore, when the algorithm terminates, the final FAM must be connected. This leads to a contradiction.
	Therefore, we conclude that each of the remaining node in the final FAM has at least 3 immediate neighbors.

	We then show that the final FAM has at least 4 nodes. Pick a node in the final FAM. Using the above argument, it has at least 3 immediate neighbors. 
	Consider the picked node and three of its immediate neighbors in the final FAM. These are 4 different nodes in the final FAM, therefore we conclude that there must be at least 4 nodes remaining in the final FAM.

	Finally, we show that the FS term has at most $\lfloor m/2 \rfloor$ layers of sum. We call a node in the original FAM prunable, if it does not remain in the final FAM, i.e. it is pruned during a step of pruning operation. 
	Denote $k$ the number of nodes in the original FAM, denote $p$ the number of prunable nodes in the original FAM, and denote $d_2$ the number of degree $2$ nodes in the original FAM.
	
	Consider the number of prunable nodes in the original FAM. Note that every node with degree $2$ must be prunable, since they can have no more than $2$ neighbors. Therefore, we conclude that
	\begin{equation}\label{eq:degreerelation1}
		p \geq d_2.
	\end{equation}
	Consider the sum of degrees of all nodes in the original FAM. Since the original FAM has $m$ edges, the sum of degrees of all nodes in the original FAM is $2m$.
	By Lemma~\ref{lem:graphrequirementdetails}, the original FAM is connected, and every node in the original FAM has even degrees. 
	This means that the degree of each node in the original FAM can only be positive even numbers. Therefore, any node whose degree is not $2$ has a degree of at least $4$. 
	Therefore, the sum of degrees of all nodes in the original FAM is no less than $2d_2+4(k-d_2)$. Hence, we have the following inequality:
	\begin{equation}\label{eq:degreerelation2}
		2m \geq 2d_2+4(k-d_2).
	\end{equation}
	
	By \eqref{eq:degreerelation2}, we conclude that $4(k-d_2)\leq 2m$. Therefore, by \eqref{eq:degreerelation1}, we have 
	\begin{equation}   
		k-p \leq k-d_2\leq m/2. 
	\end{equation}
	Since the number of remaining nodes in the final FAM must be an integer, we conclude that the number of remaining nodes in the final FAM is no greater than $\lfloor m/2 \rfloor$. Since each node in the final FAM corresponds to one layer of summation, we conclude that the FS term has at most $\lfloor m/2 \rfloor$ layers of sum. This completes the proof. \qed

\section{Details of The LLMs under investigation}
While we primarily use DeepSeek-R1 in our hugAI approach, we have also tested 11 other models. Their relevant information is summarized in Table~\ref{tab:11LLMs}. 
	
\begin{table}[htbp]
\centering
\caption{The 11 LLMs under comparison.  All models are queried in their ``reasoning'' or ``thinking'' mode where available.} \label{tab:11LLMs}
		\label{tab:models}
		\scalebox{1.1}{
			\begin{tabular}{@{}llll@{}}
				\toprule
				\textbf{Model} & \textbf{Provider} & \textbf{Reasoning mode} & \textbf{API identifier} \\
				\midrule
				Claude Opus~4.8     & Anthropic  & Extended thinking  & \texttt{anthropic/claude-opus-4.8} \\
				Claude Sonnet~4.6   & Anthropic  & Extended thinking  & \texttt{anthropic/claude-sonnet-4.6} \\
				DeepSeek~V4 Pro     & DeepSeek   & High effort        & \texttt{deepseek/deepseek-v4-pro} \\
				GPT-5.5             & OpenAI     & High effort        & \texttt{openai/gpt-5.5} \\
				Gemini~3.1 Pro      & Google     & Thinking mode      & \texttt{google/gemini-3.1-pro-preview} \\
				Gemini~3.5 Flash    & Google     & Thinking mode      & \texttt{google/gemini-3.5-flash} \\
				GLM~5.1             & Zhipu AI   & Default            & \texttt{z-ai/glm-5.1} \\
				Kimi~K2.6           & Moonshot   & Default            & \texttt{moonshotai/kimi-k2.6} \\
				MiniMax-M3          & MiniMax    & Default            & \texttt{minimax/minimax-m3} \\
				MiniMax-M2.7        & MiniMax    & Default            & \texttt{minimax/minimax-m2.7} \\
				DeepSeek~V3.2       & DeepSeek   & High effort        & \texttt{deepseek/deepseek-v3.2} \\
				\bottomrule
		\end{tabular}}
	\end{table}

\section{Prompts} \label{app:prompt}
		
		The complete two-stage prompt used in the evaluation is reproduced below; see Figures~\ref{fig:prompt-stage1}-\ref{fig:prompt-stage2}.
		Stage~1 is sent first; after structural verification, Stage~2 is sent with
		the model's own Stage~1 code included as context. 
		%The prompt contains no paper citations, author names, or searchable identifiers.
		
		% ====== FIGURE* 1 : System + Stage 1 ======
		\begin{figure*}[htb!]
			
			\textbf{System Prompt (Shared by Both Stages)}
			
			\begin{promptbox}
\begingroup\setlength{\parskip}{0pt}\linespread{0.96}\selectfont\obeylines\obeyspaces%
You are a mathematician implementing a combinatorial algorithm. Your task is to derive an explicit formula for a cycle count statistic and output it as executable Python code. Follow the algorithm described below precisely, and produce code output as instructed.
\endgroup
			\end{promptbox}
			
			%\vspace{15pt}
			\textbf{Stage 1 User Prompt: Merging Process}
			
			\begin{promptbox}
\begingroup\setlength{\parskip}{0pt}\linespread{0.96}\selectfont\obeylines\obeyspaces%
PROBLEM. For a symmetric matrix A in R\^\{n x n\} with zero diagonal (A\_\{ii\} = 0), the order-m cycle count statistic is:

    C\_m(A) = sum\_\{i\_1,...,i\_m all distinct\} A\_\{i\_1,i\_2\} * A\_\{i\_2,i\_3\} * ... * A\_\{i\_m,i\_1\}

The brute-force cost is O(n\^m). We want a Computationally Efficient Equivalent Form (CEEF): a formula expressing C\_m as a linear combination of terms, each computable in polynomial time using matrix operations.

YOUR TASK (Step 1 of 2): Implement the MERGING PROCESS that decomposes C\_m into canonical multi-graph types with aggregated Mobius coefficients. Your code must accept m as a parameter and work for any m \textgreater{}= 2.

STEP 1: MERGING PROCESS (Partitions -\textgreater{} Multi-graphs -\textgreater{} Canonical Types)

1(a) Generate all multi-graphs.
-- Fix the index set S = \{i\_1,...,i\_m\} with a cycle graph: edges i\_1-i\_2, i\_2-i\_3, ..., i\_m-i\_1.
-- A partition sigma of S merges some indices, producing a multi-graph on k nodes. The edge weight w(a,b) equals the total number of edges (in both directions) between blocks a and b after merging.
-- Exclude any multi-graph with self-loops (A\_\{ii\} = 0).

1(b) Check graph isomorphism.
-- Group multi-graphs by simultaneous row-and-column permutation of the edge weight matrix. Pick the lexicographically smallest representative.

1(c) Find linear coefficients.
-- mu(0hat, sigma) = (-1)\^\{m-k\} * prod\_\{i=1\}\^\{k\} (|B\_i| - 1)!
-- Sum these Mobius values over all partitions in the same isomorphism class.

\textgreater{}\textgreater{}\textgreater{} CODE OUTPUT \textless{}\textless{}\textless{}
Write a self-contained Python script defining step1(m) that returns a list of (weight\_matrix, coefficient) pairs.

Sanity checks:
-- m=2: 1 surviving partition, 1 canonical type
-- m=3: 1 surviving partition, 1 canonical type
-- m=4: 4 surviving partitions, 3 canonical types

REMINDERS:
-- In Stage 1, graph isomorphism is checked using the symmetric edge-weight matrix, because before pruning all edge labels are inherited from the symmetric matrix A.
-- The Stage 1 weight\_matrix is only a canonical pre-pruning representation.
-- Stage 2 will convert each weight\_matrix into FAM edge records before pruning.
-- The cycle wraps around: position m connects back to position 1.
-- Self-loop filtering is essential.
\endgroup
			\end{promptbox}
			
			\caption{System prompt and Stage~1 (Merging Process) of the benchmark evaluation prompt.}
			\label{fig:prompt-stage1}
		\end{figure*}
		
		% ====== FIGURE* 2 : Stage 2 ======
		\begin{figure*}[htb!]
			
			\textbf{Stage 2 User Prompt: Pruning Process}
			
			\begin{promptbox}
\begingroup\setlength{\parskip}{0pt}\linespread{0.96}\selectfont\obeylines\obeyspaces%

YOUR TASK (Step 2 of 2): Using the Step 1 output, implement the PRUNING PROCESS that converts each FS term into a computationally efficient SEA or IFS expression.

STEP 2: PRUNING PROCESS (FS terms -\textgreater{} SEA/IFS terms -\textgreater{} Final Formula)

An FAM is represented by:
-- A dictionary node\_label[v], where each node v has a vector label. Initially every node label is ones\_n.
-- A list of oriented edge records (src, dst, matrix\_label). The edge contributes matrix\_label[i\_src, i\_dst] to the Full Sum.
-- Parallel edges are allowed. Do not assume that matrix labels are symmetric after pruning.

Initialization from a Stage-1 weight\_matrix:
-- For each pair u \textless{} v with weight\_matrix[u][v] = w \textgreater{} 0, create one oriented edge record (u, v, A\_hadamard\_w), where A\_hadamard\_w means A Hadamard-powered w times.
-- This initial orientation u -\textgreater{} v is arbitrary but fixed. It is valid because the initial labels are symmetric.
-- After pruning, newly created matrix labels may be non-symmetric, so every later access must respect the stored orientation.

Orientation helper:
For an edge between nodes a and b with stored record (src, dst, M):
-- If src == a and dst == b, then the matrix oriented from a to b is M.
-- If src == b and dst == a, then the matrix oriented from a to b is M.T.
This rule must be used whenever a pruning formula requires a matrix oriented in a specific direction.

2(a) Identify pendants.
-- A Type-I pendant has exactly 1 distinct neighbor, ignoring edge directions.
-- A Type-II pendant has exactly 2 distinct neighbors, ignoring edge directions.

2(b-I) Type-I pruning.
Suppose p is the pendant and h is the hinge. Let v\_p be the node label of p and u\_h be the node label of h.
Collect all edge labels between h and p, oriented from h to p using the orientation helper.
Let M\_prod be their Hadamard product.
Update the hinge label by

    node\_label[h] = node\_label[h] * (M\_prod @ node\_label[p])

where * is element-wise vector product and @ is matrix-vector multiplication.
Remove p and all edges incident to p.

2(b-II) Type-II pruning.
Suppose p is the pendant and x,y are the two hinges. Use a deterministic order for the hinges, for example x \textless{} y.
Collect all edge labels between p and x, oriented from p to x. Let their Hadamard product be Q.
Collect all edge labels between p and y, oriented from p to y. Let their Hadamard product be R.
Create a new oriented edge from x to y with label

    E = Q.T @ diag(node\_label[p]) @ R

Remove p and all edges incident to p. Add the new edge record (x, y, E).

Important:
-- Do not assume E is symmetric.
-- Do not replace an existing x-y edge by E unless you correctly preserve the product of all parallel edge labels.
-- The safest implementation is to keep E as an additional parallel oriented edge record.

Repeat until:
-- If only 1 node remains, output/evaluate an SEA term.
-- If no pendant remains and more than one node remains, output/evaluate an IFS term.

IFS rule:
For an IFS term, each oriented edge record (src, dst, M) contributes M[index[src], index[dst]].
Do not symmetrize M. Do not replace M[index[src], index[dst]] by M[index[dst], index[src]] unless the stored orientation requires a transpose through the orientation helper.

\textgreater{}\textgreater{}\textgreater{} CODE OUTPUT \textless{}\textless{}\textless{}
Produce a self-contained Python script defining C\_m(A, m) that works for any m \textgreater{}= 2.

REMINDERS:
-- Edge orientation matters after pruning.
-- Initial labels are symmetric, but pruning can create non-symmetric matrix labels.
-- Type-I pruning requires all incident edge labels oriented from hinge to pendant.
-- Type-II pruning requires Q and R oriented from pendant to each hinge before computing Q.T @ diag(v\_p) @ R.
-- Keep parallel edges unless you are absolutely sure the Hadamard combination preserves orientation.
-- Carefully distinguish @, *, Hadamard product, transpose, and diag.
-- For IFS terms, use directed edge indices exactly as stored.
\endgroup
			\end{promptbox}
			
			\caption{Stage~2 (Pruning Process) of the benchmark evaluation prompt.}
			\label{fig:prompt-stage2}
		\end{figure*}

\section{The output CEEF formulas}\label{app:formula}
We now present the complete formulas of $C_m$ that we have calculated for $m=3,4, \ldots,12$ as follows. 
We have grouped the terms according to the node count of the original multi-graph.
%Here we note that the binary and non-binary case coincide when $m \leq 6$, so for the binary case we will start from $m=7$.

%\subsection{Non-binary Case}\label{non-binary}

\small 

\allowdisplaybreaks[4]
\vspace{-0.5em}
\subsection{m=3}
\vspace{-0.5em}
\begin{longtable}{@{}p{0.1\linewidth}@{}p{0.9\linewidth}@{}}
	\multicolumn{2}{@{}l@{}}{$C_3 =$} \\
	\multicolumn{2}{@{}l@{}}{\text{Unique graphs for k=3: (1 graphs)}} \\
	+1 & ${\bf 1}_n' \cdot ((A \circ (A \cdot A)) \cdot {\bf 1}_n)$ \\
\end{longtable}

\vspace{-2em}
\subsection{m=4}
%\vspace{-0.5em}
\begin{longtable}{@{}p{0.1\linewidth}@{}p{0.9\linewidth}@{}}
	\multicolumn{2}{@{}l@{}}{$C_4 =$} \\
	\multicolumn{2}{@{}l@{}}{\text{Unique graphs for k=2: (1 graphs)}} \\
	 +1 & ${\bf 1}_n' \cdot ((A \circ A \circ A \circ A) \cdot {\bf 1}_n)$ \\
	\multicolumn{2}{@{}l@{}}{\text{Unique graphs for k=3: (1 graphs)}} \\
	 -2 & ${\bf 1}_n' \cdot (((A \circ A) \cdot {\bf 1}_n) \circ ((A \circ A) \cdot {\bf 1}_n))$ \\
	\multicolumn{2}{@{}l@{}}{\text{Unique graphs for k=4: (1 graphs)}} \\
	 +1 & ${\bf 1}_n' \cdot ((A \circ ((A \cdot A) \cdot A)) \cdot {\bf 1}_n)$ \\
\end{longtable}

\vspace{-2em}
\subsection{m=5}
\vspace{-0.5em}
\begin{longtable}{@{}p{0.1\linewidth}@{}p{0.9\linewidth}@{}}
	\multicolumn{2}{@{}l@{}}{$C_5 =$} \\
	\multicolumn{2}{@{}l@{}}{\text{Unique graphs for k=3: (1 graphs)}} \\
	+5 & ${\bf 1}_n' \cdot ((A \circ A \circ A \circ (A \cdot A)) \cdot {\bf 1}_n)$ \\
    \multicolumn{2}{@{}l@{}}{\text{Unique graphs for k=4: (1 graphs)}} \\
	-5 & ${\bf 1}_n' \cdot (((A \circ A) \cdot {\bf 1}_n) \circ ((A \circ (A \cdot A)) \cdot {\bf 1}_n))$ \\
    \multicolumn{2}{@{}l@{}}{\text{Unique graphs for k=5: (1 graphs)}} \\
	+1 & ${\bf 1}_n' \cdot ((A \circ (((A \cdot A) \cdot A) \cdot A)) \cdot {\bf 1}_n)$ \\
\end{longtable}

\vspace{-2em}
\subsection{m=6}
\vspace{-0.5em}
\begin{longtable}{@{}p{0.1\linewidth}@{}p{0.9\linewidth}@{}}
	\multicolumn{2}{@{}l@{}}{$C_6 =$} \\
	\multicolumn{2}{@{}l@{}}{\text{Unique graphs for k=2: (1 graphs)}} \\
	+4 & ${\bf 1}_n' \cdot ((A \circ A \circ A \circ A \circ A \circ A) \cdot {\bf 1}_n)$ \\
   \multicolumn{2}{@{}l@{}}{\text{Unique graphs for k=3: (2 graphs)}} \\
	-12 & ${\bf 1}_n' \cdot (((A \circ A) \cdot {\bf 1}_n) \circ ((A \circ A \circ A \circ A) \cdot {\bf 1}_n))$ \\
	-4 & ${\bf 1}_n' \cdot ((A \circ A \circ ((A \circ A) \cdot (A \circ A))) \cdot {\bf 1}_n)$ \\
   \multicolumn{2}{@{}l@{}}{\text{Unique graphs for k=4: (4 graphs)}} \\
	+6 & ${\bf 1}_n' \cdot ((A \circ A \circ A \circ ((A \cdot A) \cdot A)) \cdot {\bf 1}_n)$ \\
	+4 & ${\bf 1}_n' \cdot ((((A \circ A) \cdot {\bf 1}_n) \circ ((A \circ A) \cdot {\bf 1}_n)) \circ ((A \circ A) \cdot {\bf 1}_n))$ \\
	+3 & ${\bf 1}_n' \cdot (((A \circ A) \cdot ((A \circ A) \cdot {\bf 1}_n)) \circ ((A \circ A) \cdot {\bf 1}_n))$ \\
	+9 & ${\bf 1}_n' \cdot ((A \circ A \circ (A \cdot A) \circ (A \cdot A)) \cdot {\bf 1}_n)$ \\
   \multicolumn{2}{@{}l@{}}{\text{Unique graphs for k=5: (2 graphs)}} \\
	-6 & ${\bf 1}_n' \cdot (((A \circ A) \cdot {\bf 1}_n) \circ ((A \circ ((A \cdot A) \cdot A)) \cdot {\bf 1}_n))$ \\
	-3 & ${\bf 1}_n' \cdot (((A \circ (A \cdot A)) \cdot {\bf 1}_n) \circ ((A \circ (A \cdot A)) \cdot {\bf 1}_n))$ \\
   \multicolumn{2}{@{}l@{}}{\text{Unique graphs for k=6: (1 graphs)}} \\
	+1 & ${\bf 1}_n' \cdot ((A \circ ((((A \cdot A) \cdot A) \cdot A) \cdot A)) \cdot {\bf 1}_n)$ \\
\end{longtable}

\vspace{-2em}
\subsection{m=7}
\vspace{-0.5em}
\begin{longtable}{@{}p{0.1\linewidth}@{}p{0.9\linewidth}@{}}
	\multicolumn{2}{@{}l@{}}{$C_7 =$} \\
	\multicolumn{2}{@{}l@{}}{\text{Unique graphs for k=3: (2 graphs)}} \\
	+28 & ${\bf 1}_n' \cdot ((A \circ A \circ A \circ A \circ A \circ (A \cdot A)) \cdot {\bf 1}_n)$ \\
	+28 & ${\bf 1}_n' \cdot ((A \circ A \circ A \circ ((A \circ A \circ A) \cdot A)) \cdot {\bf 1}_n)$ \\
   \multicolumn{2}{@{}l@{}}{\text{Unique graphs for k=4: (4 graphs)}} \\
	-14 & ${\bf 1}_n' \cdot (((A \circ A \circ A \circ A) \cdot {\bf 1}_n) \circ ((A \circ (A \cdot A)) \cdot {\bf 1}_n))$ \\
	-56 & ${\bf 1}_n' \cdot (((A \circ A) \cdot {\bf 1}_n) \circ ((A \circ A \circ A \circ (A \cdot A)) \cdot {\bf 1}_n))$ \\
	-7 & ${\bf 1}_n' \cdot ((A \circ A \circ A \circ (A \cdot \mathrm{d}((A \circ A) \cdot {\bf 1}_n) \cdot A)) \cdot {\bf 1}_n)$ \\
\end{longtable}

\vspace{-2em}
\subsection{m=8}
\vspace{-0.5em}
\begin{longtable}{@{}p{0.1\linewidth}@{}p{0.9\linewidth}@{}}
	\multicolumn{2}{@{}l@{}}{$C_8 =$} \\
	\multicolumn{2}{@{}l@{}}{\text{Unique graphs for k=2: (1 graphs)}} \\
	+36 & ${\bf 1}_n' \cdot ((A \circ A \circ A \circ A \circ A \circ A \circ A \circ A) \cdot {\bf 1}_n)$ \\
   \multicolumn{2}{@{}l@{}}{\text{Unique graphs for k=3: (3 graphs)}} \\
	-96 & ${\bf 1}_n' \cdot (((A \circ A) \cdot {\bf 1}_n) \circ ((A \circ A \circ A \circ A \circ A \circ A) \cdot {\bf 1}_n))$ \\
	-36 & ${\bf 1}_n' \cdot (((A \circ A \circ A \circ A) \cdot {\bf 1}_n) \circ ((A \circ A \circ A \circ A) \cdot {\bf 1}_n))$ \\
	-112 & ${\bf 1}_n' \cdot ((A \circ A \circ A \circ A \circ ((A \circ A) \cdot (A \circ A))) \cdot {\bf 1}_n)$ \\
   \multicolumn{2}{@{}l@{}}{\text{Unique graphs for k=4: (11 graphs)}} \\
\end{longtable}

\vspace{-2em}
\subsection{m=9}
\vspace{-0.5em}
\begin{longtable}{@{}p{0.1\linewidth}@{}p{0.9\linewidth}@{}}
	\multicolumn{2}{@{}l@{}}{$C_9 =$} \\
	\multicolumn{2}{@{}l@{}}{\text{Unique graphs for k=3: (3 graphs)}} \\
	+324 & ${\bf 1}_n' \cdot ((A \circ A \circ A \circ A \circ A \circ A \circ A \circ (A \cdot A)) \cdot {\bf 1}_n)$ \\
	+648 & ${\bf 1}_n' \cdot ((A \circ A \circ A \circ A \circ A \circ ((A \circ A \circ A) \cdot A)) \cdot {\bf 1}_n)$ \\
	+176 & ${\bf 1}_n' \cdot ((A \circ A \circ A \circ ((A \circ A \circ A) \cdot (A \circ A \circ A))) \cdot {\bf 1}_n)$ \\
   \multicolumn{2}{@{}l@{}}{\text{Unique graphs for k=4: (11 graphs)}} \\
	-108 & ${\bf 1}_n' \cdot (((A \circ A \circ A \circ A \circ A \circ A) \cdot {\bf 1}_n) \circ ((A \circ (A \cdot A)) \cdot {\bf 1}_n))$ \\
	-648 & ${\bf 1}_n' \cdot (((A \circ A) \cdot {\bf 1}_n) \circ ((A \circ A \circ A \circ A \circ A \circ (A \cdot A)) \cdot {\bf 1}_n))$ \\
\end{longtable}

\vspace{-2em}
\subsection{m=10}
\vspace{-0.5em}
\begin{longtable}{@{}p{0.1\linewidth}@{}p{0.9\linewidth}@{}}
	\multicolumn{2}{@{}l@{}}{$C_{10} =$} \\
	\multicolumn{2}{@{}l@{}}{\text{Unique graphs for k=2: (1 graphs)}} \\
	+576 & ${\bf 1}_n' \cdot ((A \circ A \circ A \circ A \circ A \circ A \circ A \circ A \circ A \circ A) \cdot {\bf 1}_n)$ \\
   \multicolumn{2}{@{}l@{}}{\text{Unique graphs for k=3: (4 graphs)}} \\
	-1440 & ${\bf 1}_n' \cdot (((A \circ A) \cdot {\bf 1}_n) \circ ((A \circ A \circ A \circ A \circ A \circ A \circ A \circ A) \cdot {\bf 1}_n))$ \\
	-960 & ${\bf 1}_n' \cdot (((A \circ A \circ A \circ A) \cdot {\bf 1}_n) \circ ((A \circ A \circ A \circ A \circ A \circ A) \cdot {\bf 1}_n))$ \\
	-1800 & ${\bf 1}_n' \cdot ((A \circ A \circ A \circ A \circ A \circ A \circ ((A \circ A) \cdot (A \circ A))) \cdot {\bf 1}_n)$ \\
	-2160 & ${\bf 1}_n' \cdot ((A \circ A \circ A \circ A \circ ((A \circ A \circ A \circ A) \cdot (A \circ A))) \cdot {\bf 1}_n)$ \\
\end{longtable}

\normalsize

\subsection{$m=11$ and $m=12$}
These formulas are too long to present in the paper. They can be found in the Supplementary Material.

\end{document}